%% file: main.tex
\documentclass[a4paper,fleqn]{cas-sc}

\input{pkg_macro}
\pdfoutput=1

\allowdisplaybreaks[4]
\usepackage[rightcaption]{sidecap}

\newcommand{\comp}{\circ}
\newcommand{\eqn}[1]{{}\medskip\newline\centerline{$\displaystyle#1$}\medskip\newline}

\usepackage[square,numbers]{natbib}

\begin{document}
\let\WriteBookmarks\relax
\def\floatpagepagefraction{1}
\def\textpagefraction{.001}

\shorttitle{Belief Acquisition as Stochastic Filtering}   
\shortauthors{Chen, Lloyd, Yang-Zhao, \& Ng}  
\title [mode = title]{Belief Acquisition as Stochastic Filtering}

\author[1]{Dawei Chen}[orcid=0000-0001-6063-2622]
\ead{dawei.chen@anu.edu.au}
\credit{Data Curation, Investigation, Methodology, Software, Visualization, Writing - Original Draft}

\author[1]{John Lloyd}
\ead{john.lloyd@anu.edu.au}
\credit{Methodology, Supervision, Visualization, Writing - Original Draft, Writing - Review and Editing}

\author[1]{Samuel Yang-Zhao}[orcid=0000-0002-8742-7036]
\ead{samuel.yang-zhao@anu.edu.au}
\credit{Investigation, Software, Writing - Original Draft}

\author[1]{Kee Siong Ng}[orcid=0000-0003-0701-8783]
\ead{keesiong.ng@anu.edu.au}
\credit{Conceptualization, Supervision, Writing - Original Draft}

\affiliation[1]{organization={School of Computing},
            addressline={Australian National University}, 
            city={Canberra},
            postcode={2600}, 
            state={ACT},
            country={Australia}%
}

\cortext[1]{Corresponding authors}

\input{abstract}


\maketitle

\input{introduction}

\input{beliefs}

\input{theory}

\input{epidemic}

\input{factored_cpf}

\input{related_work}

\input{conclusion}

\appendix

\renewcommand*{\thesection}{Appendix~\Alph{section}}

\input{filters_cmp}
\input{transition_observation_update}
\input{KL_derivation}
\input{performance_evaluation}
\input{state_error_derivation}
\input{appendix_practical_consideration}

\bibliographystyle{cas-model2-names}
\bibliography{ref}

\end{document}

%% file: pkg_macro.tex
\usepackage{algorithm}
\usepackage{algorithmic}
\usepackage{enumitem}

\usepackage{tikz}
\usetikzlibrary{calc,fit,positioning,shapes,decorations.pathmorphing}

\newcommand{\alg}{\mathcal{F}}

\newcommand{\NCal}{\mathcal{N}}

\newcommand{\UCal}{\mathcal{U}}

\newcommand{\R}{\mathbb{R}}
\newcommand{\N}{\mathbb{N}}

\newcommand{\Pos}{\texttt{+} }
\newcommand{\Neg}{\texttt{-} }
\newcommand{\Unk}{\texttt{?} }
\newcommand{\sS}{\mathsf{S}}
\newcommand{\sE}{\mathsf{E}}
\newcommand{\sI}{\mathsf{I}}
\newcommand{\sR}{\mathsf{R}}

\newcommand{\dd}{{\; d}}

\newcommand{\ie}{i.e.,\ }
\newcommand{\eg}{e.g.,\ }
\newcommand{\diag}{\, \text{diag}}

\newcommand{\ind}[1]{\mathbb{I}(#1)}

\DeclareMathOperator*{\argmin}{argmin}
\newcommand{\surl}[1]{\urlstyle{same}\url{#1}}

\newcommand{\widebar}[1]{\overline{#1}}
\newcommand{\normal}{\mathcal{N}}

\newcommand{\daweicomment}[1]{\textcolor[rgb]{0.63,0.13,0.94}{#1}\newline}
\renewcommand{\daweicomment}[1]{}

\usepackage{lscape}
\usepackage[all]{xy}
\CompileMatrices

\usepackage{amsmath}
\usepackage{amsthm}
\usepackage{graphicx}

\usepackage{tikz}
\usetikzlibrary{matrix,intersections,arrows,decorations.pathmorphing,backgrounds,positioning,fit,petri}
\usetikzlibrary{decorations.pathmorphing} 
\usetikzlibrary{fit}					
\usetikzlibrary{backgrounds}	

\theoremstyle{plain}

\theoremstyle{definition}
\newtheorem{definition}{Definition}[section]

\newcommand{\expect}{\mathsf{E}}

\newcommand{\prob}{\mathsf{P}}

\newcommand{\bool}{o}

\usepackage{scalerel,stackengine}
\stackMath
\newcommand\reallywidehat[1]{%
\savestack{\tmpbox}{\stretchto{%
  \scaleto{%
    \scalerel*[\widthof{\ensuremath{#1}}]{\kern-.6pt\bigwedge\kern-.6pt}%
    {\rule[-\textheight/2]{1ex}{\textheight}}
  }{\textheight}%
}{0.5ex}}%
\stackon[1pt]{#1}{\tmpbox}%
}

%% file: abstract.tex
\begin{abstract}
This paper studies how belief acquisition can be accomplished using stochastic filtering.  
First, a theoretical foundation for empirical beliefs is outlined.
Then stochastic filtering in this context is studied.
The paper introduces factored conditional filters, new filtering algorithms for simultaneously tracking states and estimating parameters in high-dimensional state spaces.
The conditional nature of the algorithms is used to estimate parameters and the factored nature is used to decompose the state space into low-dimensional subspaces in such a way that filtering on these subspaces gives distributions whose product is a good approximation to the distribution on the entire state space.
The conditions for successful application of the algorithms are that observations be available at the subspace level and that the transition schema can be factored into local transition schemas that are approximately confined to the subspaces; these conditions are widely satisfied in computer science, engineering, and geophysical filtering applications.
Experimental results on tracking epidemics and estimating parameters in large contact networks show the effectiveness of the approach.
\end{abstract}

%% file: introduction.tex
\section{Introduction}
\label{sec:intro}

The context of this paper is the challenge of building agents that require large and complex empirical belief bases to achieve their goals.
Empirical beliefs are beliefs that are acquired from observations by an agent situated in an environment.
Empirical belief bases often consist of a single state distribution that is maintained by stochastic filtering.
However, applications increasingly require the agent to maintain a belief base that consists of numerous beliefs and, furthermore, the beliefs can be conditional beliefs, that is, take the form of a conditional probability density, for example.
Assuming that it is desirable to maintain individual conditional beliefs, the problem then arises as to how this might be achieved.
Since stochastic filtering is an attractive method for belief acquisition, we are led inevitably to the problem of designing conditional filters, that is, filters that operate on conditional beliefs.
As a concrete example that has been widely investigated by a variety of methods, consider the problem of estimating parameters while tracking a state distribution.
The conditional empirical belief for this situation is a conditional probability density that is a mapping from the parameter space to the space of probability densities on the state.
Thus, as we show below, this problem can be naturally modelled as a conditional filtering problem. 
If, in addition, the dimension of the state space is large, there is also the problem of filtering in high dimensions which can often be addressed by factoring.
In this parameter estimation problem, one can make the assumption that the parameter is fixed (but unknown, of course). 
This assumption greatly simplifies the problem and is the setting for this paper.
(The general case, not considered here, is where the parameter space is replaced by an arbitrary space and the stochastic process on this space is arbitrary.)
These considerations supply the motivation for the factored conditional filters of this paper.

To provide a foundation for the filters studied in this paper, a theoretical foundation for empirical beliefs rather more general than is normally considered for stochastic filtering is required.
Thus the definitions of empirical schemas (from which empirical beliefs are derived), transition schemas, and observation schemas are given in appropriate generality.
Furthermore, actions are included and, at least initially, no conditional independence assumptions are made;
thus empirical schemas, transition schemas, and observation schemas all have history arguments.
Then a systematic presentation of twelve filtering algorithms is given. 
These algorithms are deliberately written at a high level of abstraction to provide the clearest possible description of them. 
Practical applications may well require refinements of these descriptions.
In effect, we give a landscape of possible filtering algorithms that could be employed in artificial intelligence (and other) applications.
For any particular application, the idea is to look amongst the algorithms to find which is most appropriate and then perhaps refine it as necessary.

 Of the filtering algorithms presented here, the paper focusses primarily on the three factored conditional algorithms together with the application area, epidemic spreading on contact networks, used for our experiments.
So consider an agent being employed to assist in controlling an epidemic.
The environment for the agent is the territory where the epidemic is taking place, the people and their locations in the territory, transport services between locations in the territory, and so on.
Observations available to the agent could include numbers of people with the disease, numbers of deaths and recoveries, test results, and so on.
Actions for the agent could include advice to human experts about which tests to perform on which people, the location and duration of lockdowns, and so on. 

The most fundamental capability needed of such an agent is to be able to track the state of an epidemic from observations.
Assume that the territory of the epidemic is modelled by a graph, where nodes in the graph could represent any of people, households, neighbourhoods, suburbs, towns, or cities.
There is an undirected edge between each pair of nodes for which transmission of the disease between the two nodes is possible.
Thus the model employs a contact network.
Such a model is more complex than the more typical compartmental model but provides more detail about the epidemic.
A state consists of the contact network with nodes labelled by information about the epidemic at that node, for example, whether the person represented by that node is susceptible, exposed, infectious, or recovered. 
Suppose that the agent knows the transition schema and observation schema of the epidemic. 
Then it can track the epidemic by computing a state distribution at each time step.
The agent uses knowledge of the state distribution to select actions.
This paper is concerned only with the problem of tracking the state of contact networks, not selecting actions.

Considered more abstractly, for a fixed contact network, the state space can be modelled by a product space, where each element of the product space is a particular state of the epidemic.
For given transition and observation schemas, the generic problem is that of filtering on a product space representing a graph.
In addition, the graph may have tens or hundreds of thousands, or even millions, of nodes, so the problem is one of filtering in high dimensions. 
Furthermore, the transition and observation schemas may have parameters that need to be estimated by the filtering process.

In high-dimensional spaces, exact probabilistic inference is usually infeasible; in such cases it is necessary to resort to approximation.
The two main kinds of approximation commonly employed are based on Monte Carlo methods and variational methods.
For filtering, the Monte Carlo method is manifested in the form of particle filtering and we discuss this first.

A well-known fundamental difficulty with particle filtering in high dimensions is that two distinct probability measures in high-dimensional spaces are nearly mutually singular.
(For example, in high dimensions, Gaussian distributions with the identity matrix as covariance matrix have nearly all their probability mass in a thin annulus around a hypersphere with radius $\sqrt{d}$, where $d$ is the dimension of the space.
(See the Gaussian Annulus Theorem in \cite{blum-hopcroft-kannan}.)
It follows that two distinct Gaussian distributions in a high-dimensional space are nearly mutually singular.)
In particular, in high dimensions, the distribution obtained after a transition update and the distribution obtained after the subsequent observation update are nearly mutually singular.
As a consequence, the particle family obtained by resampling gives a poor approximation of the distribution obtained from the observation update.
Typically, what happens is that one particle from the transition update has (normalized) weight very nearly equal to one and so the resampled particle family degenerates to a single particle.
See the discussion about this in \cite{snyder-bengtsson-bickel-anderson}, for example.
Intuitively, the problem could be solved with a large enough particle family.
Unfortunately, it has been shown that to avoid degeneracy the size of the particle family has to be at least exponential in the problem size.
More precisely, the size of the particle family must be exponential in the variance of the observation log likelihood, which depends not only on the state dimension but also on the distribution after the state transition and the number and character of observations.
Simulations confirm this result.
For the details, see \cite{bengtsson-bickel-li,bickel-li-bengtsson, snyder-bengtsson-bickel-anderson}.

In spite of these difficulties, it is often possible to exploit structure in the form of spatial locality of a particular high-dimensional problem to filter effectively, and this will be the case for filtering epidemics.
Let the state space be $\prod_{i=1}^m Y_i$ and $\{ C_1, \ldots, C_p \}$ a partition of the index set 
$\{ 1, \ldots, m \}$.
Suppose, for $l = 1, \ldots, p$, the size of $C_l$ is small and that observations are available for 
subspaces of the form $\prod_{i \in C_l} Y_i$.
It may even be the case that each $C_l$ is a singleton.
Then, since each $C_l$ is small, the degeneracy difficulties mentioned above do not occur for the observation update for each $\prod_{i \in C_l} Y_i$.
In addition, an assumption needs to be made about the transition schema.
It is too much to expect there to be local transition schemas completely confined to each $\prod_{i \in C_l} Y_i$.
If this were true, it would be possible to filter the entire space by independently filtering on each of the subspaces.
But what is often true is that the domain of transition schema for $\prod_{i \in C_l} Y_i$ depends only on a subset of the $Y_j$, where index $j$ is a neighbour of, or at least close by, an index in $C_l$.
The use of such local transition schemas introduces an approximation of the state distribution but, as our experiments here indicate, the error can be surprisingly small even for epidemics on large graphs.
The resulting 
algorithm is called the factored filter, and the particle version of it is 
similar to the local particle filter in \cite{rebeschini,Rebeschini:2015}.
Note that we have adopted the name factored filter that is used in the artificial intelligence literature rather than local filter that is used in the data assimilation literature.

Next we discuss the use of variational methods in filtering.
The general setting for this discussion is that of assumed-density filtering for which the state distribution is approximated by a density from some suitable space of densities, often an exponential subfamily.
Essentially, there is an approximation step at the end of the filtering algorithm in which the density obtained from the observation update is projected into the space of approximating densities.
(See the discussion and references in \cite[Section 18.5.3]{Murphy:2012}.)
Here we concentrate on variational methods for this approximation.
This is based on minimizing a divergence measure between two distributions.
In variational inference~\cite{blei-kucukelbir-mcauliffe, JordanVariationalMethods}, the corresponding optimization problem is to find $\argmin_{q} \it{KL}(q \| p)$,
where $q$ is the approximation of $p$;
in expectation propagation \cite{MinkaUAI,vehtari-gelman}, the corresponding optimization problem is to find $\argmin_{q} \it{KL}(p \| q)$.
To present the variational algorithms at a suitable level of abstraction, the $\alpha$-divergence \cite{bishop2006, MinkaDivergence} 
is employed.
This is defined by 
\[ D_\alpha(p \| q) = \frac{4}{1 - \alpha^2} \left( 1 - \int_Y p^{(1+\alpha)/2} q^{(1-\alpha)/2} \; d \upsilon_Y \right), \]
where $\alpha \in \mathbb{R}$.
Note that 
\begin{gather*}
\lim_{\alpha \rightarrow 1} D_\alpha(p \| q) = \it{KL}(p \| q) \\
\lim_{\alpha \rightarrow -1} D_\alpha(p \| q) = \it{KL}(q \| p),
\end{gather*}
so that, with a suitable choice of $\alpha$, forward (inclusive) KL-divergence or reverse (exclusive) KL-divergence or a combination of these can be specified.
For each application, a value for $\alpha$ and a corresponding optimization algorithm are chosen for use in the variational algorithms. 

Variational methods are an attractive alternative to Monte Carlo methods: often variational methods are faster than Monte Carlo methods and scale better. 
We note a further advantage of variational methods: they produce closed-form versions of state distributions in contrast to the Dirac mixture distributions produced by Monte Carlo methods.
This means that the output of a variational method can be comprehensible to people.
In the context of the general setting for filtering in Section 2 for which the aim is to acquire empirical beliefs, comprehensibility can be crucial for interrogating an agent to understand why it acted the way it did, to check for various kinds of bias, and so on.

If, in addition to tracking a state distribution, it is necessary to estimate parameters, the problem becomes rather more complicated.
Our solution to this problem involves having two filters, a nonconditional one for the parameters and a conditional one for the states.
There is considerable freedom about the nature of each kind of filter and, depending on the application, they can run largely independently of one another.
A key issue for this approach is that, especially in the case of parameter estimation, it is difficult, if not impossible, for the designer to know what the observation schema should be for the parameter filter.
Fortunately, it turns out that the parameter observation schema can be computed from other information that is known.
In the particular case where the parameter filter is a particle filter and state filter is a (non-factored or factored) particle filter, we compare the resulting algorithm with the nested particle filter in \cite{Crisan:2017,Crisan:2018}.

Finally, putting the conditional filter and the factored filter together, we obtain the factored conditional filter that can be used in concert with a nonconditional parameter filter to both track high-dimensional states and estimate parameters.
The parameter filter can be the same as for the non-factored case.
However, the state filter is different to the conditional filter of the preceding paragraph because it is now factored.

In summary, the starting point for our development is the \emph{standard} filter; the particle version of the standard filter is the bootstrap particle filter in \cite{Gordon:1993}.
With such a filter, it is possible to track states in low-dimensional state spaces.
One extension of the standard filter is the \emph{conditional} filter;
with such a filter, it is possible to track states and estimate parameters in low-dimensional state spaces.
A different extension of the standard filter is the \emph{factored} filter;
with such a filter, it is possible to track states in high-dimensional state spaces.
The \emph{factored conditional} filter, which combines the notions of a conditional filter and a factored filter, makes it possible under certain circumstances to track states and estimate parameters in high-dimensional state spaces.
For each of these four cases, there are basic, particle, and variational versions of the corresponding filter.

The main contributions of this paper are the following:
\begin{itemize}
\itemsep 0em
\item 
Based on the concept of an empirical belief, we describe a general setting for filtering. 
Empirical beliefs provide a sophisticated approach to knowledge representation especially suited to artificial intelligence applications.
\item We introduce the concept of a conditional filter which is a straightforward conditionalization of the standard filter and has mathematical properties analogous to the standard filter.
\item We provide a framework for investigating the space of filtering algorithms at a suitable level of abstraction and we place factored conditional filtering in this framework.
\item We propose basic, particle, and variational versions of algorithms for factored conditional filtering, which make it possible for suitable applications to simultaneously track states and estimate parameters
in high-dimensional state spaces.
\item We empirically evaluate the factored conditional filtering algorithms in the application domain of epidemics spreading on contact networks.
For a variety of large real-world contact networks, we demonstrate that the algorithms are able to accurately track epidemic states and estimate parameters.
\end{itemize}

The paper is organised as follows.
Section~\ref{sec:empirical_beliefs} introduces some notation that will be used throughout the paper.
Then a theoretical foundation for empirical beliefs is outlined.
For this, a general setting for an agent situated in some environment is presented, where the agent can apply actions to the environment and can receive observations from the environment.
The basic concepts of an empirical schema, empirical belief, transition schema, and observation schema are defined, concentrating on the conditional setting.
Some remarks situate the approach to knowledge representation in this paper within the traditional logical approach.
Section~\ref{sec:theory} outlines a framework for investigating the space of filtering algorithms that includes the standard, conditional, factored, and factored conditional filters, together with their particle and variational versions, and other relevant details.
Section~\ref{sec:epi} introduces the application of primary interest in this paper: epidemics on contact networks.
It describes the network model of epidemics and compares it with the more commonly used compartmental model. 
In Section~\ref{sec:factored_conditional_filters}, experimental results from the employment of the factored conditional filters to track states and estimate parameters of epidemic processes on contact networks are reported.
Section~\ref{sec:related} covers related work.
The conclusion is given in Section~\ref{sec:conclusion}, which also suggests directions for future research.
Several appendices provide additional material: the first appendix empirically compares our factored conditional particle filter with two related filtering algorithms in the literature, three further appendices give details of some mathematical derivations, the fifth appendix discusses how the performance of filters can be evaluated, and the sixth appendix considers some practical implementation issues.
A GitHub 
repository is available at \texttt{\url{https://github.com/cdawei/filters}} that contains additional details and the complete set of experimental results.

%% file: beliefs.tex
\section{Empirical Beliefs}
\label{sec:empirical_beliefs}

This section introduces the theory of empirical beliefs on which the filtering algorithms in this paper depend.
While some formal definitions are given, the account is rather informal without precise statements of theorems; 
an extensive and rigorous account of the theory underlying the algorithms, and the statements and proofs of the results alluded to in this paper, are given in \cite{lloyd-empirical-beliefs}.
The notation used in the paper will also be introduced here;
it is deliberately as precise as possible to enable rigorous definitions, and statements and proofs of results given in \cite{lloyd-empirical-beliefs}.

\subsection{Notation}

Here are some general notational conventions.
Functions usually have their signature given explicitly: the notation $f : X \rightarrow Y$ means that $f$ is a function with domain $X$ and codomain $Y$.
The notation $f : X \rightarrow Y \rightarrow Z$ means $f : X \rightarrow (Y \rightarrow Z)$, so that the codomain of $f$ is the function space $Z^Y$, consisting of all functions from $Y$ to $Z$.
Denoting by $\mathcal{D}(Y)$ the set of probability densities on the measure space $Y$, a function $f : X \rightarrow \mathcal{D}(Y)$ can also be written as $f : X \rightarrow Y \rightarrow \mathbb{R}$.
The notation $\triangleq$ means `stand(s) for'.

The lambda calculus is a convenient way of compactly and precisely defining anonymous functions. 
Let $X$ and $Y$ be sets, $x$ a variable ranging over values in $X$, and $\varphi$ an expression taking values in $Y$.
Then $\lambda x. \varphi$ denotes the function from $X$ to $Y$ defined by $x \mapsto \varphi$, for all $x \in X$.
The understanding is that, for each element $e$ of $X$, the corresponding value of the function is obtained by replacing each free occurrence of $x$ in $\varphi$ by $e$.
For example, $\lambda x. x^2 : \mathbb{R} \rightarrow \mathbb{R}$ denotes the square function defined by $x \mapsto x^2$, for all $x \in \mathbb{R}$.
For another example, if $f : X \rightarrow Y \rightarrow Z$, then, for all $y \in Y$, $\lambda x. f(x)(y) : X \rightarrow Z$ is the function defined by 
$x \mapsto f(x)(y)$, for all $x \in X$.

The notation $\int_X f \; d \upsilon$ denotes the (Lebesgue) integral of the real-valued, measurable function $f : X \rightarrow \mathbb{R}$, where $X$ is a measure space with measure $\upsilon$.
If $X$ is a Euclidean space, then $\upsilon$ is often Lebesgue measure.
If $X$ is a countable space and $\upsilon$ is counting measure on $X$, then $\int_X f \; d \upsilon = \sum_{x \in X} f(x)$.

$\mathcal{P}(Y)$ denotes the set of probability measures on the measurable space $Y$.
If $\mu_1: \mathcal{P}(X_1)$ and $\mu_2 : X_1 \rightarrow \mathcal{P}(X_2)$, 
then the product $\mu_1 \otimes \mu_2 : \mathcal{P}(X_1 \times X_2)$ of $\mu_1$ and $\mu_2$ is defined by
$\mu_1 \otimes \mu_2 = \lambda A. \int_{X_1} (\lambda x_1. \int_{X_2} \lambda x_2. \mathbf{1}_A(x_1, x_2) \; d \mu_2(x_1))  \; d\mu_1$.
Also the fusion $\mu_1 \odot \mu_2 : \mathcal{P}(X_2)$ of $\mu_1$ and $\mu_2$ is defined by
$\mu_1 \odot \mu_2 = \lambda A_2. \int_{X_1} \lambda x_1. \mu_2(x_1)(A_2)  \; d\mu_1$.
These definitions extend in the obvious way to 
$\mu_1: X_0 \rightarrow \mathcal{P}(X_1)$ and $\mu_2 : X_0 \times X_1 \rightarrow \mathcal{P}(X_2)$.

Let $(X, \mathcal{X}, \nu)$ be a measure space and $h : X \rightarrow \mathbb{R}$
a measurable function.
Then $h$ is a probability density if (i) $h(x) \geq 0$, for all $x \in X$, and (ii) $\int_X h \, d\nu = 1$.
The set of all probability densities on $X$ is denoted by $\mathcal{D}(X)$.

The basic probability space is denoted by $(\Omega, \alg, \prob)$, where $\alg$ is a $\sigma$-algebra on $\Omega$ and $\prob$ is a probability measure on $\alg$.

The conditional expectation of a random variable $f : \Omega \rightarrow \mathbb{R}$ given  $\sigma$-algebra $\mathcal{G}$ is denoted by $\expect(f \, | \, \mathcal{G})$.
The conditional probability of a measurable set $A$ given $\mathcal{G}$,
denoted by $\prob(A \, | \, \mathcal{G})$, is defined to be $\expect(\mathbf{1}_A \, | \, \mathcal{G})$.

$\mathbb{R}$\index{$\mathbb{R}$} is the set of real numbers, $\mathbb{N}$\index{$\mathbb{N}$} is the set of positive integers, and $\mathbb{N}_0$\index{$\mathbb{N}_0$} is the set of non-negative integers.

\subsection{Empirical, Transition, and Observation Schemas}

To assist an agent in deciding which actions to perform in order to achieve its goal(s), 
the agent has a belief base that consists of certain function definitions,
embodying information about, for example, the environment, its location, and beliefs of other agents. 
Included in the belief base are empirical beliefs, those beliefs acquired from observations.
Empirical beliefs are derived from the fundamental concept of an empirical schema. 
For stochastic filtering, also needed are transition schemas and observation schemas.
The key idea for the definitions of these three concepts is that they depend upon the underlying stochastic processes of the setting.
These definitions form the basis of the theory of empirical beliefs; 
for example, they are needed to prove the correctness of the filter recurrence equations for both nonconditional 
and conditional filters.

A measurable space is a pair $(X, \mathcal{X})$, where $X$ is a set and $\mathcal{X}$ is a $\sigma$-algebra on $X$.
If $(X, \cal{X})$ and $(Y, \mathcal{Y})$ are measurable spaces, a measurable function $\mu : X \rightarrow \mathcal{P}(Y)$ is called a probability kernel.

There is an action space $(A, \mathcal{A})$, an observation space $(O, \mathcal{O})$,
an action process $\mathbf{a} : \Omega \rightarrow A^{\mathbb{N}}$, and an observation process $\mathbf{o} : \Omega \rightarrow O^{\mathbb{N}}$.
For all $n \in \mathbb{N}_0$, $H_n \triangleq A \times O \times \cdots \times A \times O$, 
where there are $n$ occurrences of $A$ and $n$ occurrences of $O$, and
$\mathbf{h}_n \triangleq (\mathbf{a}_1, \mathbf{o}_1, \ldots, \mathbf{a}_n, \mathbf{o}_n) : \Omega \rightarrow H_n$.
If $\mathbf{x} : \Omega \rightarrow X^{\mathbb{N}_0}$, then, for all $n \in \mathbb{N}_0$,
$\mathbf{x}_n : \Omega \rightarrow X$ is defined by $\mathbf{x}_n(\omega) = \mathbf{x}(\omega)(n)$, for all $\omega \in \Omega$.\index{$\mathbf{x}_n$}
Similarly, for $\mathbf{y} : \Omega \rightarrow Y^{\mathbb{N}_0}$ and $\mathbf{y}_n : \Omega \rightarrow Y$. 

In the following, the conditional case is considered first; the nonconditional case is considered at the end of this section.
The definitions implicitly employ the probabilistic concept of a regular conditional distribution \cite{kallenberg2021, klenke}.

 Here is the fundamental concept of an empirical schema.

\begin{definition}
\label{empirical_schema_defn}
Let $(\Omega, \alg, \prob)$ be a probability space,
$(A, \mathcal{A})$ an action space, $(O, \mathcal{O})$ an observation space,
$(X, \mathcal{X})$ and $(Y, \mathcal{Y})$ measurable spaces, 
$\mathbf{a} : \Omega \rightarrow A^{\mathbb{N}}$ an action process, 
$\mathbf{o} : \Omega \rightarrow O^{\mathbb{N}}$ an observation process, and
$\mathbf{x} : \Omega \rightarrow X^{\mathbb{N}_0}$ and $\mathbf{y} : \Omega \rightarrow Y^{\mathbb{N}_0}$ stochastic processes.\index{$\mathbf{x}$}
An {\em empirical schema}\index{empirical schema} ({\em for} $\mathbf{y}$ {\em given} $\mathbf{x}$)
is a sequence $\mu \triangleq (\mu_n)_{n \in \mathbb{N}_0}$, where 
\[ \mu_n  : H_n \times X \rightarrow \mathcal{P}(Y) \]
is a regular probability kernel of $\mathbf{y}_n$ given $(\mathbf{h}_n, \mathbf{x}_n)$, for all $n \in \mathbb{N}_0$.
\end{definition}

\begin{figure}[ht]
\framebox[\linewidth]{
\begin{tikzpicture}[-latex]
\matrix [matrix of math nodes, ampersand replacement=\&, row sep=1em, column sep = 6em]
{
                                              \&   |(Y)|  H_n \times X  \&    |(Y1)|   H_n \times X    \\
    |(X)| \Omega   \&                                                        \\
                                              \&   |(Z)|  Y             \&    |(PZ)|  \mathcal{P}(Y)   \\
};
\begin{scope}[every node/.style={midway,auto}]
\draw [thick, -to] (X) to node {$(\mathbf{h}_n, \mathbf{x}_n)$} (Y);
\draw [thick, -to] (X) to node {$\mathbf{y}_n$} (Z);
\draw [thick, -to] (Y1) to node {$\mu_n$} (PZ);
\end{scope}
\end{tikzpicture}
}
\caption{A component of an empirical schema}
\label{empirical_schema_fig}
\end{figure}

Regularity means that, for all $n \in \mathbb{N}_0$, $\mu_n$ is a probability kernel that satisfies the condition
\[ \lambda \omega. \mu_n((\mathbf{h}_n, \mathbf{x}_n)(\omega))(B) = \prob(\mathbf{y}_n^{-1}(B) \; | \; (\mathbf{h}_n, \mathbf{x}_n))  \; \text{a.s.}, \]
for all $B \in \mathcal{Y}$.
Under a weak restriction on $Y$, for each $n \in \mathbb{N}$, such a $\mu_n$ exists 
and is essentially unique.

The definition of each $\mu_n : H_n \times X \rightarrow \mathcal{P}(Y)$ 
is strongly constrained by the requirement that $\mu_n$ must satisfy the regularity assumption.
Intuitively this means that $\mu_n$ represents a conditional probabilistic relationship between the stochastic processes that generate actions in $A$, 
observations in $O$, and the spaces $X$ and $Y$; 
this regularity assumption is crucial in the theory of stochastic filtering.

Next is the definition of an empirical belief.

\begin{definition}
\label{empirical_belief_defn}
Let $(\Omega, \alg, \prob)$ be a probability space,
$(A, \mathcal{A})$ an action space, $(O, \mathcal{O})$ an observation space,
$(X, \mathcal{X})$ and $(Y, \mathcal{Y})$ measurable spaces, 
$\mathbf{a} : \Omega \rightarrow A^{\mathbb{N}}$ an action process, 
$\mathbf{o} : \Omega \rightarrow O^{\mathbb{N}}$ an observation process, and
$\mathbf{x} : \Omega \rightarrow X^{\mathbb{N}_0}$ and $\mathbf{y} : \Omega \rightarrow Y^{\mathbb{N}_0}$ stochastic processes.
An {\em empirical belief} ({\em for} $\mathbf{y}$ {\em given} $\mathbf{x}$)\index{empirical belief} is a probability kernel
\[ \lambda x.\mu_n(h, x) : X \rightarrow \mathcal{P}(Y), \] 
where $(\mu_n : H_n \times X \rightarrow \mathcal{P}(Y))_{n \in \mathbb{N}_0}$ is an empirical schema for $\mathbf{y}$ given $\mathbf{x}$
and $h \in H_n$, for some $n \in \mathbb{N}_0$.
\end{definition}

There is no requirement that an empirical belief be a regular probability kernel and, 
in fact, there cannot be such a requirement since there are 
many $\lambda x. \mu_n(h, x) : X \rightarrow \mathcal{P}(Y)$, each depending on the particular history $h$.

An empirical belief is contingent: given the specific history that has taken place,
for any value in $X$, an empirical belief determines the distribution on the possible values in $Y$.
The adjective `empirical' is apt since an empirical belief depends upon the observations in the history.
An empirical belief is a measure of an agent's state of knowledge about its environment, not necessarily a measure of the environment itself.

 An empirical belief $\lambda x.\mu_n(h, x) : X \rightarrow \mathcal{P}(Y)$ can thought of as a `ground truth' conditional distribution, consistent with the underlying stochastic processes.
An agent's estimate, denoted by $\reallywidehat{\lambda x.\mu_n(h, x)}$, of $\lambda x. \mu_n(h, x)$ is used by the agent to select actions.
In this paper, it is proposed that $\reallywidehat{\lambda x.\mu_n(h, x)}$ be computed by the agent using stochastic filtering.
In only few cases can $\lambda x.\mu_n(h, x)$ be computed exactly, even in principle, by a filter; generally, for tractability reasons, most kinds of filters make approximations.

Throughout, it is assumed that the stochastic process $\mathbf{x} : \Omega \rightarrow X^{\mathbb{N}_0}$ is constant-valued a.s.,
that is, 
\[ \prob (\{ \omega \in \Omega \; | \; \mathbf{x}_n(\omega) = \mathbf{x}_{n+1}(\omega), \text{ for all } n \in \mathbb{N}_0 \}) = 1. \]
This condition, which characterizes the constant-valued conditional setting, is satisfied for the applications in this paper, since $X$ is a space of parameter values that are unknown but fixed;  
the condition simplifies the statements of the conditional filtering algorithms which are more complicated if the condition is dropped. 
The general case where this condition is dropped is discussed in \cite{lloyd-empirical-beliefs}.

After an action is performed and an observation is received, the definition of $\lambda x.\mu_n(h_n, x)$ needs to be updated to 
a modified definition $\lambda x.\mu_{n+1}(h_{n+1}, x)$ at the next time step by the filtering process.
Stochastic filtering in the conditional case is the process of tracking the empirical belief $\lambda x.\mu_n(h_n, x)$ over time.
For this process, a transition schema and an observation schema are needed;
here are their definitions.

\begin{definition}
\label{transition_model_defn2}
Let $(\Omega, \alg, \prob)$ be a probability space, 
$(A, \mathcal{A})$ an action space,
$(O, \mathcal{O})$ an observation space,
$(X, \mathcal{X})$ and $(Y, \mathcal{Y})$ measurable spaces,
$\mathbf{a} : \Omega \rightarrow A^{\mathbb{N}}$ an action process,
$\mathbf{o} : \Omega \rightarrow O^{\mathbb{N}}$ an observation process, and
$\mathbf{x} : \Omega \rightarrow X^{\mathbb{N}_0}$ and $\mathbf{y} : \Omega \rightarrow Y^{\mathbb{N}_0}$ stochastic processes. 
A {\em transition schema}\index{transition schema} ({\em for} $\mathbf{y}$ {\em given} $\mathbf{x}$) is a sequence $\tau \triangleq (\tau_n)_{n \in \mathbb{N}}$,
where 
\[ \tau_n : H_{n-1} \times A \times X \times Y \rightarrow \mathcal{P}(Y) \]
is a regular probability kernel of $\mathbf{y}_n$ given $(\mathbf{h}_{n-1}, \mathbf{a}_n, \mathbf{x}_n, \mathbf{y}_{n-1})$, 
for all $n \in \mathbb{N}$.

Each component $\tau_n : H_{n-1} \times A \times X \times Y \rightarrow \mathcal{P}(Y)$ of $\tau$ is called a {\em transition model}.\index{transition model}
\end{definition}

\begin{figure*}[htbp]
\framebox[\linewidth]{
\begin{tikzpicture}[-latex]
\matrix [matrix of math nodes, ampersand replacement=\&, row sep=1em, column sep = 3em]
{
                   \&   |(Y)|  H_{n-1} \times A \times X \times Y  \&    |(Y1)|   H_{n-1} \times A \times X \times Y      \\
    |(X)| \Omega   \&                                                      \\
                   \&   |(Z)|  Y           \&    |(PZ)|  \mathcal{P}(Y)    \\
};
\begin{scope}[every node/.style={midway,auto}]
\draw [thick, -to] (X) to node {$(\mathbf{h}_{n-1}, \mathbf{a}_n, \mathbf{x}_n, \mathbf{y}_{n-1})$} (Y);
\draw [thick, -to] (X) to node {$\mathbf{y}_n$} (Z);
\draw [thick, -to] (Y1) to node {$\tau_n$} (PZ);
\end{scope}
\end{tikzpicture}
}
\caption{A transition model}
\label{}
\end{figure*}

In other words, for all $n \in \mathbb{N}$, $\tau_n$ is a probability kernel that satisfies the condition
\[ \lambda \omega. \tau_n((\mathbf{h}_{n-1}, \mathbf{a}_n, \mathbf{x}_n, \mathbf{y}_{n-1})(\omega))(C) = 
         \prob(\mathbf{y}_n^{-1}(C) \; | \; (\mathbf{h}_{n-1}, \mathbf{a}_n, \mathbf{x}_n, \mathbf{y}_{n-1})) \; \text{ a.s.}, \]
for all $C \in \mathcal{Y}$.
Under a weak restriction on $Y$, for each $n \in \mathbb{N}$, such a $\tau_n$ exists 
and is essentially unique.

A transition model takes as input an element in $H_{n-1} \times A \times X \times Y$
and returns a distribution on the values in $Y$ that could result from the transition.
Here, $H_{n-1}$ is the set of histories up to the current observation received by the agent.
Having this extra argument is a requirement that comes from replacing a state space by an arbitrary space $Y$;
elements of $Y$ (or even $X \times Y$) may not force the conditional independence properties that states do. 

This signature for a transition schema is unusual in that it includes $H_{n-1}$ in the domain.
It turns out that it is natural for the general theory to include the history space in the domain; 
under a conditional independence assumption, the history argument can be dropped.

Now comes the definition of an observation schema.

\begin{definition}
\label{generalized_observation_model_defn}
Let $(\Omega, \alg, \prob)$ be a probability space, 
$(A, \mathcal{A})$ an action space,
$(O, \mathcal{O})$ an observation space,
$(X, \mathcal{X})$ and $(Y, \mathcal{Y})$ measurable spaces,
$\mathbf{a} : \Omega \rightarrow A^{\mathbb{N}}$ an action process,
$\mathbf{o} : \Omega \rightarrow O^{\mathbb{N}}$ an observation process, and
$\mathbf{x} : \Omega \rightarrow X^{\mathbb{N}_0}$ and $\mathbf{y} : \Omega \rightarrow Y^{\mathbb{N}_0}$ stochastic processes. 
An {\em observation schema} ({\em for} $\mathbf{y}$ {\em given} $\mathbf{x}$)\index{observation schema} is a sequence 
$\xi \triangleq (\xi_n)_{n \in \mathbb{N}}$, where 
\[ \xi_n : H_{n-1} \times A \times X \times Y \rightarrow \mathcal{P}(O) \]
is a regular probability kernel for 
$\mathbf{o}_n$ given $(\mathbf{h}_{n-1}, \mathbf{a}_n, \mathbf{x}_n, \mathbf{y}_n)$, for all $n \in \mathbb{N}$.

Each component $\xi_n : H_{n-1} \times A \times X \times Y \rightarrow \mathcal{P}(O)$ of $\xi$ is called an {\em observation model}.\index{observation model}
\end{definition}

\begin{figure*}[htbp]
\framebox[\linewidth]{
\begin{tikzpicture}[-latex]
\matrix [matrix of math nodes, ampersand replacement=\&, row sep=1em, column sep = 6em]
{
                   \&   |(Y)|  H_{n-1} \times A \times X \times Y      \&    |(Y1)|  H_{n-1} \times A \times X \times Y \\
    |(X)| \Omega   \&                                                                                       \\
                   \&   |(Z)|  O                                       \&    |(PZ)|  \mathcal{P}(O)         \\
};
\begin{scope}[every node/.style={midway,auto}]
\draw [thick, -to] (X) to node {$(\mathbf{h}_{n-1}, \mathbf{a}_n, \mathbf{x}_n, \mathbf{y}_n)$} (Y);
\draw [thick, -to] (X) to node {$\mathbf{o}_n$} (Z);
\draw [thick, -to] (Y1) to node {$\xi_n$} (PZ);
\end{scope}
\end{tikzpicture}
}
\caption{An observation  model}
\label{generalized_observation_model_fig}
\end{figure*}

In other words, for all $n \in \mathbb{N}$, $\xi_n$ is a probability kernel that satisfies the condition
\[ \lambda \omega. \xi_n((\mathbf{h}_{n-1}, \mathbf{a}_n,  \mathbf{x}_n, \mathbf{y}_n)(\omega))(B) =
                   \prob(\mathbf{o}_n^{-1}(B) \; | \; (\mathbf{h}_{n-1}, \mathbf{a}_n, \mathbf{x}_n, \mathbf{y}_n)) \; \text{ a.s.}, \]
for all $B \in \mathcal{O}$.
Under a weak restriction on $O$, for each $n \in \mathbb{N}$, such a $\xi_n$ exists 
and is essentially unique. 

An observation  model takes as input an element in $H_{n-1} \times A \times X \times Y$ and returns a distribution on the observations 
that could be received by the agent.

It turns out that it is natural for the general theory to include the history and action spaces in the domain of an observation schema; 
under a conditional independence assumption, the history and action arguments can be dropped.

The setting above is sufficiently general to provide a theoretical framework for the acquisition of empirical beliefs by an agent situated in some environment for which the constant-valued assumption is satisfied.
Furthermore, empirical beliefs can be far more general than the standard case of state distributions.

Now the nonconditional case of the theory is considered.
For this, the space $X$ in the above development is dropped.
Thus empirical schemas have the form
\[ (\mu_n : H_n \rightarrow \mathcal{P}(Y))_{n \in \mathbb{N}_0}. \]
Regularity of an empirical schema for the nonconditional case means that, for all $n \in \mathbb{N}_0$, $\mu_n$ is a probability kernel that satisfies the condition
\[ \lambda \omega. \mu_n(\mathbf{h}_n(\omega))(B) = \prob(\mathbf{y}_n^{-1}(B) \; | \; \mathbf{h}_n)  \; \text{a.s.}, \]
for all $B \in \mathcal{Y}$.
An empirical belief then has the form $\mu_n(h) : \mathcal{P}(Y)$.
Similarly, a transition schema has the form 
\[ (\tau_n : H_{n-1} \times A \times Y \rightarrow \mathcal{P}(Y))_{n \in \mathbb{N}} \]
and an observation schema has the form
\[ (\xi_n : H_{n-1} \times A \times Y \rightarrow \mathcal{P}(O))_{n \in \mathbb{N}}. \]
The regularity conditions for transition and observations schemas in the nonconditional case are analogous to that for empirical schemas.
This formalization generalizes the usual state distribution setting of stochastic filtering.

For the development of the theory of filtering, $\mathcal{P}(Y)$ is the natural codomain of empirical schemas, empirical beliefs, and transition schemas.
In contrast, for a practical application,
it may be more convenient to work with $\mathcal{D}(Y)$, the set of probability densities on the space $Y$.
The density version of an empirical schema has the form
\[ (\breve{\mu}_n : H_n \rightarrow \mathcal{D}(Y))_{n \in \mathbb{N}_0}, \]
where $\breve{\mu}_n$ denotes the conditional probability density of $\mu_n$ with respect to the underlying measure.
Similarly, the density version of a transition schema has the form 
\[ (\breve{\tau_n} : H_{n-1} \times A \times X \times Y \rightarrow \mathcal{D}(Y))_{n \in \mathbb{N}} \]
and the density version of an observation schema has the form
\[ (\breve{\xi_n} : H_{n-1} \times A \times X \times Y \rightarrow \mathcal{D}(O))_{n \in \mathbb{N}}. \]
Informally, it is possible to regard either $\mathcal{P}(Y)$ or $\mathcal{D}(Y)$ as distributions on $Y$.

Now the discussion turns to the correctness of filtering.
For this, the relevant filtering algorithms below are the standard filter, used in the nonconditional case, and the conditional filter.
(The other ten algorithms use approximations.)
For each of these algorithms, correctness is captured by the filter recurrence relations for each setting; these recurrence relations are expressed directly in the corresponding Algorithms 1 and 4 below.
The key ingredients in the proofs of each of these theorems are the regularity conditions satisfied by the empirical schemas, transition schemas, and observation schemas for each case.
The statements and proofs of the two correctness theorems can be found in \cite{lloyd-empirical-beliefs}.

Note that the space $Y$ above can be arbitrarily structured which provides a rich language for knowledge representation;\,  
the data types thus provided include tuples, lists, strings, sequences, sets, multisets, graphs, and function spaces.
The impact of the structure of $Y$, particularly for belief acquisition, is discussed in 
\cite{lloyd-empirical-beliefs}.

Finally, it is helpful to situate the contribution of this paper in the context of traditional knowledge representation in artificial knowledge that employs various logics. 
The motivation for introducing logics in the early days of artificial intelligence, and still today, is that they provide the ability to reason: that agents reason correctly is crucially important for achieving their goals.
Now logics have two major parts: syntax and semantics.
Syntax is used to express terms and formulas in the logic and to reason,
while semantics provides interpretations for the syntax.
The setting of this paper is entirely in the {\em semantics} of a modal higher-order logic, which is a suitably powerful logic for agents \cite{lloyd-empirical-beliefs, lloyd-ng-JAAMAS}.
In particular, empirical beliefs, transition models, and observation models are function definitions which form part of the intended interpretation.
The semantics of this logic is sufficient for the many kinds of computation an agent must perform in order to select actions that will help it achieve its goals.
Typical such computations are computing expected utility, for some utility function.

However, for reasoning, agents employ the syntactic part of the logic in the form of a theory for which the intended interpretation is a model.
For this purpose, a logicization step is necessary: enough of the intended interpretation must be materialized and there needs to be an algorithm that, from the intended interpretation, produces a sufficiently encompassing theory that has the intended interpretation as a model.
For example, an empirical belief is converted into a formula in the logic.
Then, by the soundness of the logic, any theorem that can be proved from the theory must be true in the intended interpretation.

%% file: theory.tex
\section{A General Setting for Filtering}
\label{sec:theory}

To better appreciate factored conditional filtering that we introduce in this paper, it is useful to place it in the space of filtering algorithms.
For this reason, in the next four subsections, we provide a framework for filters based on their main characteristics, 
such as being factored or conditional, as well as being Monte Carlo or variational.
This leads to twelve filters in all.
There are the four different groups of standard, conditional, factored, and factored conditional.
Inside each group there are three algorithms depending on whether they are basic, particle, or variational filters.
The twelve algorithms are summarized in Table~\ref{tab:filter_algorithms}.

\begin{table*}[htbp]
\centering
\begin{tabular}{@{}rll@{}}
{} & \textbf{Nonconditional} & \textbf{Conditional} \\
\cmidrule{1-3}
\textbf{Nonfactored} & 
    \begin{tabular}{@{}l@{}}
    (1) Standard filter \\
    (2) Standard particle filter \\
    (3) Standard variational filter \\
    \end{tabular} &
    \begin{tabular}{@{}l@{}}
    (4) Conditional filter \\
    (5) Conditional particle filter \\
    (6) Conditional variational filter \\
    \end{tabular} \\
\cmidrule{1-3}
\textbf{Factored} &
    \begin{tabular}{@{}l@{}}
    (7) Factored filter \\
    (8) Factored particle filter \\
    (9) Factored variational filter \\
    \end{tabular} &
    \begin{tabular}{@{}l@{}}
    (10) Factored conditional filter \\
    (11) Factored conditional particle filter \\
    (12) Factored conditional variational filter \\
    \end{tabular} \\
\cmidrule{1-3}
\end{tabular}
\caption{The filtering algorithms}
\label{tab:filter_algorithms}
\end{table*}

Each of the twelve algorithms can be run independently of the others as a standalone algorithm.
But, for many applications, it is the combination of algorithms that is important and this will be the case for the applications considered in this paper.
The following intended use of these filtering algorithms should be kept in mind.

An empirical belief base of an agent typically contains a collection of nonconditional and conditional empirical schemas.
A nonconditional empirical schema of the form $(\nu_n : H_n \rightarrow \mathcal{P}(X))_{n \in \mathbb{N}_0}$ 
and a conditional empirical schema of the form
$(\mu_n : H_n \times X \rightarrow \mathcal{P}(Y))_{n \in \mathbb{N}_0}$ are said to be {\em paired}
if they have the same underlying stochastic process taking values in the common space $X$.
There may be some nonconditional empirical schemas that do not have a paired conditional empirical schema 
and there may be some conditional empirical schemas that do not have a paired nonconditional empirical schema.
Given a pair of empirical schemas, a nonconditional filter is associated with the nonconditional empirical schema and
a conditional filter is associated with the conditional empirical schema.
The corresponding nonconditional filter and conditional filter are said to be {\em paired}.
In principle, for a conditional empirical schema, there is no restriction on the choice of conditional filter and, for a nonconditional empirical schema, there is no restriction on the choice of nonconditional filter. 
(Of course, there are likely to be {\em practical} reasons for choosing one filter over another).
Having this freedom is attractive from a modelling point of view.
Since there are six nonconditional and six conditional filters in Table~\ref{tab:filter_algorithms},
there are 36 possible pairs of filters, all of which have plausible scenarios where they could be employed.

There are compelling reasons for distinguishing conditional filters and nonconditional filters.
Imagine a large and complex empirical belief base for an AI application.
It makes sense to exploit the intrinsic modularity of such an empirical belief base by having a filter associated with each empirical schema, conditional or nonconditional; at each time step, each empirical belief can be updated using its associated filter.
Then, according to the current task of the agent, the results of the updates of several empirical beliefs may be used to perform some computation.
Note, however, that paired nonconditional and conditional filters may need to
coordinate with each other because of the common stochastic process on the space $X$.
The details of how this is handled will be elaborated in Section~\ref{sec:interface}.
Until then, the presentation of the filtering algorithms is at a sufficiently high level of abstraction 
to portray the conditional and the nonconditional filters as independent from one another.

\subsection{Standard Filtering}
\label{sec:theory:filtering}
Suppose now that the empirical belief $\breve{\mu}_{n-1}(h_{n-1}) : \mathcal{D}(Y)$ at time $n-1$ is known.
Given the action $a_n$ applied by the agent and the observation $o_n$ received from the environment, 
the agent needs to compute the empirical belief $\breve{\mu}_n(h_n) : \mathcal{D}(Y)$ at the next time step.
This is achieved by first applying the $n$th component of the transition schema $\breve{\tau}$
to the empirical belief $\breve{\mu}_{n-1}(h_{n-1})$ 
in the transition update to obtain the intermediate distribution 
\[ \lambda y. \int_Y \lambda y'. \breve{\tau}_n(h_{n-1}, a_n, y')(y) \; \breve{\mu}_{n-1}(h_{n-1}) \; d \upsilon_Y \]
on $Y$.
(For a fixed $y$, $\lambda y'. \breve{\tau}_n(h_{n-1}, a_n, y')(y)$  is a real-valued function on $Y$. 
Also $\breve{\mu}_{n-1}(h_{n-1})$  is a real-valued function on $Y$.
Multiply these two functions together and then integrate the product. 
Now let $y$ vary to get the new distribution on $Y$.)
Here, $\upsilon_Y$ is the underlying measure on the space $Y$. 
Then, in the observation update, the observation schema and the actual observation are used to modify the intermediate distribution 
to obtain $\breve{\mu}_n(h_n)$ via
\begin{equation*}
\breve{\mu}_n(h_n) =
K_n^{-1} \, \lambda y. \breve{\xi}_n(h_{n-1}, a_n, y)(o_n) \;\, \lambda y. \int_Y \lambda y'. \breve{\tau}_n(h_{n-1}, a_n, y')(y) \; \breve{\mu}_{n-1}(h_{n-1}) \; d \upsilon_Y,
\end{equation*}
where $K_n$ is a normalization constant.
There is also a probability measure version of the preceding filter recurrence equation
\cite[Section 4.2]{lloyd-empirical-beliefs}.
The corresponding algorithm for the standard filter is given in Algorithm~\ref{algo:sf}.
(This is the basic version of the standard filter.)

\begin{algorithm}[htbp]
\caption{
{\it Standard Filter} \\
{\bf returns} Empirical belief $\breve{\mu}_n(h_n)$ at time $n$ \\
{\bf inputs:} Empirical belief $\breve{\mu}_{n-1}(h_{n-1})$ at time $n-1$, \\
\rule{3.6em}{0pt} history $h_{n-1}$ up to time $n-1$, \\
\rule{3.6em}{0pt} action $a_n$ at time $n$, \\
\rule{3.6em}{0pt} observation $o_n$ at time $n$.
}
\label{algo:sf}

\begin{algorithmic}[1]
\smallskip
\STATE $\displaystyle{\breve{\overline{\mu}}_n(h_{n-1}, a_n) := \lambda y. \int_Y \lambda y'. \breve{\tau}_n(h_{n-1}, a_n, y')(y) \; \breve{\mu}_{n-1}(h_{n-1}) \; d \upsilon_Y}$
\smallskip
\STATE $\displaystyle{\breve{\mu}_n(h_n) := \frac{\lambda y. \breve{\xi}_{n}(h_{n-1}, a_n, y)(o_n) \; \breve{\overline{\mu}}_n(h_{n-1}, a_n)}
{\int_Y \lambda y. \breve{\xi}_n(h_{n-1}, a_n, y)(o_n) \; \breve{\overline{\mu}}_n(h_{n-1}, a_n)  \; d \upsilon_Y}}$
\medskip
\RETURN{$\breve{\mu}_n(h_n)$}
\end{algorithmic}
\end{algorithm}

Particle versions of the filter of Algorithm~\ref{algo:sf} are widely used in practice.
In this case, empirical beliefs are approximated by mixtures of Dirac measures, so that
\[ \mu_n(h_n) \approx \frac{1}{N} \sum_{i=1}^N \delta_{y_n^{(i)}}, \]
where each particle $y_n^{(i)} \in Y$ and each $\delta_{y_n^{(i)}}$ is the Dirac measure at $y_n^{(i)}$.
Particle filters have the advantage of being applicable when closed-form expressions for the $\mu_n(h_n)$ are not available from the filter recurrence equations,
as is usually the case. 
The corresponding particle filter, which can be derived from Algorithm~\ref{algo:sf}, is given in Algorithm~\ref{algo:spf} below.

Given the particle family $(y_{n-1}^{(i)})_{i=1}^N$ at time $n-1$, the history $h_{n-1}$, as well as 
the action $a_n$ and observation $o_n$ at time $n$, the standard particle filter as shown
in Algorithm~\ref{algo:spf} produces the particle family $(y_n^{(i)})_{i=1}^N$ at time $n$ using
the transition schema $\tau$ (via the transition update in Line 2) and the observation schema $\breve{\xi}$ (via the observation update in Line 3).
The resampling step (Line 9) 
with respect to the normalized weights of particles (Line 6) 
serves to alleviate the degeneracy of particles~\cite{Gordon:1993}.
We remark that Algorithm~\ref{algo:spf} requires $O(N)$ transition and observation updates on average, and in practice a more efficient implementation with lower variance is usually preferred. 
(See Appendix~\ref{appendix:prac} for further details.)

\begin{algorithm}[ht]
\caption{
{\it Standard Particle Filter} \\
{\bf returns} Approximation $\frac{1}{N} \sum_{i=1}^N \delta_{y_n^{(i)}}$ of empirical belief $\mu_n(h_n)$ at time $n$ \\
{\bf inputs:} Approximation $\frac{1}{N} \sum_{i=1}^N \delta_{y_{n-1}^{(i)}}$ of empirical belief $\mu_{n-1}(h_{n-1})$ at time $n-1$, \\
\rule{3.6em}{0pt} history $h_{n-1}$ up to time $n-1$, \\
\rule{3.6em}{0pt} action $a_n$ at time $n$, \\
\rule{3.6em}{0pt} observation $o_n$ at time $n$.
}
\label{algo:spf}
\begin{algorithmic}[1]
\medskip
\FOR{$i := 1$ to $N$}
    \STATE sample $\widebar{y}_n^{(i)} \sim \frac{1}{N} \sum_{i'=1}^N \tau_n(h_{n-1}, a_n, y_{n-1}^{(i')})$
    \STATE $w_n^{(i)} := \breve{\xi}_n(h_{n-1}, a_n, \widebar{y}_n^{(i)})(o_n)$
\ENDFOR
\FOR{$i := 1$ to $N$}
    \STATE $\widebar{w}_n^{(i)} := \frac{w_n^{(i)}}{\sum_{i'=1}^N w_n^{(i')}}$
\ENDFOR
\FOR{$i := 1$ to $N$}
    \STATE sample $y_n^{(i)} \sim \sum_{i'=1}^N \widebar{w}_n^{(i')} \delta_{\widebar{y}_n^{(i')}}$
\ENDFOR
\RETURN{$\frac{1}{N} \sum_{i=1}^N \delta_{y_n^{(i)}}$}
\end{algorithmic}
\end{algorithm}

\begin{algorithm}[htbp]
\caption{
{\it Standard Variational Filter} \\
{\bf returns} Approximation $q_n$ of empirical belief $\breve{\mu}_n(h_n)$ at time $n$ \\
{\bf inputs:} Approximation $q_{n-1}$ of empirical belief $\breve{\mu}_{n-1}(h_{n-1})$ at time $n-1$, \\
\rule{3.6em}{0pt} history $h_{n-1}$ up to time $n-1$, \\
\rule{3.6em}{0pt} action $a_n$ at time $n$, \\
\rule{3.6em}{0pt} observation $o_n$ at time $n$.
}
\label{algo:svf}

\begin{algorithmic}[1]
\smallskip
\STATE $\displaystyle{\overline{q}_n := \lambda y. \int_Y \lambda y'. \breve{\tau}_n(h_{n-1}, a_n, y')(y) \; q_{n-1} \; d \upsilon_Y}$
\medskip
\STATE $\displaystyle{p_n := \frac{\lambda y. \breve{\xi}_{n}(h_{n-1}, a_n, y)(o_n) \; \overline{q}_n}
{\int_Y \lambda y. \breve{\xi}_n(h_{n-1}, a_n, y)(o_n) \; \overline{q}_n  \; d \upsilon_Y}}$
\medskip
\STATE $\displaystyle{q_n := \argmin_{q \in Q} D_\alpha(p_n \| q)}$
\smallskip
\RETURN{$q_n$}

\end{algorithmic}
\end{algorithm}

Algorithm~\ref{algo:svf} is a variational version of the standard filter. 
It is  an assumed-density filter for which the approximation step is based on a variational principle.
Line 1 is the transition update using the approximation $q_{n-1}$ of $\breve{\mu}_{n-1}(h_{n-1})$.
In Line 2, the density $p_n$ is obtained from the observation update.
Line 3 is the variational update.
For this, $Q \subseteq \mathcal{D}(Y)$ is a suitable subclass of densities on $Y$ from which the approximation to $\breve{\mu}_n(h_n)$ is chosen; 
for example, $Q$ could be some subclass of the exponential family of distributions.
The density $q_n \in Q$ that minimizes the $\alpha$--divergence $D_\alpha(p_n \| q)$ between $p_n$ and $q \in Q$ is computed.
Depending on the application, a value for $\alpha$ and a corresponding optimization algorithm are chosen. 
Thus Algorithm 3 allows the particular method of variational approximation to depend upon the application.
For example, a common choice for Algorithm 3 could be that of reverse KL-divergence as the divergence measure and variational inference as the approximation method.

\subsection{Conditional Filtering}
\label{sec:theory:conditional}

The discussion starts with some motivation of the need for conditional filters.
Suppose a state space is a product space of the form $X \times Y$ and one wants to filter on this space.
One can, of course, perform filtering directly on $X \times Y$ using one of the filters from 
Section~\ref{sec:theory:filtering}.
But there are situations where this is not desirable.
A common situation is where the dimension of $X \times Y$ is large, so particle filtering on $X \times Y$ is infeasible, but particle filtering on $X$ and exact filtering on $Y$ are both possible.  
This is the setting of Rao-Blackwellized filtering \cite{sarkka-svensson}.
Typically, in this setting, a particle filter is utilized on $X$ and a Kalman filter, conditioned on the values in $X$, is utilized on $Y$. 
This approach reduces the originally infeasible particle filtering problem on $X \times Y$ to particle filtering on the space $X$ having a lower dimension than $X \times Y$ together with efficient Kalman filtering on $Y$.
In the descriptions of Rao-Blackwellized filtering in the literature, particles in the particle family on $X$ are simply concatenated with the corresponding Gaussian distribution on $Y$.
Implicit in this description is the concept of a conditional filter that we abstract and define explicitly below.
We argue that the concept of a conditional filter clarifies the description of Rao-Blackwellized filtering and has many other uses besides.

In the conditional filter setting, the space $X$ becomes an argument of the domains for the empirical, transition, and observation schemas relevant to $Y$.
Thus, in the conditional case,
an empirical schema has the form
\[ (\mu_n : H_n \times X \rightarrow \mathcal{P}(Y))_{n \in \mathbb{N}_0}, \]
a transition schema has the form 
\[ (\tau_n : H_{n-1} \times A \times X \times Y \rightarrow \mathcal{P}(Y))_{n \in \mathbb{N}}, \] 
and an observation schema has the form
\[ (\xi_n : H_{n-1} \times A \times X \times Y \rightarrow \mathcal{P}(O))_{n \in \mathbb{N}}. \] 
In this setting, as discussed above, conditional empirical beliefs have the form 
$\lambda x. \mu_n(h_n, x) : X \rightarrow \mathcal{P}(Y)$, where $h_n \in H_n$.
Thus conditional filtering means filtering {\em conditional} empirical beliefs.

Isolating and abstracting the concept of a conditional filter has a number of advantages:
\begin{enumerate}

\item The conditional filter is a straightforward conditionalization of the standard filter and has mathematical properties analogous to the standard filter; for example, there are conditional filter recurrence equations.

\item Consider a state space of the form $X \times Y$.
Using a nonconditional filter, one can acquire empirical beliefs of the form $\mu_n(h) : \mathcal{P}(X \times Y)$,
where $(\mu_n :H_n \rightarrow \mathcal{P}(X \times Y))_{n \in \mathbb{N}_0}$ is the relevant empirical schema.
However, sometimes it is necessary to factorize such product empirical beliefs.
For this one needs two empirical beliefs: $\mu_n^{(1)}(h) : \mathcal{P}(X)$ acquired by a nonconditional filter and 
$\lambda x. \mu_n^{(2)}(h, x) : X \rightarrow \mathcal{P}(Y)$ acquired by a conditional filter such that 
$\mu_n = \mu_n^{(1)} \otimes \mu_n^{(2)}$.
There are even applications where just the factor $\lambda x. \mu_n^{(2)}(h, x): X \rightarrow \mathcal{P}(Y)$ 
alone is needed;
this can happen if $X$ is a space without any probabilistic significance in the application.

\item It clarifies how Rao-Blackwellized filtering works: there are {\em two} filters, 
one producing empirical beliefs having signature $\mathcal{P}(X)$ and one producing conditional empirical beliefs
having signature $X \rightarrow \mathcal{P}(Y)$ that can then be combined in various ways discussed below.

\item There is considerable freedom in the form each of these filters can take; there is, in general, no need to make the restrictions on the forms of the two filters that Rao-Blackwellisation requires. 

\item The concept of a conditional filter is abstracted so that it can be used in a wide variety of situations, for example, in the FastSLAM algorithm \cite{montemerlo-thrun-koller-wegbreit} and in filtering dynamic Bayesian networks \cite{doucet-freitas-murphy-russell}, both of which are Rao-Blackwellized filtering algorithms; in estimating unknown parameters in transition and observation schemas, as in this paper; and in other applications.

\item Conditional filters will be seen below to be a crucial ingredient in the framework for investigating the space of filtering algorithms that combines conditional filters with various other kinds of filters.

\end{enumerate}

The conditional filter is given in Algorithm~\ref{algo:cf}.
(This is the basic version of the conditional filter.)
It is just the conditional version of the standard filter.
An assumption sufficient to prove the correctness of Algorithm~\ref{algo:cf} is that the stochastic process on $X$ is constant-valued a.s.
This condition is satisfied for the applications concerning estimation of parameters considered in this paper.
There is also a more general version of the conditional filter that does not make this assumption and is needed for general Rao-Blackwellized filtering applications. 
For the details, see \cite{lloyd-empirical-beliefs}.

\begin{algorithm*}[ht]
\caption{
{\it Conditional Filter} \\
{\bf returns} Empirical belief $\lambda x. \breve{\mu}_n(h_n, x)$ at time $n$ \\
{\bf inputs:} Empirical belief $\lambda x. \breve{\mu}_{n-1}(h_{n-1}, x)$ at time $n-1$, \\
\rule{3.6em}{0pt} history $h_{n-1}$ up to time $n-1$, \\
\rule{3.6em}{0pt} action $a_n$ at time $n$, \\
\rule{3.6em}{0pt} observation $o_n$ at time $n$.
}
\label{algo:cf}

\begin{algorithmic}[1]
\smallskip
\STATE $\displaystyle{\lambda x. \breve{\overline{\mu}}_n(h_{n-1}, a_n, x) := \lambda x. \lambda y. \int_Y \lambda y'. \breve{\tau}_n(h_{n-1}, a_n, x, y')(y) \; \breve{\mu}_{n-1}(h_{n-1}, x) \; d \upsilon_Y}$ \\ 
\medskip
\STATE $\displaystyle{\lambda x. \breve{\mu}_n(h_n, x) := \lambda x. \frac{\lambda y. \breve{\xi}_n(h_{n-1}, a_n, x, y)(o_n) \; \breve{\overline{\mu}}_n(h_{n-1}, a_n, x)}
{\int_Y \lambda y. \breve{\xi}_n(h_{n-1}, a_n, x, y)(o_n) \; \breve{\overline{\mu}}_n(h_{n-1}, a_n, x) \; d \upsilon_Y}}$
\medskip
\RETURN{$\lambda x. \breve{\mu}_n(h_n, x)$}
\end{algorithmic}
\end{algorithm*}

There is a particle version of Algorithm~\ref{algo:cf}. 
For this, we need to define a probability kernel mapping from $X$ to Dirac mixtures on $Y$ and
that requires an indexed family of functions $(\phi_n : X \rightarrow Y^M)_{n \in \mathbb{N}_0}$.
Intuitively, $\phi_n(x) = (\phi_n(x)^{(1)}, \ldots, \phi_n(x)^{(M)}) $ is a tuple of $M$ particles in $Y$ that correspond to the value $x \in X$ at time $n$.
Thus $\lambda x. \frac{1}{M} \sum_{j=1}^M \delta_{\phi_n(x)^{(j)}} : X \rightarrow \mathcal{P}(Y)$ is a conditional Dirac mixture, for all $n \in \mathbb{N}_0$.
The conditional particle filter is given in Algorithm~\ref{algo:cpf}.
For this algorithm, the common case is that the empirical belief $\lambda x. \mu_n(h_n, x)$ is defined for all $x \in X$.
However, this may not always be the case; see Section~\ref{sec:partial_def}.

\begin{algorithm}[ht]
\caption{
{\it Conditional Particle Filter} \\
{\bf returns} Approximation $\lambda x. \frac{1}{M} \sum_{j=1}^M \delta_{\phi_n(x)^{(j)}}$ of empirical belief $\lambda x. \mu_n(h_n, x)$ at time $n$ \\
{\bf inputs:} Approximation $\lambda x. \frac{1}{M} \sum_{j=1}^M \delta_{\phi_{n-1}(x)^{(j)}}$ of empirical belief $\lambda x. \mu_{n-1}(h_{n-1}, x)$ \\
\rule{3.6em}{0pt} at time $n-1$, \\
\rule{3.6em}{0pt} history $h_{n-1}$ up to time $n-1$, \\
\rule{3.6em}{0pt} action $a_n$ at time $n$, \\
\rule{3.6em}{0pt} observation $o_n$ at time $n$.
}
\label{algo:cpf}

\begin{algorithmic}[1]
\medskip
\FOR{$x \in X$}
    \FOR{$j := 1$ to $M$}
        \STATE sample $\widebar{\phi}_n(x)^{(j)} \sim \frac{1}{M} \sum_{j'=1}^M \tau_n ( h_{n-1}, \, a_n, \, x, \, \phi_{n-1}(x)^{(j')})$
        \STATE $w_n(x)^{(j)} := \breve{\xi}_n ( h_{n-1}, \, a_n, \, x, \, \widebar{\phi}_n(x)^{(j)} )(o_n)$
    \ENDFOR
    \FOR{$j := 1$ to $M$}
        \STATE $\widebar{w}_n(x)^{(j)} := \frac{w_n(x)^{(j)}}{\sum_{j'=1}^M w_n(x)^{(j')}}$
    \ENDFOR
    \FOR{$j := 1$ to $M$}
        \STATE sample $\phi_n(x)^{(j)} \sim \sum_{j'=1}^M \widebar{w}_n(x)^{(j')} \delta_{\widebar{\phi}_n(x)^{(j')}}$
    \ENDFOR
\ENDFOR
\RETURN{$\lambda x. \frac{1}{M} \sum_{j=1}^M \delta_{\phi_n(x)^{(j)}}$}
\end{algorithmic}
\end{algorithm}

To end this subsection, Algorithm~\ref{algo:cvf} is the conditional variational filter.
Let $Q \subseteq \mathcal{D}(Y)$ be a suitable class of densities; 
for example, $Q$ could be a subclass of the exponential family of distributions.
The approximation to $\lambda x. \breve{\mu}_n(h_n, x)$ is chosen from the space $Q^X$ of conditional densities.

\begin{algorithm}[ht]
\caption{
{\it Conditional Variational Filter} \\
{\bf returns} Approximation $q_n$ of empirical belief $\lambda x. \breve{\mu}_n(h_n, x)$ at time $n$ \\
{\bf inputs:} 
Approximation $q_{n-1}$ of empirical belief $\lambda x. \breve{\mu}_{n-1}(h_{n-1}, x)$ at time $n-1$, \\
\rule{3.6em}{0pt} history $h_{n-1}$ up to time $n-1$, \\
\rule{3.6em}{0pt} action $a_n$ at time $n$, \\
\rule{3.6em}{0pt} observation $o_n$ at time $n$.
}
\label{algo:cvf}

\begin{algorithmic}[1]
\smallskip
\STATE $\displaystyle{\overline{q}_n := \lambda x. \lambda y. \int_Y \lambda y'. \breve{\tau}_n(h_{n-1}, a_n, x, y')(y) \; q_{n-1}(x) \; d \upsilon_Y}$
\medskip
\STATE $\displaystyle{p_n := \lambda x. \frac{\lambda y. \breve{\xi}_n(h_{n-1}, a_n, x, y)(o_n) \; \overline{q}_n(x)}
{\int_Y \lambda y. \breve{\xi}_n(h_{n-1}, a_n, x, y)(o_n) \; \overline{q}_n(x) \; d \upsilon_Y}}$
\medskip
\STATE $\displaystyle{q_n : = \lambda x. \argmin_{q \in Q} D_\alpha(p_n(x) \| q)}$
\RETURN{$q_n$}
\end{algorithmic}
\end{algorithm}

\subsection{Factored  Filtering}
\label{sec:theory:factored}

Next, factored filtering is discussed.
Commonly, the space $Y$ is a product space, say, $\prod_{i=1}^m Y_i$.
In the high-dimensional case, particle filtering is known to have difficulties with particle degeneracy;
see, for example, \cite{djuric-bugallo, doucet-johansen, snyder-bengtsson-bickel-anderson}. 
The standard filter can also become computationally infeasible in the high-dimensional case.
This paper explores a solution in certain circumstances to this problem based on partitioning the index set for the product into smaller subsets,
as studied in \cite{BoyenKoller:1998,djuric-bugallo,Ng:2002, Rebeschini:2015}, for example. 
Filters are then employed on the product subspaces based on each of these subsets of indices. 
From the distributions on each of these product subspaces, a distribution on the entire product space can be obtained which approximates the actual distribution. 
In the case where the dimensions of the subspaces are low and distributions on the subspaces are `almost' independent, 
the approximation to the actual distribution on the entire space can be quite accurate. 

Consider the lattice of all partitions of the index set $\{ 1, \ldots, m \}$ for the product space $\prod_{i=1}^m Y_i$.
Each element of a partition is called a cluster.
The partial order $\leq$ on this lattice is defined by $\mathfrak{A} \leq \mathfrak{B}$ if, for each cluster $C \in \mathfrak{A}$,
there exists a cluster $D \in \mathfrak{B}$ such that $C \subseteq D$.
It is said that $\mathfrak{A}$ is finer than $\mathfrak{B}$ or, equivalently, $\mathfrak{B}$ is coarser than $\mathfrak{A}$.
The coarsest of all partitions is $\{ \{ 1, \ldots, m \} \}$;
the finest of all partitions is $\{ \{ 1 \}, \ldots, \{m \} \}$.

The basic idea of factored filters is to choose a suitable partition of the index set 
$\{ 1, \ldots, m \}$ and employ filters 
on the subspaces of $\prod_{i=1}^m Y_i$ that are obtained by forming the product subspace associated with each of the clusters in the partition.
Thus let $\{ C_1, \ldots, C_p \}$ be a partition of the index set $\{ 1, \ldots, m \}$.
By permuting indices if necessary, it can be assumed that $\prod_{l = 1}^p \prod_{i \in C_l} Y_i$ can be identified with $\prod_{i=1}^m Y_i$.
For $l = 1, \ldots, p$, $y_{C_l}$ denotes a typical element of $\prod_{i \in C_l} Y_i$.
Similarly, $(y_{C_1}, \ldots, y_{C_p})$ denotes a typical element of $\prod_{i=1}^m Y_i$.

The observation space corresponding to the $l$th cluster is $O_l$, for $l = 1, \ldots, p$.
Thus $O = \prod_{l=1}^p O_l$.
An observation for the $l$th cluster is denoted by $o^{(l)} \in O_l$.
A history for all the clusters together at time $n$ is
\[ h_n \triangleq (a_1, (o_1^{(1)}, \ldots, o_1^{(p)}), a_2, (o_2^{(1)}, \ldots, o_2^{(p)}), \ldots, a_n, (o_n^{(1)}, \ldots, o_n^{(p)})), \]
while a history for the $l$th cluster at time $n$ is
\[ h_n^{(l)} \triangleq (a_1, o_1^{(l)}, a_2, o_2^{(l)}, \ldots, a_n, o_n^{(l)}). \]
If $H_n$ is the set of all histories at time $n$ for all the clusters together, then $H_n^{(l)}$ denotes the set of all histories at time $n$ for the $l$th cluster.

To get started, the transition and the observation schemas are assumed to have particular factorizable forms 
depending on the choice of partition $\{ C_1, \ldots, C_p \}$.
(In the following, it will be convenient to work with products of densities rather than products of probability measures.
If $f_i : \mathcal{D}(Y_i)$ is a density, for $i = 1, \ldots, n$,
then $\bigotimes_{i=1}^n f_i : \mathcal{D}(\prod_{i=1}^n Y_i)$ is defined by
$\bigotimes_{i=1}^n f_i = \lambda (y_1, \ldots, y_n). \prod_{i=1}^n f_i(y_i)$.)\smallskip

The transition schema
\[ (\breve{\tau}_n : H_{n-1} \times A \times \prod_{i=1}^m Y_i \rightarrow \mathcal{D}(\prod_{i=1}^m Y_i))_{n \in \mathbb{N}} \]
is defined by
\[ \breve{\tau}_n = \lambda (h, a, (y_{C_1}, \ldots, y_{C_p})). \bigotimes_{l=1}^p \breve{\tau}_n^{(l)}(h^{(l)}, a, y_{C_l}), \]
for all $n \in \mathbb{N}$, where
\[ (\breve{\tau}_n^{(l)} : H_{n-1}^{(l)} \times A \times \prod_{i \in C_l} Y_i \rightarrow \mathcal{D}(\prod_{i \in C_l} Y_i))_{n \in \mathbb{N}} \]
is the transition schema associated with the cluster $C_l$, for $l = 1, \ldots, p$.

The observation schema
\[ (\breve{\xi}_n : H_{n-1} \times A \times \prod_{i=1}^m Y_i \rightarrow \mathcal{D}(\prod_{l=1}^p O_l))_{n \in \mathbb{N}} \]
is defined by
\[ \breve{\xi}_n = \lambda (h, a, (y_{C_1}, \ldots, y_{C_p})). \bigotimes_{l=1}^p \breve{\xi}_n^{(l)}(h^{(l)}, a, y_{C_l}), \]
for all $n \in \mathbb{N}$, where
\[ (\breve{\xi}_n^{(l)} : H_{n-1}^{(l)} \times A \times \prod_{i \in C_l} Y_i \rightarrow \mathcal{D}(O_l))_{n \in \mathbb{N}} \]
is the observation schema associated with the cluster $C_l$, for $l = 1, \ldots, p$.

Then, with this setting, under some conditional independence assumptions
\cite[Section A.16]{lloyd-empirical-beliefs} that also appear in FastSLAM algorithm setting in the context of factorizing the map space, it can be shown that the empirical schema 
\[ (\breve{\mu}_n : H_n \rightarrow \mathcal{D}(\prod_{i=1}^m Y_i))_{n \in \mathbb{N}_0} \] 
is given by
\[ \breve{\mu}_n = \lambda h. \bigotimes_{l=1}^p \breve{\mu}_n^{(l)}(h^{(l)}), \]
for all $n \in \mathbb{N}_0$, where
\[ (\breve{\mu}_n^{(l)} : H_n^{(l)} \rightarrow \mathcal{D}(\prod_{i \in C_l} Y_i))_{n \in \mathbb{N}_0} \] 
is the empirical schema associated with the cluster $C_l$, for $l = 1, \ldots, p$.
Thus an empirical belief on the entire space $\prod_{i=1}^m Y_i$ factorizes exactly into the product of empirical beliefs 
on each $\prod_{i \in C_l} Y_i$.

In summary so far, under the above assumptions for the factorization of the transition and observation schemas, 
the distribution $\breve{\mu}_n(h_n)$ on $\prod_{i=1}^m Y_i$ is the product of the marginal distributions $\breve{\mu}_n^{(l)}(h_n^{(l)})$ on the $\prod_{i \in C_l} Y_i$.
Furthermore, instead of running a filter on $\prod_{i=1}^m Y_i$, one can equivalently run (independent) filters on each $\prod_{i \in C_l} Y_i$.
These are nice results, but the problem is that the conditions under which they hold are too strong for most practical situations.
Instead it is necessary to weaken the assumption on the transition schema somewhat.
However, since for high-dimensional spaces it is only feasible to filter locally on each $\prod_{i \in C_l} Y_i$ 
rather than globally on $\prod_{i=1}^m Y_i$, the price to be paid for weakening the assumption is that an approximation is thus introduced.

Consider now the assumptions as above, except assume instead that, for $l = 1, \ldots, p$, the transition schema $\breve{\tau}^{(l)}$ has the form
\[ (\breve{\tau}_n^{(l)} : H_{n-1}^{(l)} \times A \times \prod_{i=1}^m Y_i \rightarrow \mathcal{D}(\prod_{i \in C_l} Y_i))_{n \in \mathbb{N}}. \]
So now the $l$th transition schema depends not just on $\prod_{i \in C_l} Y_i$ but on $\prod_{i=1}^m Y_i$ or,  more realistically, on $\prod_{i \in D} Y_i$,
for some set $D \subseteq \{ 1, \ldots, m\}$ that is a little larger than $C_l$.
Also, by definition,  
\[ \breve{\tau}_n = \lambda (h, a, y). \bigotimes_{l=1}^p \breve{\tau}_n^{(l)}(h^{(l)}, a, y), \]
for all $n \in \mathbb{N}$.
Everything else is the same.
Now assume that $\breve{\mu}_n(h_n)$ is obtained by first computing $\breve{\mu}_n^{(l)}(h_n^{(l)})$, for $l = 1, \ldots, p$, 
and then using the approximation $\breve{\mu}_n(h_n) \approx \bigotimes_{l=1}^p \breve{\mu}_n^{(l)}(h_n^{(l)})$.
Then the above change to each transition schema $\breve{\tau}^{(l)}$ means that $\breve{\mu}_n(h_n)$ and each $\breve{\mu}_n^{(l)}(h_n^{(l)})$ can generally only be computed approximately.
However, there are applications where these approximations are close; a factored filter exploits this situation.

The factored filter is given in Algorithm~\ref{algo:ff}.
(This is the basic version of the factored filter.)
It consists of largely independent filters on the subspaces generated by each of the $p$ clusters.
However, the filters on each of these subspaces are not completely independent because the variable $y'$ in the transition schema ranges over the entire product space $\prod_{i=1}^m Y_i$, not just the subspace $\prod_{i \in C_l} Y_i$ generated by the cluster $C_l$.
In Line 2, the distribution $\bigotimes_{l'=1}^p q_{n-1}^{(l')}$ approximates the empirical belief on the entire product space $\prod_{i=1}^m Y_i$ at time $n-1$. 
Also, in Line 2, $y$ ranges over $\prod_{i \in C_l} Y_i$,  while $y'$ ranges over $\prod_{i=1}^m Y_i$    
because of the extension of the domain of the transition schema.
In contrast, each observation schema is completely confined to the subspace generated by the relevant cluster.
Thus, in Line 3, the integral in the normalization constant is computed using an integral over the space $\prod_{i \in C_l} Y_i$.

\begin{algorithm*}[htbp]
\caption{
{\it Factored Filter} \\
{\bf returns} Approximation $q_n^{(l)}$ of empirical belief $\breve{\mu}_n^{(l)}(h_n^{(l)})$ at time $n$, for $l = 1, \ldots, p$ \\
{\bf inputs:} Approximation $q_{n-1}^{(l)}$ of empirical belief $\breve{\mu}_{n-1}^{(l)}(h_{n-1}^{(l)})$ at time $n-1$, for $l = 1, \ldots, p$, \\
\rule{3.6em}{0pt} history $h_{n-1}^{(l)}$ up to time $n-1$, for $l = 1, \ldots, p$, \\
\rule{3.6em}{0pt} action $a_n$ at time $n$, \\
\rule{3.6em}{0pt} observation $o_n^{(l)}$ at time $n$, for $l = 1, \ldots, p$.
}
\label{algo:ff}

\begin{algorithmic}[1]
\medskip

\FOR{$l := 1$ to $p$}
\STATE $\displaystyle{\overline{q}_n^{(l)} :=\lambda y. \int_{\prod_{i=1}^m Y_i} \lambda y'. \breve{\tau}_n^{(l)}(h_{n-1}^{(l)}, a_n, y')(y) \; \bigotimes_{l'=1}^p q^{(l')}_{n-1} \; d \bigotimes_{i=1}^m \upsilon_{Y_i}}$ \\
\medskip
\STATE $\displaystyle{q_n^{(l)} := \frac{\lambda y. \breve{\xi}_n^{(l)}(h_{n-1}^{(l)}, a_n, y)(o_n^{(l)}) \; \overline{q}_n^{(l)}}
{\int_{\prod_{i \in C_l} Y_i} \lambda y. \breve{\xi}_n^{(l)}(h_{n-1}^{(l)}, a_n, y)(o_n^{(l)}) \; \overline{q}_n^{(l)}  \; d \bigotimes_{i \in C_l} \upsilon_{Y_i}}}$
\ENDFOR 
\RETURN{$q_n^{(l)}$, for $l = 1, \ldots, p$}
\end{algorithmic}
\end{algorithm*}

Two special cases of Algorithm~\ref{algo:ff} are of interest. 
First, if there is only a single cluster which contains all the nodes (\ie $p = 1$), then Algorithm~\ref{algo:ff} is the same as the standard filter in Algorithm~\ref{algo:sf} (for a product space). 
Second, if each cluster is a singleton (\ie $p = m$), then  Algorithm~\ref{algo:ff} is called the \emph{fully factored} filter.

There is a particle version of Algorithm~\ref{algo:ff}.
The factored particle filter is given in Algorithm~\ref{algo:fpf}. 
A factored particle family is just a tuple of particle families.
Besides an additional loop over the $p$ clusters, this algorithm parallels Algorithm~\ref{algo:spf} with the transition update in Line 3, the observation update in Line 4, and the resampling step in Line 10.
Note that the transition update works by sampling from the `product' of the particle families of all clusters which approximates the actual distribution for the entire product space (given by the empirical belief $\breve{\mu}_n(h_n)$ at time $n$).

\begin{algorithm*}[htbp]
\caption{
{\it Factored Particle Filter} \\
{\bf returns} Approximation $\frac{1}{N_l} \sum_{i_l=1}^{N_l} \delta_{y_n^{(C_l, i_l)}}$ of empirical belief $\mu_n^{(l)}(h_n^{(l)})$ 
at time $n$, \\
\rule{3.6em}{0pt} for $l = 1, \ldots, p$ \\
{\bf inputs}: Approximation $\frac{1}{N_l} \sum_{i_l=1}^{N_l} \delta_{y_{n-1}^{(C_l, i_l)}}$ of empirical belief $\mu_{n-1}^{(l)}(h_{n-1}^{(l)})$ at time $n-1$, \\
\rule{3.6em}{0pt} for $l = 1, \ldots, p$, \\
\rule{3.6em}{0pt} history $h_{n-1}^{(l)}$ up to time $n-1$, for $l = 1, \ldots, p$, \\
\rule{3.6em}{0pt} action $a_n$ at time $n$, \\
\rule{3.6em}{0pt} observation $o_n^{(l)}$ at time $n$, for $l = 1, \ldots, p$.
}
\label{algo:fpf}
\begin{algorithmic}[1]
\medskip
\FOR{$l := 1$ to $p$}
  \FOR{$i_l := 1$ to $N_l$}
      \STATE sample $\overline{y}^{(C_l,i_l)}_n \sim \frac{1}{\prod_{l=1}^p N_l} \sum_{i_1'=1}^{N_1} \ldots \sum_{i_p'=1}^{N_p} \tau_n^{(l)}(h_{n-1}^{(l)}, a_n, (y^{(C_1,i_1')}_{n-1}, \ldots, y^{(C_p,i_p')}_{n-1}))$
      \STATE $w^{(C_l,i_l)}_n := \breve{\xi}_n^{(l)}(h_{n-1}^{(l)}, a_n, \overline{y}^{(C_l,i_l)}_n)(o_n^{(l)})$
  \ENDFOR
  
  \FOR{$i_l := 1$ to $N_l$}
      \STATE $\widebar{w}^{(C_l,i_l)}_n := \frac{w^{(C_l,i_l)}_n}{\sum_{i_l'=1}^{N_l} w^{(C_l,i_l')}_n}$
  \ENDFOR
  
  \FOR{$i_l := 1$ to $N_l$}
      \STATE sample $y^{(C_l,i_l)}_n \sim \sum_{i_l'=1}^{N_l}  \widebar{w}^{(C_l,i_l')}_n \delta_{\overline{y}^{(C_l,i_l')}_n}$
  \ENDFOR
\ENDFOR
\RETURN{$\frac{1}{N_l} \sum_{i_l=1}^{N_l} \delta_{y_n^{(C_l, i_l)}}$, for $l = 1, \ldots, p$}
\end{algorithmic}
\end{algorithm*}

Two special cases of Algorithm~\ref{algo:fpf} are of interest. 
First, if there is only a single cluster which contains all the nodes (\ie $p = 1$), then Algorithm~\ref{algo:fpf} is the same as the standard particle filter in Algorithm~\ref{algo:spf} (for a product space). 
Second, if each cluster is a singleton (\ie $p = m$), then Algorithm~\ref{algo:fpf} is called the \emph{fully factored} particle filter.

Algorithm~\ref{algo:fpf} is now compared with the block particle filter in \cite{Rebeschini:2015}.
For the block particle filter, transition and observation schemas are first defined at the node level.
Then, for whatever partition is chosen, transition and observation schemas are defined at the cluster level by forming products of the node-level transition and observation schemas.
Thus, for the block particle filter, transition and observation schemas have a particular factorizable form at the cluster level.
For Algorithm~\ref{algo:fpf}, the partition is chosen first and then transition and observation schemas at the cluster level are defined without the restrictions of the block particle filter.
Thus Algorithm~\ref{algo:fpf} is more general in this regard than the block particle filter; otherwise the two algorithms are the same. 
Furthermore, for the fully factored case when each cluster is a single node, the two algorithms are the same.
Theorem 2.1 in \cite{Rebeschini:2015} that gives an approximation error bound for the block particle filter also applies to Algorithm~\ref{algo:fpf} (at least restricted to the form of transition and observation schemas that are allowed for the block particle filter).

Algorithm~\ref{algo:fvf} is the factored variational filter.
For this, $Q^{(l)}$ is a suitable subclass of approximating densities in $\mathcal{D}(\prod_{i \in C_l} Y_i)$, for $l = 1, \ldots, p$.
Note that the variational update in Line 4 is likely to be simpler than the corresponding step in Algorithm~\ref{algo:svf} 
because here each $\prod_{i \in C_l} Y_i$ is likely to be a low-dimensional space; 
in fact, in the fully factored case, each $\prod_{i \in C_l} Y_i$ is one-dimensional.

\begin{algorithm*}[ht]
\caption{
{\it Factored Variational Filter} \\
{\bf returns} Approximation $q_n^{(l)}$ of empirical belief $\breve{\mu}_n^{(l)}(h_n^{(l)})$ at time $n$, for $l = 1, \ldots, p$ \\
{\bf inputs:} Approximation $q_{n-1}^{(l)}$ of empirical belief $\breve{\mu}_{n-1}^{(l)}(h_{n-1}^{(l)})$ at time $n-1$, for $l = 1, \ldots, p$, \\
\rule{3.6em}{0pt} history $h_{n-1}^{(l)}$ up to time $n-1$, for $l = 1, \ldots, p$, \\
\rule{3.6em}{0pt} action $a_n$ at time $n$, \\
\rule{3.6em}{0pt} observation $o_n^{(l)}$ at time $n$, for $l = 1, \ldots, p$.
}
\label{algo:fvf}

\begin{algorithmic}[1]
\medskip

\FOR{$l := 1$ to $p$}
\STATE $\displaystyle{\overline{q}_n^{(l)} :=\lambda y. \int_{\prod_{i=1}^m Y_i} \lambda y'. \breve{\tau}_n^{(l)}(h_{n-1}^{(l)}, a_n, y')(y) \; \bigotimes_{l'=1}^p q^{(l')}_{n-1} \; d \bigotimes_{i=1}^m \upsilon_{Y_i}}$ \\
\medskip
\STATE $\displaystyle{p_n^{(l)} := \frac{\lambda y. \breve{\xi}_n^{(l)}(h_{n-1}^{(l)}, a_n, y)(o_n^{(l)}) \; \overline{q}_n^{(l)}}
{\int_{\prod_{i \in C_l} Y_i} \lambda y. \breve{\xi}_n^{(l)}(h_{n-1}^{(l)}, a_n, y)(o_n^{(l)}) \; \overline{q}_n^{(l)}  \; d \bigotimes_{i \in C_l} \upsilon_{Y_i}}}$
\medskip
\STATE $\displaystyle{q_n^{(l)} := \argmin_{q \in Q^{(l)}} D_\alpha(p_n^{(l)} \| q)}$
\ENDFOR 
\RETURN{$q_n^{(l)}$, for $l = 1, \ldots, p$}
\end{algorithmic}
\end{algorithm*}

To conclude this subsection, here are some remarks concerning modelling with the factored filter.
In practice, the designers of an agent system are trying to model a partly unknown real world and, as such, there will inevitably be approximations in the model. 
The choice of a particular partition of the index set will normally be an approximation to ground truth. 
And then there is the approximation that comes from the extension of the domain of a transition schema outside its relevant cluster. 
What comes first is the choice of partition; the clusters are likely to be largely determined by the availability of observation schemas. 
From the point of view of avoiding degeneracy, the smaller the cluster sizes the better. 
From the point of view of accuracy of the approximation, bigger cluster sizes are better. 
Thus there is a trade-off. 
Having made the choice of partition, the (potential) need to extend the domain of a transition schema outside its cluster is a consequence of this choice.

\subsection{Factored Conditional Filtering}

Since there are filtering applications where it is necessary to estimate parameters in situations for which the state space is high-dimensional, it is natural to put the two ideas of conditioning and factorization together in one algorithm. 
For this setting, the transition and observation schemas are assumed to have particular forms depending on the choice of partition $\{ C_1, \ldots, C_p \}$; they are conditional versions of the transition and observation schemas from Section~\ref{sec:theory:factored}.

The transition schema
\[ (\breve{\tau}_n : H_{n-1} \times A \times X \times \prod_{i=1}^m Y_i \rightarrow \mathcal{D}(\prod_{i=1}^m Y_i))_{n \in \mathbb{N}} \]
is defined by
\[ \breve{\tau}_n = \lambda (h, a, x, y). \bigotimes_{l=1}^p \breve{\tau}_n^{(l)}(h^{(l)}, a, x, y), \]
for all $n \in \mathbb{N}$, where
\[ (\breve{\tau}_n^{(l)} : H_{n-1}^{(l)} \times A \times X \times \prod_{i=1}^m Y_i \rightarrow \mathcal{D}(\prod_{i \in C_l} Y_i))_{n \in \mathbb{N}} \]
is the transition schema associated with the cluster $C_l$, for $l = 1, \ldots, p$.

The observation schema
\[ (\breve{\xi}_n : H_{n-1} \times A \times X \times \prod_{i=1}^m Y_i \rightarrow \mathcal{D}(\prod_{l=1}^p O_l))_{n \in \mathbb{N}} \]
is defined by
\[ \breve{\xi}_n = \lambda (h, a, x, (y_{C_1}, \ldots, y_{C_p})). \bigotimes_{l=1}^p \breve{\xi}_n^{(l)}(h^{(l)}, a, x, y_{C_l}), \]
for all $n \in \mathbb{N}$, where
\[ (\breve{\xi}_n^{(l)} : H_{n-1}^{(l)} \times A \times X \times \prod_{i \in C_l} Y_i \rightarrow \mathcal{D}(O_l))_{n \in \mathbb{N}} \]
is the observation schema associated with the cluster $C_l$, for $l = 1, \ldots, p$.

Then the empirical schema 
\[ (\breve{\mu}_n : H_n \times X \rightarrow \mathcal{D}(\prod_{i=1}^m Y_i))_{n \in \mathbb{N}_0} \] 
is approximated by
\[ \breve{\mu}_n \approx \lambda (h, x). \bigotimes_{l=1}^p \breve{\mu}_n^{(l)}(h^{(l)},x), \]
for all $n \in \mathbb{N}_0$, where
\[ (\breve{\mu}_n^{(l)} : H_n^{(l)} \times X \rightarrow \mathcal{D}(\prod_{i \in C_l} Y_i))_{n \in \mathbb{N}_0} \] 
is the empirical schema associated with the cluster $C_l$, for $l = 1, \ldots, p$.

Thus what is needed is an algorithm for computing (an approximation of) the empirical belief $\lambda x. \breve{\mu}_n^{(l)}(h_n^{(l)}, x)$, 
for $l = 1, \ldots, p$ and all $n \in \mathbb{N}_0$.
This is achieved by combining Algorithm~\ref{algo:cf} and  Algorithm~\ref{algo:ff} in the obvious way to give the factored conditional filter that is Algorithm~\ref{algo:fcf}.
(This is the basic version of the factored conditional filter.)
With the factored conditional filter, it is possible, for applications with the appropriate structure, to accurately track states 
in high-dimensional spaces and estimate parameters. 

\begin{algorithm*}[ht]
\caption{
{\it Factored Conditional Filter} \\
{\bf returns} Approximation $q_n^{(l)}$ of empirical belief $\lambda x. \breve{\mu}_n^{(l)}(h_n^{(l)}, x)$ at time $n$, for $l = 1, \ldots, p$ \\
{\bf inputs:} Approximation $q_{n-1}^{(l)}$ of empirical belief $\lambda x. \breve{\mu}_{n-1}^{(l)}(h_{n-1}^{(l)}, x)$ at time $n-1$, \\
\rule{3.6em}{0pt} for $l = 1, \ldots, p$, \\
\rule{3.6em}{0pt} history $h_{n-1}^{(l)}$ up to time $n-1$, for $l = 1, \ldots, p$, \\
\rule{3.6em}{0pt} action $a_n$ at time $n$, \\
\rule{3.6em}{0pt} observation $o_n^{(l)}$ at time $n$, for $l = 1, \ldots, p$.
}
\label{algo:fcf}

\begin{algorithmic}[1]
\medskip

\FOR{$l := 1$ to $p$}
\STATE $\displaystyle{\overline{q}_n^{(l)} := \lambda x. \lambda y. \int_{\prod_{i=1}^m Y_i} \lambda y'. \breve{\tau}_n^{(l)}(h_{n-1}^{(l)}, a_n, x, y')(y) \; \bigotimes_{l'=1}^p q_{n-1}^{(l')}(x) \; d \bigotimes_{i=1}^m \upsilon_{Y_i}}$
\medskip
\STATE $\displaystyle{q_n^{(l)} := \lambda x. \frac{\lambda y. \breve{\xi}_n^{(l)}(h_{n-1}^{(l)}, a_n, x, y)(o_n^{(l)}) \;  \overline{q}_n^{(l)}(x)}
{\int_{\prod_{i \in C_l} Y_i} \lambda y. \breve{\xi}_n^{(l)}(h_{n-1}^{(l)}, a_n, x, y)(o_n^{(l)}) \;  \overline{q}_n^{(l)}(x) \; d \bigotimes_{i \in C_l} \upsilon_{Y_i}}}$
\ENDFOR 
\RETURN{$q_n^{(l)}$, for $l = 1, \ldots, p$}
\end{algorithmic}
\end{algorithm*}

There is also a particle version of Algorithm~\ref{algo:fcf}.
For this, we need to define a probability kernel mapping from $X$ to Dirac mixtures on $\prod_{i \in C_l} Y_i$ and
that requires an indexed family of functions $(\phi_n^{(l)} : X \rightarrow (\prod_{i \in C_l} Y_i)^{M_l})_{n \in \mathbb{N}_0}$, for $l = 1, \ldots, p$.
Intuitively, $\phi_n^{(l)}(x) = (\phi_n^{(l)}(x)^{(1)}, \ldots, \phi_n^{(l)}(x)^{(M_l)})$ is a tuple of $M_l$ particles in $\prod_{i \in C_l} Y_i$ that correspond to the value $x \in X$ at time $n$.
Thus $\lambda x. \frac{1}{M_l} \sum_{j_l=1}^{M_l} \delta_{\phi_n^{(l)}(x)^{(j_l)}} : X \rightarrow \mathcal{P}(\prod_{i \in C_l} Y_i)$ is a conditional Dirac mixture, for all $n \in \mathbb{N}_0$ and for $l = 1, \ldots, p$.
The factored conditional particle filter is given in Algorithm~\ref{algo:fcpf} below.

\begin{algorithm*}[ht]
\caption{
{\it Factored Conditional Particle Filter} \\
{\bf returns} Approximation $\lambda x. \frac{1}{M_l} \sum_{j_l=1}^{M_l} \delta_{\phi_n^{(l)}(x)^{(j_l)}}$ 
of empirical belief $\lambda x. \mu_n^{(l)}(h_n^{(l)}, x)$ at time $n$, \\
\rule{3.6em}{0pt} for $l = 1, \ldots, p$ \\
{\bf inputs:} Approximation $\lambda x. \frac{1}{M_l} \sum_{j_l=1}^{M_l} \delta_{\phi_{n-1}^{(l)}(x)^{(j_l)}}$ 
of empirical belief $\lambda x. \mu_{n-1}^{(l)}(h_{n-1}^{(l)}, x)$ at time \\
\rule{3.6em}{0pt}  $n-1$, for $l = 1, \ldots, p$, \\
\rule{3.6em}{0pt} history $h_{n-1}^{(l)}$ up to time $n-1$, for $l = 1, \ldots, p$, \\
\rule{3.6em}{0pt} action $a_n$ at time $n$, \\
\rule{3.6em}{0pt} observation $o_n^{(l)}$ at time $n$, for $l = 1, \ldots, p$.
}
\label{algo:fcpf}
\begin{algorithmic}[1]
\medskip
\FOR{$x \in X$}
    \FOR{$l :=1$ to $p$}
        \FOR{$j_l := 1$ to $M_l$}
            \STATE sample $\widebar{\phi}_n^{(l)}(x)^{(j_l)} \sim \frac{1}{\prod_{l=1}^p M_l} \sum_{j'_1=1}^{M_1} \ldots \sum_{j'_p=1}^{M_p} \tau_n^{(l)}(h_{n-1}^{(l)}, a_n, x, (\phi_{n-1}^{(l)}(x)^{(j'_1)}, \ldots, \phi_{n-1}^{(l)}(x)^{(j'_p)}))$
            \STATE $w_n^{(l)}(x)^{(j_l)} := \breve{\xi}_n^{(l)}(h_{n-1}^{(l)}, a_n, x, \widebar{\phi}_n^{(l)}(x)^{(j_l)})(o_n^{(l)})$
        \ENDFOR
    
        \FOR{$j_l := 1$ to $M_l$}
            \STATE $\widebar{w}_n^{(l)}(x)^{(j_l)} := \frac{w_n^{(l)}(x)^{(j_l)}}{\sum_{j'_l=1}^{M_l} w_n^{(l)}(x)^{(j'_l)}}$
        \ENDFOR
    
        \FOR{$j_l := 1$ to $M_l$}
            \STATE sample $\phi_n^{(l)}(x)^{(j_l)} \sim \sum_{j'_l=1}^{M_l} \widebar{w}_n^{(l)}(x)^{(j'_l)} \delta_{\widebar{\phi}_n^{(l)}(x)^{(j'_l)}}$
        \ENDFOR
    \ENDFOR
\ENDFOR
\RETURN{$\lambda x. \frac{1}{M_l} \sum_{j_l=1}^{M_l} \delta_{\phi_n^{(l)}(x)^{(j_l)}}$}, for $l = 1, \ldots, p$
\end{algorithmic}
\end{algorithm*}

Note that Algorithm~\ref{algo:fcpf} is applicable in more situations than Algorithm~\ref{algo:fcf}, for the same reasons that the standard particle filter is more widely applicable than the standard filter.

Algorithm~\ref{algo:fcvf} is the factored conditional variational filter.
For this, $Q^{(l)} \subseteq \mathcal{D}(\prod_{i \in C_l} Y_i)$ is a family of densities such that $(Q^{(l)})^X$ is 
a suitable space of approximating conditional densities for $\lambda x. \breve{\mu}_n^{(l)}(h_n^{(l)}, x)$, for $l = 1, \ldots, p$.

\begin{algorithm*}[ht]
\caption{
{\it Factored Conditional Variational Filter} \\
{\bf returns} Approximation $q_n^{(l)}$ of empirical belief $\lambda x. \breve{\mu}_n^{(l)}(h_n^{(l)}, x)$ at time $n$, for $l = 1, \ldots, p$ \\
{\bf inputs:} 
Approximation $q_{n-1}^{(l)}$ of empirical belief $\lambda x. \breve{\mu}_{n-1}^{(l)}(h_{n-1}^{(l)}, x)$ at time $n-1$, \\
\rule{3.6em}{0pt} for $l = 1, \ldots, p$, \\
\rule{3.6em}{0pt} history $h_{n-1}^{(l)}$ up to time $n-1$, for $l = 1, \ldots, p$, \\
\rule{3.6em}{0pt} action $a_n$ at time $n$, \\
\rule{3.6em}{0pt} observation $o_n^{(l)}$ at time $n$, for $l = 1, \ldots, p$.
}
\label{algo:fcvf}

\begin{algorithmic}[1]
\medskip

\FOR{$l := 1$ to $p$}
\STATE $\displaystyle{\overline{q}_n^{(l)} := \lambda x. \lambda y. \int_{\prod_{i=1}^m Y_i} \lambda y'. \breve{\tau}_n^{(l)}(h_{n-1}^{(l)}, a_n, x, y')(y) \; \bigotimes_{l'=1}^p q_{n-1}^{(l')}(x) \; d \bigotimes_{i=1}^m \upsilon_{Y_i}}$
\medskip
\STATE $\displaystyle{p_n^{(l)} := \lambda x. \frac{\lambda y. \breve{\xi}_n^{(l)}(h_{n-1}^{(l)}, a_n, x, y)(o_n^{(l)}) \;  \overline{q}_n^{(l)}(x)}
{\int_{\prod_{i \in C_l} Y_i} \lambda y. \breve{\xi}_n^{(l)}(h_{n-1}^{(l)}, a_n, x, y)(o_n^{(l)}) \; \overline{q}_n^{(l)}(x)  \; d \bigotimes_{i \in C_l} \upsilon_{Y_i}}}$
\medskip
\STATE $\displaystyle{q_n^{(l)} : = \lambda x. \argmin_{q \in Q^{(l)}} D_\alpha(p_n^{(l)}(x) \| q)}$
\ENDFOR 
\RETURN{$q_n^{(l)}$, for $l = 1, \ldots, p$}
\end{algorithmic}
\end{algorithm*}

\subsection{Interface between conditional and nonconditional filters}
\label{sec:interface}

Several issues concerning the interface between conditional and nonconditional filters are now discussed.
Consider a nonconditional empirical schema, transition schema, and observation schema
\begin{gather*}
(\nu_n : H_n \rightarrow \mathcal{P}(X))_{n \in \mathbb{N}_0}, \\
(\eta_n : H_{n-1} \times A \times X \rightarrow \mathcal{P}(X))_{n \in \mathbb{N}}, \text{ and} \\
(\zeta_n : H_{n-1} \times A \times X \rightarrow \mathcal{P}(O))_{n \in \mathbb{N}}, 
\end{gather*}
and a conditional empirical schema, transition schema, and observation schema
\begin{gather*}
(\mu_n : H_n \times X \rightarrow \mathcal{P}(Y))_{n \in \mathbb{N}_0}, \\
(\tau_n : H_{n-1} \times A \times X \times Y \rightarrow \mathcal{P}(Y))_{n \in \mathbb{N}}, \text{ and} \\
(\xi_n : H_{n-1} \times A \times X \times Y \rightarrow \mathcal{P}(O))_{n \in \mathbb{N}}. 
\end{gather*}
Suppose that the stochastic process on $X$ for $\nu$ is the same as the stochastic process on $X$ for $\mu$, so 
that the two empirical schemas are paired.
Let the nonconditional filter $F_\nu$ be associated with $\nu$ and the conditional filter $F_\mu$ be associated with $\mu$.
We consider what information may need to be passed between $F_\nu$ and $F_\mu$ in certain particular circumstances.

\subsubsection{Observation schema synthesis}
\label{sec:obs_synthesis}

It can happen that $F_\nu$ needs access to $F_\mu$.
Suppose the designer does not know the observation schema for $F_\nu$.
In the context of a typical nonconditional empirical belief in a belief base, this is likely to be a rare occurrence.
However, it can happen when $X$ is a parameter space, the setting of the experiments in this paper.

There is a result concerning the observation update in Line 3 of Algorithm~\ref{algo:spf} (for the space $X$),
\[ w_n^{(i)} := \breve{\zeta}_n(h_{n-1}, a_n, \widebar{x}_n^{(i)})(o_n), \]
that makes it possible to estimate parameters using the combination of a nonconditional and a conditional filter.
The transition schema $\tau$, the observation schema $\xi$, and the transition schema $\eta$ can be expected to be known by the designer of the system (except perhaps for some parameters in $\tau$ and $\xi$).
But the observation schema $\zeta$ may not be obvious at all for the parameter estimation problem.
Fortunately, $\zeta$ can be computed at each time step from other information that is known.

For this, note that because of the requirements of the definitions of 
empirical schemas, transition schemas, and observation schemas, 
$\nu$, $\eta$, $\zeta$, $\mu$, $\tau$, and $\xi$ are not independent of one another.
For example, there is a result which states that, for all $n \in \mathbb{N}_0$,
\eqn{%
\zeta_{n+1} = (\lambda (h, a, x). \mu_n(h, x) \odot \tau_{n+1}) \odot \xi_{n+1}, }
a.e. for a certain distribution.
A proof of this result is given in \cite[Section 4.3]{lloyd-empirical-beliefs}, where it is called observation schema synthesis.
The proof requires a natural conditional independence assumption and the assumption that, almost surely, the stochastic process on $X$ produces the same value in $X$ at each time step.

As consequence of this result, one can show that, for all $n \in \mathbb{N}_0$ and (almost) all $h_n \in H_n$, $a_{n+1} \in A$, $x \in X$, 
and $o \in O$,
\begin{equation*}
\begin{aligned}
   \breve{\zeta}_{n+1}(h_n, a_{n+1}, x)(o) 
   &= \int_Y \lambda y. \breve{\xi}_{n+1}(h_n, a_{n+1}, x, y)(o) \; d (\mu_n(h_n, x) \odot \lambda y'. \tau_{n+1}(h_n, a_{n+1}, x, y')).
\end{aligned}
\end{equation*}
To understand this result, note that the probability measure $\mu_n(h_n, x) \odot \lambda y'. \tau_{n+1}(h_n, a_{n+1}, x, y')$ gives the state distribution after the transition update; then the $Y$ argument in
$\lambda y. \breve{\xi}_{n+1}(h_n, a_{n+1}, x, y)(o)$ is integrated out using this updated state distribution to give $\breve{\zeta}_{n+1}(h_n, a_{n+1}, x)(o)$.
Thus $\breve{\zeta}_{n+1}$ can be constructed from the known $\breve{\xi}_{n+1}$, $\mu_n$, and $\tau_{n+1}$.
This result is crucial for our approach: since the component $\breve{\zeta}_{n+1}$ of the observation schema can be constructed from the known $\breve{\xi}_{n+1}$, $\mu_n$, and $\tau_{n+1}$, everything that is needed for the statement of the parameter filter is available. 

It follows that the observation update in Line 3 of Algorithm~\ref{algo:spf} (for the space $X$) can be replaced by
\vspace{-.7em}
\[ 
w_n^{(i)} := \int_Y \lambda y. \breve{\xi}_n (h_{n-1}, a_n, \overline{x}_n^{(i)}, y)(o_n) \dd (\mu_{n-1}(h_{n-1}, \overline{x}_n^{(i)}) \odot \lambda y. \tau_n(h_{n-1}, a_n, \overline{x}_n^{(i)}, y)).
\vspace{-.5em}
\]
where the integral can be approximated using Monte Carlo integration by sampling from the probability measure 
$\mu_{n-1}(h_{n-1}, \overline{x}_n^{(i)}) \odot \lambda y. \tau_n(h_{n-1}, a_n, \overline{x}_n^{(i)}, y)$.
This is enough to allow the particle filter $F_\nu$ on $X$ to estimate the parameters accurately provided the approximation of $\mu_{n-1}(h_{n-1}, \overline{x}_n^{(i)})$ by the conditional filter $F_\mu$ is accurate enough.

The observation schema synthesis result is applicable whenever the unknown $\zeta$ occurs in a nonconditional filter, under the conditions stated above.
Conversely, if the observation schema $\zeta$ is known a priori for any of the six nonconditional filters (for the space $X$), then those filters do not need to communicate at run time with their paired conditional filter.
(This does not preclude the possible need for the conditional filter to communicate with the nonconditional filter; see below.)

In the experimental results reported in this paper, the parameter filter is always the standard particle filter Algorithm~\ref{algo:spf} using observation schema synthesis (and a jittering transition schema; see below).

\subsubsection{Partial Definitions of Conditional Empirical Beliefs}
\label{sec:partial_def}

As can be seen from the descriptions of the conditional filters, the corresponding conditional empirical beliefs are defined for all $x \in X$.
For the purposes of some reasoning tasks, this may be essential.
However, for certain pairings of filters, a partial definition may suffice.
Here is a typical such situation.

Consider Algorithm~\ref{algo:cpf}, for example.
(Analogous remarks apply to Algorithm~\ref{algo:fcpf}.)
Suppose the nonconditional filter on $X$ that is paired with Algorithm~\ref{algo:cpf} is the particle filter 
Algorithm~\ref{algo:spf}, for the space $X$. 
(Analogous remarks apply to Algorithm~\ref{algo:fpf}.)
Then $\phi_n$ may be defined for all $x \in X$ or some smaller subset, but it must at least be
defined on the elements $x^{(1)}_n, \ldots, x^{(N)}_n$ of $X$ that are in the particle family at time $n$.
This latter requirement can be achieved as follows.

Suppose that $x \in X$ and $x = x^{(i)}_n$, for some $i \in \{ 1, \ldots, N \}$.
The problem is to know the value of $\phi_{n-1}(x)$ for use in Line 3 of Algorithm~\ref{algo:cpf}. 
Now recall that, for the correctness of the conditional filter, the unknown value in $X$ is assumed to be fixed.
This implies that the transition schema $\eta$ for the nonconditional filter Algorithm~\ref{algo:spf} should be a no-op having the definition $\eta_n = \lambda (h, a, x). \delta_x$, for all $n \in \mathbb{N}$.
In this case, the transition update in Line 2 of Algorithm~\ref{algo:spf} is
\vspace{-.5em}
\[ \text{sample } \overline{x}^{(i)}_n \sim \frac{1}{N} \sum_{i'=1}^N \delta_{x^{(i')}_{n-1}}. 
\vspace{-.3em}
\]
If Algorithm~\ref{algo:cpf} is paired with this version of Algorithm~\ref{algo:spf},
then $x^{(i)}_n = x^{(i^*)}_{n-1}$, for some $i^* \in \{ 1, \ldots, N \}$.
Thus $\phi_{n-1}(x)$ is $\phi_{n-1}(x^{(i^*)}_{n-1})$.
This is used in Line 3 of Algorithm~\ref{algo:cpf}.

However, in practice, this property of the transition schema being a no-op can lead to degeneracy of the parameter particle family. 
Thus a jittering transition schema is used instead.
Rather than keep the parameter fixed, the transition schema perturbs it slightly, thus avoiding degeneracy, as shown in~\cite{Chopin:2013, Crisan:2018}.
Commonly, the parameter space $X$ is $\mathbb{R}^m$, for some $m \geq 1$.
For this case, $\breve{\eta}_n$ is defined, for all $n \in \mathbb{N}$, by
\vspace{-.3em}
\begin{equation}
    \label{eq:jittering_transition_model}
    \breve{\eta}_n = \lambda (h, a, x). \mathcal{N}(x, \Sigma_n),
\vspace{-.3em}
\end{equation}
where $\mathcal{N}(\mu, \Sigma)$ denotes the Gaussian density with mean $\mu$ and covariance matrix $\Sigma$, and $\Sigma_n$ is a (diagonal) covariance matrix with suitably small diagonal values that may change over time according to some criterion in case the jittering is adaptive.
With this modification, Line 2 of Algorithm~\ref{algo:spf} now becomes
\vspace{-.7em}
\begin{equation*}
\text{sample } \widebar{x}_n^{(i)} \sim 
\frac{1}{N} \sum_{i'=1}^N \NCal(x_{n-1}^{(i')}, \, \Sigma_n).
\vspace{-.5em}
\end{equation*}
If Algorithm~\ref{algo:cpf} is paired with this version of Algorithm~\ref{algo:spf},
then $x^{(i)}_n \sim \mathcal{N}(x^{(i^*)}_{n-1}, \Sigma_n)$, for some $i^* \in \{ 1, \ldots, N \}$.
Thus $x^{(i)}_n$ is a perturbation of $x^{(i^*)}_{n-1}$.
Then $\phi_{n-1}(x^{(i^*)}_{n-1})$ replaces $\phi_{n-1}(x)$ and this is used in Line 3 of Algorithm~\ref{algo:cpf}.

The version of Algorithm~\ref{algo:spf} that is used in this paper for estimating parameters in the applications employs such a jittering transition schema.

\subsubsection{Computing $\nu_n(h_n) \otimes \lambda x. \mu_n(h_n, x)$ and $\nu_n(h_n) \odot \lambda x. \mu_n(h_n, x)$}

An agent may need to calculate, for $h_n \in H_n$,
\[ \nu_n(h_n) \otimes \lambda x. \mu_n(h_n, x) : \mathcal{P}(X \times Y) \] 
and
\vspace{-.5em}
\[ \nu_n(h_n) \odot \lambda x. \mu_n(h_n, x) : \mathcal{P}(Y). \] 
Note that 
\[ \nu_n(h_n) \otimes \lambda x. \mu_n(h_n, x) = (\nu_n \otimes \mu_n)(h_n) \text{ and }
\nu_n(h_n) \odot \lambda x. \mu_n(h_n, x) = (\nu_n \odot \mu_n)(h_n). \]
Also $\nu_n(h_n) \odot \lambda x. \mu_n(h_n, x)$ is the marginal distribution of $\nu_n(h_n) \otimes \lambda x. \mu_n(h_n, x)$ for $Y$.
For any $F_\nu$ and $F_\mu$, $\nu_n(h_n) \otimes \lambda x. \mu_n(h_n, x)$ and $\nu_n(h_n) \odot \lambda x. \mu_n(h_n, x)$ can be approximated as follows.

To explain this, it will be convenient to introduce the concept of a conditional particle family.
A conditional particle (from $X$ to $Y$) is a pair $(x, (y^{(j)})_{j=1}^M)$, 
where $x$ is a particle in $X$ and $(y^{(j)})_{j=1}^M$ is a particle family in $Y$.
A conditional particle family (from $X$ to $Y$) is an indexed family of the form 
$((x^{(i)}, (y^{(i,j)})_{j=1}^M))_{i=1}^N$, where each $(x^{(i)}, (y^{(i,j)})_{j=1}^M)$ is a conditional particle.
So a conditional particle family is an indexed family of conditional particles.

Returning to the computation of 
$\nu_n(h_n) \otimes \lambda x. \mu_n(h_n, x)$ and $\nu_n(h_n) \odot \lambda x. \mu_n(h_n, x)$,
sample a particle family $(x^{(i)}_n)_{i=1}^N$ from $\nu_n(h_n)$.
Then, from each $\mu_n(h_n, x^{(i)}_n)$, sample a particle family $(y^{(i,j)}_n)_{j=1}^M$.
This provides a conditional particle family $((x^{(i)}_n, (y^{(i,j)}_n)_{j=1}^M))_{i=1}^N$.
It follows that, for all $n \in \mathbb{N}$,
\[ \nu_n(h_n) \otimes \lambda x. \mu_n(h_n, x) \approx \frac{1}{NM} \sum_{i=1}^N \sum_{j=1}^M \delta_{(x^{(i)}_n, y^{(i,j)}_n)}. \]
Hence, if $f : X \times Y \rightarrow \mathbb{R}$ is an integrable function, then 
\vspace{-.7em}
\[ \int_{X \times Y} f \; d (\nu_n(h_n) \otimes \lambda x. \mu_n(h_n, x)) \approx \frac{1}{NM} \sum_{i=1}^N \sum_{j=1}^M f(x^{(i)}_n, y^{(i,j)}_n). 
\vspace{-.5em}
\]

In addition, for all $n \in \mathbb{N}$,
\vspace{-.7em}
\[  \nu_n(h_n) \odot \lambda x. \mu_n(h_n, x) \approx \frac{1}{NM} \sum_{i=1}^N \sum_{j=1}^M \delta_{y^{(i,j)}_n}. 
\vspace{-.5em}
\]
Hence, if $g : Y \rightarrow \mathbb{R}$ is an integrable function, then
\vspace{-.7em}
\[ \int_Y g \; d (\nu_n(h_n) \odot \lambda x. \mu_n(h_n, x)) \approx \frac{1}{NM} \sum_{i=1}^N \sum_{j=1}^M g(y^{(i,j}_n). 
\vspace{-.5em}
\]

If $\lambda x. \mu_n(h_n, x)$ is defined everywhere, there are no issues.
If not, then $\lambda x. \mu_n(h_n, x)$ needs to be defined at least on the particle family $(x^{(i)}_n)_{i=1}^N$, a family that can be expected to change over time.
If $F_\nu$ is a particle filter, then this can be achieved by updating with $F_\nu$ first and then making the particle family $(x^{(i)}_n)_{i=1}^N$ known to $F_\mu$ so that it can do its update.

Here is a summary of the interface considerations.
If the observation schema for a nonconditional filter is not known a priori, then the nonconditional filter must communicate with the paired conditional filter at the start of each time step;
if a nonconditional filter is a particle filter and the conditional empirical belief given by the paired conditional filter is not defined everywhere, then the conditional filter must communicate with the nonconditional filter at the start of each time step.

%% file: epidemic.tex
\section{Modelling Epidemic Processes on Contact Networks}
\label{sec:epi}

The discussion now moves to practical applications of the filtering algorithms in Section~\ref{sec:theory}.
We concentrate on the three factored conditional algorithms which are the most novel of the twelve algorithms.
Also we restrict attention to just the belief acquisition aspect of the agent cycle; in particular, we exclude considerations of action selection and application, and thus restrict the above theoretical foundation accordingly.

An important issue for public health is investigating the transmission of infectious diseases amongst the population. 
To study this, researchers typically have used compartmental models that are concerned primarily with population dynamics, that is,
the evolution of the total numbers of healthy and infected individuals~\cite{Anderson:2013, Brauer:2012, Hethcote:2000}.

A compartmental model divides the population into several compartments (or groups), such as susceptible, infectious,  
or recovered, etc.
One of the simplest compartmental models involves two compartments $\sS$ and $\sI$, where $\sS$ represents healthy individuals who are susceptible to the disease and $\sI$ infectious individuals who can infect susceptible ones before their recovery~\cite{Kermack:1927}.
The disease can be transmitted from infectious individuals to susceptible ones at a rate $\beta$ (\ie the contact, or infection, rate), and those infectious 
will recover at a rate $\gamma$ (\ie the recovery rate). 
If it is assumed that the recovery from infection does not result in immunity to the infection, such a model is known as the \emph{Susceptible-Infectious-Susceptible} model, or SIS model for short.

Additional compartments can be introduced to capture more features of a disease, such as the gain of immunity upon recovery (at least for a period of time). One such model is the \emph{Susceptible-Exposed-Infectious-Recovered-Susceptible} model, or SEIRS model for short~\cite{bjornstad-shea-krzywinski-altman, Hethcote:2000}.
Here $\sE$ represents individuals who have been exposed to the disease, but are in the pre-infectious period,
\ie individuals in compartment $\sE$ do not infect susceptible individuals; 
and $\sR$ denotes individuals recovered from infection with immunity to the disease for a period of time.
Extensions to the SEIRS model that include compartments for quarantined, hospitalized, or asymptomatic infectious, and so on, have also been investigated~\cite{Della:2020, Gatto:2020,Tang:2020}. 
However, in this paper, we focus on the SIS and SEIRS epidemic models that are illustrated in Figure~\ref{fig:epi}. 
The various rates mentioned in Figure~\ref{fig:epi} are parameters in the set of differential equations that govern the behaviour of the epidemic models.

\input{fig_epi}

A drawback of compartmental models is the simplifying assumption of a well-mixed population, which means that every individual interacts with (and thus may transmit disease to) every other individual in the population with equal probability~\cite{Nowzari:2016}.
However, most people regularly interact with only a small subset of the population, for example, family members, co-workers and neighbours, and it is more realistic to assume that a person is much more likely to contract a disease from someone amongst their regular contacts rather than a random individual in the population. 
Thus it would be better to have an accurate estimation of the distribution of the disease state of all individuals in the population.
To model this, one needs network models of the transmission of diseases that take into account which individuals come into close enough contact for the disease to be transmitted from one to another.
Therefore, in this paper, epidemic processes are modelled on contact networks~\cite{Newman:2018, Nowzari:2016, Pastor:2015}.

A contact network is an undirected graph with nodes labelled by the compartment of the individual at that node.
It will be helpful to formalize this notion.
Let $G$ be an undirected finite graph.
Thus $G = (V, E)$, where $V$ is the finite set of vertices of the graph and $E$ the set of undirected edges between vertices.

We consider several models of epidemic processes on contact networks.
These models are employed for both simulations and filters, and the model for a simulation and the model for the corresponding filter may be different.
The main choices available are how to model the states and how to model the observations.
Each of these is considered in turn.

\subsection{Modelling States}

States are tuples of labelled nodes.
The two main choices then are to label nodes by compartments or to label nodes by categorical distributions over compartments.

\subsubsection{Nodes Labelled by Compartments}
\label{sec:nodes_compartments}

Let $C$ be the set of possible compartments (of the disease) of a node;
for example, $C = \{\sS, \sI\}$ is the set of compartments for the SIS model and
$C = \{\sS, \sE, \sI, \sR\}$ the set of compartments for the SEIRS model.
A vertex-labelling function is a function $\varphi : V \rightarrow C$.
Thus a pair $(G, \varphi)$ is a labelled undirected graph that models a contact network: the nodes model individuals, 
the edges model (direct) contact between individuals, and the function $\varphi$ models the compartment of individuals.
The graph $G$ is said to be the underlying graph for the contact network.

Depending on the particular application, the underlying graph may or may not vary through time.
In this paper, we assume the vertex set $V$ is fixed, but the edge set $E$ may vary.
What certainly varies through time is the vertex-labelling function $\varphi$ as the disease condition of each individual changes.
Thus we are led to consider the function space $C^V$ of all vertex-labelling functions from $V$ to $C$.
In the context of a fixed set of vertices in an underlying graph, it is $C^V$ that is considered to be the state space and the interest is
in filtering distributions on this space.
In fact, it will be more convenient to work with a variant of  $C^V$.
Suppose that $|V|=L$.
Then $\{ 1, 2, \ldots, L \}$ is isomorphic to $V$ under any convenient indexing of the nodes and thus $C^L$ is isomorphic to $C^V$.
In the context of a fixed set of vertices, the product space $C^L$ is taken to be the state space.
If $s \in C^L$, one can write $s = (s_1, \ldots, s_L)$, where $s_k \in C$, for $k = 1, \ldots, L$. 
Filtering for a contact network means tracking states in this space.

Here are two key ingredients for filtering on contact networks where states are modelled as tuples of compartments.
First is the empirical schema.
This has the form
\[ (\breve{\mu}_n : H_n \rightarrow \mathcal{D}(C^L))_{n \in \mathbb{N}_0}.\]
Each history $h_n \in H_n$ consists just of a sequence of observations up to time $n$.
The corresponding empirical belief is $\breve{\mu}_n(h_n) \in \mathcal{D}(C^L).$
Note that $\mathcal{D}(C^L)$ is the set of densities on the finite set $C^L$ corresponding to categorical distributions.

Now consider a typical family of transition schemas for epidemic processes on contact networks.
For both the SIS and SEIRS epidemic models, the transition schema 
\vspace{-.2em}
\[ (\breve{\tau}_n : C^L \rightarrow \mathcal{D}(C^L))_{n \in \mathbb{N}} \]  
is defined by
\begin{equation}
\label{eq:trans_model}
\breve{\tau}_n(s) = \lambda s'. \prod_{k=1}^L \breve{\tau}_n^{(k)}(s)(s'_k), 
\end{equation}
for all $n \in \mathbb{N}$ and $s \in C^L$.
Here, $\breve{\tau}_n^{(k)} : C^L \rightarrow \mathcal{D}(C)$, for all $n \in \mathbb{N}$ and $k = 1, \ldots, L$.
The SIS and SEIRS models each have their own particular definition for each $\breve{\tau}_n^{(k)}$. 
For the SEIRS model, this is given below; for the SIS model, this is analogous.
In addition to the state, each $\breve{\tau}_n^{(k)}$ needs as input the set of neighbours of each node.
This information can be computed from the underlying graph.
For simplicity, it is supposed that all the necessary neighbourhood information is available to the agent. 
All the other information needed by each $\breve{\tau}_n^{(k)}$ is contained in the state.

In contrast to compartmental models, the parameters for epidemics on contact networks are probabilities.
It is assumed that the infection will only be transmitted between individuals that are neighbours, \ie connected by an edge.
Let $\beta$ be the probability per unit time (for example, a day) that the infection is transmitted from an infectious individual (\ie node) to a connected susceptible node~\cite{Newman:2018, Stilianakis:2010, Teunis:2013, Yang:2006}.
A susceptible node can become infected if at least one node amongst its infectious neighbours transmits the disease to the node.

For $k = 1, \ldots, L$, define $d_k : C^L \rightarrow \mathbb{N}_0$ by $d_k(s)$ 
is the number of infectious neighbours of node $k$ in state $s$.
Thus, if $\mathcal{N}_k$ is the set of (indices of) neighbours of node $k$,  then 
$d_k(s) = \sum_{l \in \mathcal{N}_k} \ind{s_l = \sI}$, where $\ind{\cdot}$ is the indicator function.
Note that $d_k$ can be computed from the state and neighbourhood information in the underlying graph.

Consider the SEIRS epidemic model, so that $C = \{\sS, \sE, \sI, \sR\}$.
Suppose that moving from compartment $\sE$ to compartment $\sI$ is a Bernoulli trial with probability $\sigma$ of success,
moving from compartment $\sI$ to compartment $\sR$ is a Bernoulli trial with probability $\gamma$ of success, and
moving from compartment $\sR$ to compartment $\sS$ is a Bernoulli trial with probability $\rho$ of success.
Thus the average pre-infectious period for individuals exposed to infection is $\sigma^{-1}$ days, the average infectious period is $\gamma^{-1}$ days, and the average duration that recovered individuals lose immunity is $\rho^{-1}$ days, each having a geometric distribution.

For all $n \in \mathbb{N}$ and $k = 1, \ldots, L$, define
\[ \breve{\tau}_n^{(k)} : C^L \rightarrow \mathcal{D}(C) \] 
by 
\vspace{-1em}
\begin{equation}
\label{eq:seirs}
\breve{\tau}_n^{(k)}(s)(c) = \begin{cases}
                                 1 - (1 - \beta)^{d_k(s)} & \text{if $s_k = \sS$ and $c = \sE$} \\
                                 (1 - \beta)^{d_k(s)}     & \text{\text{if $s_k = \sS$ and $c = \sS$}} \\
                                 \sigma                   & \text{if $s_k = \sE$ and $c = \sI$} \\
                                 1 - \sigma               & \text{\text{if $s_k = \sE$ and $c = \sE$}} \\
                                 \gamma                   & \text{if $s_k = \sI$ and $c = \sR$} \\
                                 1 - \gamma               & \text{\text{if $s_k = \sI$ and $c = \sI$}} \\
                                 \rho                     & \text{if $s_k = \sR$ and $c = \sS$} \\
                                 1 - \rho                 & \text{\text{if $s_k = \sR$ and $c = \sR$}} \\
                                 0                        &\text{otherwise,}
                            \end{cases}
\end{equation}
for all $s \in C^L$ and $c \in C$.
Clearly, each $\breve{\tau}_n^{(k)}$ is a conditional density and hence each $\breve{\tau}_n$ is a conditional density.

Note that 
since the parameters $\beta$, $\sigma$, $\gamma$, and $\rho$ are assumed to be fixed,
the definition of the respective transition schema only changes over time according to the number of infectious neighbours given by the function $d_k$.
For simplicity, we ignore the vital dynamics (\ie birth and death) and migrations in the population during the epidemics. 

In Section~\ref{sec:factored_conditional_filters}, conditional filters for estimating the parameters $\beta$, $\sigma$, $\gamma$, and $\rho$ will be studied.
In this case, for all $n \in \mathbb{N}$, the signatures for $\breve{\tau}_n$ and $\breve{\tau}_n^{(k)}$ become
\[ \breve{\tau}_n : [0,1]^4 \times C^L \rightarrow \mathcal{D}(C^L) \]
and 
\[ \breve{\tau}_n^{(k)} : [0,1]^4 \times C^L \rightarrow \mathcal{D}(C), \]
where $[0,1]^4$ is the parameter space,
and $\breve{\tau}_n^{(k)}(s)(c)$ in Eq.~(\ref{eq:seirs}) becomes $\breve{\tau}_n^{(k)}((\beta, \sigma, \gamma, \rho), s)(c)$.

\subsubsection{Nodes Labelled by Categorical Distributions over Compartments}\label{subsubsec:states as categorical distributions}

In an alternative approach, each node is labelled by a categorical distribution over compartments rather than simply by a compartment.

Let $S$ be the positive 3-simplex, that is,
\[ S \triangleq \{ (y_1, y_2, y_3) \in \mathbb{R}^3 \; | \; y_i > 0, \text{ for } i = 1, 2, 3, \text{ and } \sum_{i=1}^3 y_i < 1 \}. 
\]
Then $S$ is isomorphic in the obvious way to the (open) tetrahedron
\[ T \triangleq \{ (y_1, y_2, y_3, y_4) \in \mathbb{R}^4 \; | \; y_i > 0, \text{ for } i = 1, \ldots, 4, \text{ and } \sum_{i=1}^4 y_i = 1 \}.
\]
The argument $y_4 = 1 - \sum_{i=1}^3 y_i$ is called the {\em fill-up} argument and can be arbitrarily chosen amongst the arguments of tuples in $T$.
Thus each element in  $S$ can be identified with a categorical distribution on $C = \{ \mathsf{S}, \mathsf{E}, \mathsf{I},  \mathsf{R} \}$,
where no component is $0$, and we make this identification in the following.
The underlying measure $\upsilon_S$ for  $S$ is Lebesgue measure on $\mathbb{R}^3$.
Then the state space for the epidemic process is defined to be the product space $S^L$.
Thus the nodes for this model can be regarded as being labelled by categorical distributions on $C$.
An advantage of using  $S$ instead of $C$ for the node labels is that a distribution on $C$ quantifies the (usual) uncertainty in the value in $C$.
Note that, if $s \in S^L$, then $s = (s_1, \ldots, s_L)$, where each $s_k \in S$ can be identified with a categorical distribution on $C$. 

The models based on  $S$ employ Dirichlet distributions.
Let \[ \{ (y_1, \ldots, y_n) \in \mathbb{R}^n \; | \; y_i > 0, \text{ for } i = 1, \ldots, n, \text{ and } \sum_{i=1}^n y_i < 1 \} \]
be the positive $n$-simplex $S_n$.
Then a Dirichlet distribution $\mathit{Dir}(\alpha_1, \ldots, \alpha_{n+1}) : \mathcal{D}(S_n)$ has the form
\[ \mathit{Dir}(\alpha_1, \ldots, \alpha_{n+1}) \triangleq 
\lambda y. \frac{\Gamma(\sum_{i=1}^{n+1} \alpha_i)}{\prod_{i=1}^{n+1} \Gamma(\alpha_i)} (1 - \sum_{i=1}^n y_i)^{\alpha_{n+1} -1}  \prod_{i=1}^n y_i^{\alpha_i - 1}, \]
where $y = (y_1, \ldots, y_n) \in S_n$, and $\alpha_i > 0$, for $i = 1, \ldots, n+1$.
Of specific interest are Dirichlet distributions of the form $\mathit{Dir}(\alpha_1, \alpha_2, \alpha_3, \alpha_4) :  \mathcal{D}(S)$.

Going back to the epidemic model, an empirical schema has the form
\[ (\breve{\mu}_n : H_n \rightarrow \mathcal{D}(S^L))_{n \in \mathbb{N}_0}. \]
At any time $n$, the (approximate) state distribution has the form
$\bigotimes_{k=1}^L \reallywidehat{\breve{\mu}_n^{(k)}(h_n^{(k)})} : \mathcal{D}(S^L)$, where each 
$\reallywidehat{\breve{\mu}_n^{(k)}(h_n^{(k)})} :  \mathcal{D}(S)$ is assumed to be a Dirichlet distribution.

A transition schema 
$(\breve{\tau}_n : S^L \rightarrow \mathcal{D}(S^L))_{n \in \mathbb{N}}$ 
can be defined by
\[ \breve{\tau}_n(s) = \lambda s'. \prod_{k=1}^L \breve{\tau}_n^{(k)}(s)(s'_k), \]
for all $n \in \mathbb{N}$ and $s \in S^L$,
where, for $k = 1, \ldots, L$, 
\[ \breve{\tau}_n^{(k)} : S^L \rightarrow  \mathcal{D}(S) \]
is defined by 
\[ \breve{\tau}_n^{(k)}(s) = \mathit{Dir}(K\alpha_1, K\alpha_2, K\alpha_3, K\alpha_4), \]
where 
\vspace{-1em}
\begin{gather}
\label{eq:T_seirs_trans}
\begin{aligned}
\alpha_1 & \triangleq \rho s_{k,4} + \left( (1 -\beta)^{\sum_{l \in \mathcal{N}_k} s_{l,3}} \right) s_{k,1}, \\
\alpha_2 & \triangleq \left( 1 - (1-\beta)^{\sum_{l \in \mathcal{N}_k} s_{l,3}} \right) s_{k,1}  + (1 - \sigma) s_{k,2}, \\
\alpha_3 & \triangleq \sigma s_{k,2}  + (1 - \gamma) s_{k,3}, \\ 
\alpha_4 & \triangleq \gamma s_{k,3}  + (1 - \rho) s_{k,4},
\end{aligned}
\end{gather}
for all $s \in S^L$, and $s_{k,4} \triangleq1 -  \sum_{i=1}^3 s_{k,i}$.
Here, $s_{k,i}$ is the $i$th component of $s_k \in S$.
$K > 0$ is a constant that controls the variance in the codomain of the transition schema.
Eq.~(\ref{eq:T_seirs_trans}) is the analogue for $S^L$ of the transition schema for $C^L$ in Eq.~(\ref{eq:seirs}).

\subsubsection{Subpopulation Contact Networks}
We now consider a more general class of transition schemas on what we call subpopulation contact networks, which are contact networks where each node represents a subpopulation (instead of an individual) and each edge represents probabilistic interactions between two subpopulations.
A simple model that formalizes this idea is now described; the model can be extended in various ways as needed.

A subpopulation contact network is an undirected graph $G = (V,E)$, where each node $k \in V$ is labelled by a tuple $(M_k, s_k)$, with $M_k$ denoting the number of individuals in the subpopulation located at node $k$, and $s_k \in S$ denoting the node state in terms of the percentages of individuals in the subpopulation belonging to each of the four compartments $\sS, \sE, \sI, \sR$.
A subpopulation contact network represents a distribution of possible individual-level contact networks 
through the following sampling procedure: 
\begin{enumerate}[label=(\arabic*),itemsep=-5pt,topsep=2pt]
  \item Each node $k$ in the subpopulation contact network is turned into a subgraph by sampling from the Erd{\"o}s-R{\'e}nyi random graph model $ERG(M_k,\kappa_1)$, where $\kappa_1$ is the probability that each pair of individuals in node $k$ are connected. Each individual 
  is then labelled by sampling a compartment from $s_k$. 
 \item Then for each edge $e = (k,l)$ connecting $k$ and $l$ in the subpopulation contact network, we connect every pair of individuals 
 $(j_k, j_l)$
 with probability $\kappa_2$,
 where $j_k$ is an individual in the subgraph for node $k$, and $j_l$ an individual in the subgraph for node $l$.
\end{enumerate}
Note that this sampling procedure is not actually carried out in the implementation; it just provides motivation for the following definition of the transition schema.

The state of the subpopulation contact network is given by $(s_1,\ldots, s_L) \in S^L$, where $L = |V|$,  at any one time, and the state can change according to a transition schema obtained by appropriately combining the results of running 
Eq.~\eqref{eq:seirs} (together with Eq.~\ref{eq:trans_model})
on the underlying distribution of individual-level contact networks.
This motivates the transition schema $( \breve{\tau}_n  : S^L \rightarrow \mathcal{D}(S^L))_{n \in \mathbb{N}}$ for the subpopulation contact network $G = (V,E)$ defined by
\vspace{-.7em}
\begin{equation}
\label{eq:subpopulation:T_seirs_trans}
\breve{\tau}_n(s) = \lambda s'. \prod_{k=1}^L \breve{\tau}_n^{(k)}(s)(s'_k),
\vspace{-.7em}
\end{equation}
for all $n \in \mathbb{N}$ and $s \in S^L$,
where, for $k = 1, \ldots, L$,
\vspace{-.3em}
\[ \breve{\tau}_n^{(k)} : S^L \rightarrow  \mathcal{D}(S) 
\vspace{-.5em}
\]
is defined by 
\[ \breve{\tau}_n^{(k)}(s) = \mathit{Dir}(K\alpha_1, K\alpha_2, K\alpha_3, K\alpha_4), 
\]
where
\begin{equation}
\label{eq:subpopulation:T_seirs_trans2}
\begin{aligned}
& \alpha_1  \triangleq \rho s_{k,4} + 
       \left( (1 - \beta)^{(i_k + i_{\mathcal{N}_k})} \right) s_{k,1} & \hspace{3em}
& \alpha_2 \triangleq \left( 1- (1 - \beta)^{(i_k + i_{\mathcal{N}_k})} \right) s_{k,1} + (1 - \sigma) s_{k,2} \\
& \alpha_3 \triangleq \sigma s_{k,2}  + (1 - \gamma) s_{k,3} & 
& \alpha_4 \triangleq \gamma s_{k,3}  + (1 - \rho) s_{k,4} \\
& i_k  \triangleq (M_k -1) \kappa_1 s_{k,3} & 
& i_{\mathcal{N}_k} \triangleq \sum_{l \in \mathcal{N}_k} M_l \kappa_2 s_{l,3},
\end{aligned}
\end{equation}
for all $s \in S^L$, and $s_{k,4} \triangleq1 -  \sum_{i=1}^3 s_{k,i}$.
To see how Eq.~\eqref{eq:subpopulation:T_seirs_trans2} comes about, consider a node $k$ with an expected number $M_k s_{k,1}$ of susceptible individuals.
Each of these susceptible individuals is connected, on average, to $(M_k -1)\kappa_1 s_{k,3}$ infected individuals in node $k$ and $\sum_{l \in \mathcal{N}_k} M_l \kappa_2 s_{l,3}$ infected individuals in node $k$'s neighbouring nodes.
Thus, by Eq.~(\ref{eq:seirs}), we can expect to have 
\begin{equation}
\label{eq:subpopulation:susceptible}
M_k s_{k,4} \rho + 
    M_k s_{k,1} (1 - \beta)^{( (M_k -1) \kappa_1 s_{k,3} + \sum_{l \in \mathcal{N}_k} M_l \kappa_2 s_{l,3})}
\end{equation}
susceptible individuals in node $k$ after one time step. 
Dividing Eq.~\eqref{eq:subpopulation:susceptible} by $M_k$ gives us the formula for $\alpha_1$ in Eq.~(\ref{eq:subpopulation:T_seirs_trans2}).
The formulas for $\alpha_2$, $\alpha_3$, and $\alpha_4$ can be obtained using a similar reasoning.
Note that Eq.~\eqref{eq:T_seirs_trans} is a special case of Eq.~\eqref{eq:subpopulation:T_seirs_trans2} when $M_k = 1$, 
$M_l = 1$, for all $l \in \mathcal{N}_k$, and $\kappa_2 = 1$.

\subsection{Modelling Observations}

Next the observation schemas are given for different kinds of observations.
Observations are made at the node level.
Thus an observation space $O$ is introduced for observations at that level.
The observation space for the state level is $O^L$.

\subsubsection{Observations as Indicators of Presence or Absence of Disease}

Consider the SEIRS epidemic model, so that $C = \{\sS, \sE, \sI, \sR\}$.
The observations resemble the testing for an infectious disease with positive ($\Pos$) or negative ($\Neg$) outcomes. 
Let $O = \{\Pos, \Neg, \Unk\}$,
where the question mark ($\Unk$) indicates the corresponding individual does not have a test result.
Suppose that nodes are labelled by compartments.
Then, for all $n \in \mathbb{N}$ and $k = 1,\dots, L$,
\[ \breve{\xi}_n^{(k)} : C \rightarrow \mathcal{D}(O) 
\]
is defined by
\begin{equation}
\label{eq:seirs:obs}
\begin{aligned}
    & \breve{\xi}_n^{(k)}(\sS)(\Pos) = \alpha_\sS \lambda_{FP},
    & \hspace{.1em}
    & \breve{\xi}_n^{(k)}(\sE)(\Pos) = \alpha_\sE (1 - \lambda_{FN}),
    & \hspace{.1em}
    & \breve{\xi}_n^{(k)}(\sI)(\Pos) = \alpha_\sI (1 - \lambda_{FN}),
    & \hspace{.1em}
    & \breve{\xi}_n^{(k)}(\sR)(\Pos) = \alpha_\sR \lambda_{FP}, \\
    & \breve{\xi}_n^{(k)}(\sS)(\Neg) = \alpha_\sS (1 - \lambda_{FP}),
    & {}
    & \breve{\xi}_n^{(k)}(\sE)(\Neg) = \alpha_\sE \lambda_{FN},
    & {}
    & \breve{\xi}_n^{(k)}(\sI)(\Neg) = \alpha_\sI \lambda_{FN},
    & {}
    & \breve{\xi}_n^{(k)}(\sR)(\Neg) = \alpha_\sR (1 - \lambda_{FP}), \\
    & \breve{\xi}_n^{(k)}(\sS)(\Unk) = 1 - \alpha_\sS,
    & {}
    & \breve{\xi}_n^{(k)}(\sE)(\Unk) = 1 - \alpha_\sE,
    & {}
    & \breve{\xi}_n^{(k)}(\sI)(\Unk) = 1 - \alpha_\sI,
    & {}
    & \breve{\xi}_n^{(k)}(\sR)(\Unk) = 1 - \alpha_\sR.
\end{aligned}
\end{equation}

Here, $\alpha_\sS$, $\alpha_\sE$, $\alpha_\sI$, and $\alpha_\sR$ is the fraction of susceptible, exposed, infectious, and recovered individuals respectively that are tested on average at each time step.
Also $\lambda_{FP}$ is the false positive rate of the testing method employed, and $\lambda_{FN}$ the false negative rate.
Clearly, each $\breve{\xi}_n^{(k)}$ is a conditional density.

Alternatively, suppose that nodes are labelled by categorical distributions over compartments.
Then, for all $n \in \mathbb{N}$ and $k = 1,\dots, L$,
\vspace{-.3em}
\[ \breve{\xi}_n^{(k)} : S \rightarrow \mathcal{D}(O) 
\vspace{-.5em}
\]
is defined by
\begin{equation*}
\begin{aligned}
    & \breve{\xi}_n^{(k)}(y_1, y_2, y_3)(\Pos) = \alpha_\sS \lambda_{FP} y_1 + \alpha_\sE (1 - \lambda_{FN}) y_2 +
              \alpha_\sI (1 - \lambda_{FN}) y_3 + \alpha_\sR \lambda_{FP} y_4 \\
    & \breve{\xi}_n^{(k)}(y_1, y_2, y_3)(\Neg) = \alpha_\sS (1 - \lambda_{FP}) y_1 + \alpha_\sE \lambda_{FN} y_2 +
              \alpha_\sI \lambda_{FN} y_3 + \alpha_\sR (1 - \lambda_{FP}) y_4 \\
    & \breve{\xi}_n^{(k)}(y_1, y_2, y_3)(\Unk) = (1 - \alpha_\sS) y_1 + (1 - \alpha_\sE) y_2 + 
              (1 - \alpha_\sI) y_3 + (1 - \alpha_\sR) y_4,
\end{aligned}
\end{equation*}
for all 
$(y_1, y_2, y_3) \in S$, where $y_4\triangleq 1 - \sum_{i=1}^3 y_i$.

\subsubsection{Observations as Counts For Each Compartment}

Fix $m \in \mathbb{N}$ and let
$O = \{ (o_1, o_2, o_3, o_4) \in \mathbb{N}_0^4 \; | \; \sum_{j=1}^4 o_j = m \}$.
Then, for all $n \in \mathbb{N}$ and $k = 1,\dots, L$,
\vspace{-.3em}
\[ \breve{\xi}_n^{(k)} : S \rightarrow \mathcal{D}(O) 
\vspace{-.3em}
\]
is defined by
\begin{equation}
\label{obs_update_multinomial}
\breve{\xi}_n^{(k)}(s_1, s_2, s_3) = \mathit{Mult}(m, (s_1, s_2, s_3, s_4)),  
\end{equation}
for all $s = (s_1, s_2, s_3) \in S$, where $s_4 \triangleq1 - \sum_{i=1}^3 s_i$.
Here $\mathit{Mult}(m, (s_1, s_2, s_3, s_4))$ is the multinomial distribution for which the total number of draws is $m$ and $s_1$, $s_2$, $s_3$, and $s_4$ are the parameters of a categorical distribution.
So, for example, in the simulation, if the current state is $(s_1, s_2, s_3)$, an observation is generated by sampling from 
$\mathit{Mult}(m,(s_1, s_2, s_3, s_4))$.
Also there is a parameter $\alpha$ which gives the probability
a node is observed at a time step.

The multinomial observation schema in Eq.~\eqref{obs_update_multinomial} has a natural interpretation.
Imagine that a node represents a subpopulation of the total population of individuals.
Then the state labelling a node captures the population properties of that node: if the state is $s = (s_1, s_2, s_3)$, then $s_1$ is the proportion of individuals in the subpopulation that are susceptible, and so on.
An observation then gives the numbers of susceptible, exposed, infectious, and recovered individuals amongst a sample of $m$ (not necessarily distinct) individuals in the subpopulation.
This interpretation works even if a node represents a single individual: an observation then gives the results of one or more tests on that individual.
Of course, different test applications (of even the same kind of test) to a single individual can give different results.

\subsection{Experimental Setup}

The experiments in Section~\ref{sec:factored_conditional_filters} all concern SEIRS epidemics.
They use a variety of state spaces, observation spaces, transition schemas, and observation schemas to illustrate the various filtering algorithms. 
For each application, the simulation and the filter use the same state spaces and observation spaces: the observation spaces for the simulation and filter have to be the same, of course, but the respective state spaces {\em could} be different which just reflects the uncertainty in modelling real-world applications.

Each state space is either $C^L$, where $C = \{\sS, \sE, \sI, \sR\}$, with products of categorical distributions as state distributions, or $S^L$, where $S$ is the positive 3-simplex, with products of Dirichlet distributions as state distributions. 
Each observation space is either $\{\Pos, \Neg, \Unk\}$
or $\{ (o_1, o_2, o_3, o_4) \in \mathbb{N}_0^4 \; | \; \sum_{j=1}^4 o_j = m \}$.

For each application, the observation schemas for the simulation and the filter are the same.
For the applications of Algorithms 10 and 11, the transition schemas for the simulation and the filter are the same.
For the application of Algorithm 12, the filter does not use the transition schema of the simulation; instead, it uses a direct transition update that approximates the effect of applying the simulation's transition schema.
The reason for this is explained
in Section~\ref{sec:fcvf}.
The transition and observation schemas used are all defined earlier in this section.

For Algorithms 10, 11, and 12, the parameters $\beta$, $\sigma$, $\gamma$ and $\rho$ of the transition schema need to be estimated.
For these algorithms, the additional task of estimating the parameters makes the applications rather more difficult.
For all the experiments concerning conditional filters that are reported here, it was convenient to use particle filters to estimate the parameter distributions.
It would be interesting to explore other possibilities, such as Kalman filters.

For each experiment, we have concentrated on presenting a setup for which the corresponding filtering algorithm works satisfactorily rather than trying to optimize its performance;
the experimental results could all be improved by adjusting the relevant parameter values.
However, for Algorithm 12 especially, finding a setup for which the algorithm did work satisfactorily was not straightforward.
Sometimes apparently minor changes in the setup caused the algorithm to perform poorly.

Table~\ref{tab:dataset} gives the set of contact networks for which experiments were conducted.
Note that {\tt AS-733} and {\tt AS-122} are {\em dynamic} contact networks.
For reasons of space, only a fraction of the set of experiments that were carried out are reported in this paper. 
The complete set of experimental results can be found in the GitHub repository.

\begin{table*}[tb]
    \begin{tabular}{@{}lrrl@{}}
    \toprule
    {\bf Dataset} & {\bf Nodes} & {\bf Edges} & {\bf Description} \\
    \midrule
    {\tt OpenFlights}{\small\,\cite{Opsahl:2011}} & $2,905$ & $15,645$ & Airport network \\
    {\tt Facebook}{\small\,\cite{Rozemberczki:2021}} 
    & $22,470$    & $170,823$   & Social network of Facebook sites \\
    {\tt Email}{\small\,\cite{email-33k:2004}}       & $33,696$    & $180,811$   & Network of email communications \\
    {\tt GitHub}{\small\,\cite{Rozemberczki:2021}}   & $37,700$    & $289,003$   & Social network of developers on GitHub \\
    {\tt Gowalla}{\small\,\cite{Cho:2011}}           & $196,591$   & $950,327$   & Friendship network of Gowalla users \\
    {\tt Youtube}{\small\,\cite{Mislove:2007}}       & $1,134,890$ & $2,987,624$ & Friendship network of Youtube users \\
    {\tt AS-733}{\small\,\cite{Leskovec:2005}}       & $6,474$ 
    & $248\negthinspace-\negthinspace13,895$    & 733 \emph{dynamic} communication networks of autonomous systems \\
    {\tt AS-122}{\small\,\cite{Leskovec:2005}}       & $26,475$\hspace{-.2em} 
    & $18,203\negthinspace-\negthinspace53,601$ & 122 \emph{dynamic} communication networks of autonomous systems \\
    \bottomrule
    \end{tabular}
\caption{Summary of network datasets used in the experiments}
\label{tab:dataset}
\end{table*}

\begin{figure*}[htbp]
    \centering
    \setlength{\tabcolsep}{.05in}
    \begin{tabular}{cccc}
    \multicolumn{4}{r}{
    \includegraphics[width=0.71\linewidth]{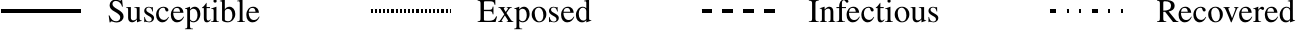} {\hspace{.2em}}
    } \\
    \includegraphics[height=1.2in]{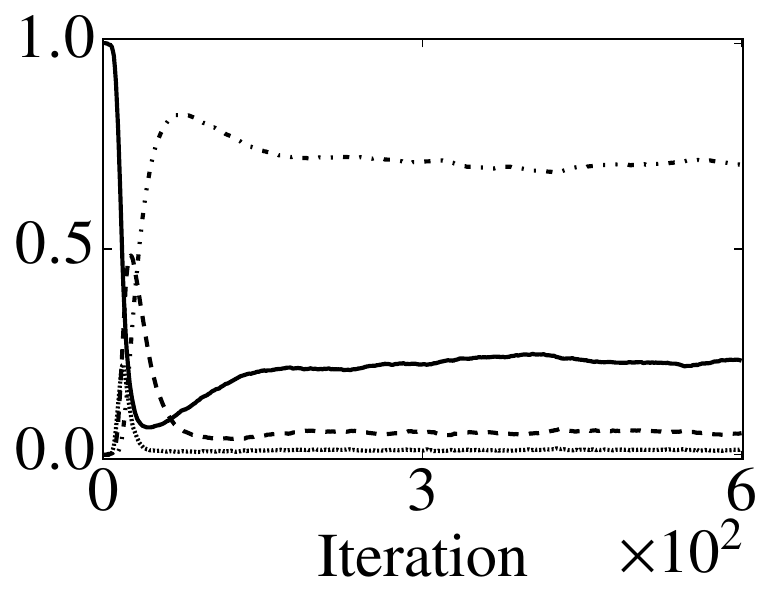} &
    \includegraphics[height=1.2in]{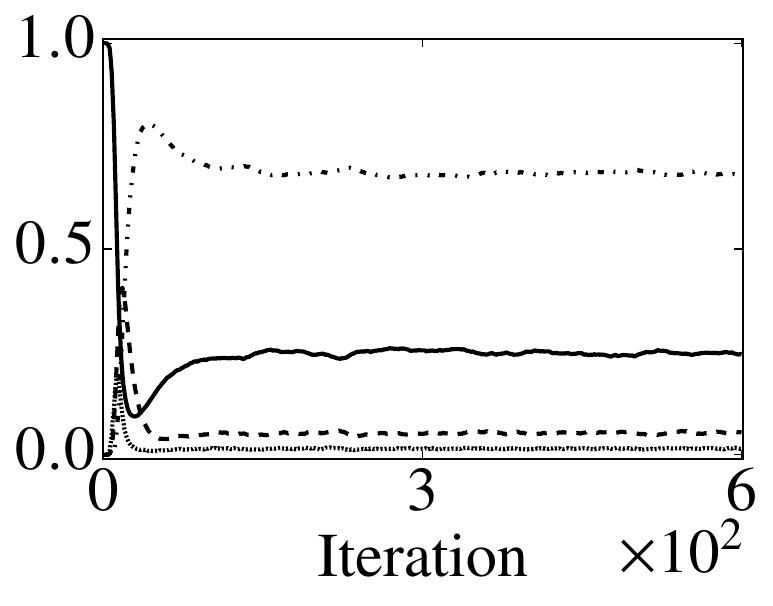} &
    \includegraphics[height=1.2in]{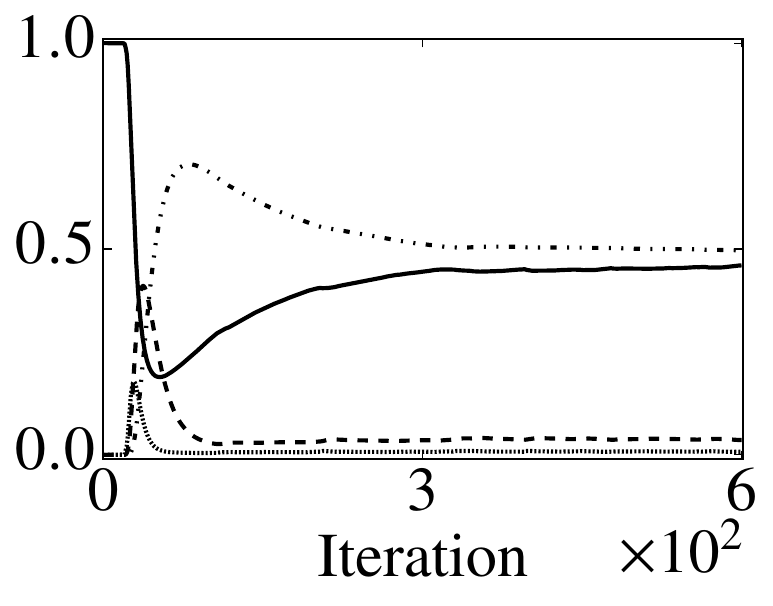} &
    \includegraphics[height=1.2in]{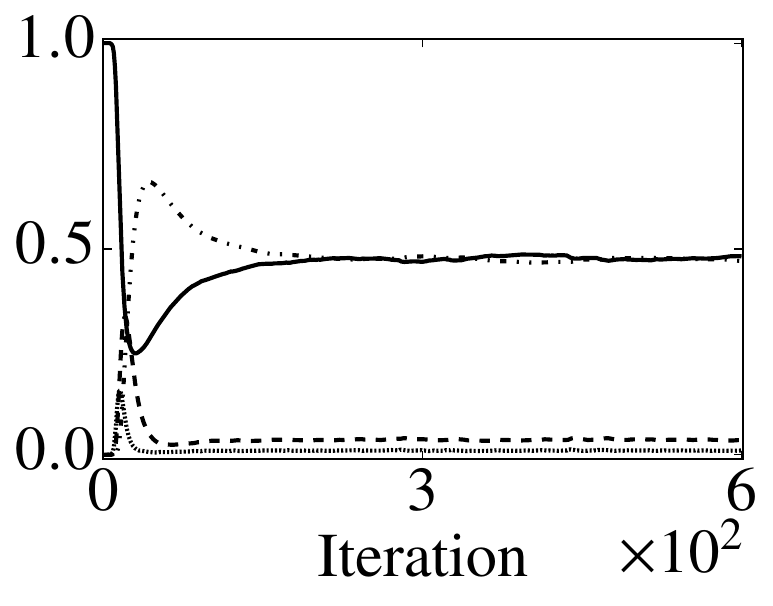} \\
    {\text{(a)}} & {(b)} & {(c)} & {(d)}
    \end{tabular}
    \caption{
        Population properties of one run of the simulation on two contact networks, {\tt Facebook} and {\tt Youtube}.
        (a) and (b): {\tt Facebook}. (c) and (d): {\tt Youtube}.
        (a) and (c) use parameters $\beta=0.2, \, \sigma=1/3, \, \gamma=1/14, \, \rho=1/180$;
        (b) and (d) use parameters $\beta=0.27, \, \sigma=1/2, \, \gamma=1/7, \, \rho=1/90$.
    }
    \label{fig:population_properties}
\end{figure*}

\subsection{Population Properties of Epidemic Processes}
\label{sec:population_properties}

This subsection provides some insight into the nature of the epidemic processes defined above.
For this purpose, population properties of epidemic processes on two contact networks are used.
For each of the compartments, $\sS$, $\sE$, $\sI$, and $\sR$, at each time step, one considers the percentage of nodes in the contact network that are labelled by a particular compartment.

Consider one run of the simulation for which the state space is $C^L$ and the transition schema is given by Eq.~\eqref{eq:seirs}.
Figure~\ref{fig:population_properties} gives the four population properties at each time step for two contact networks from Table~\ref{tab:dataset}, for each of two different sets of epidemic process parameters.
From Figure~\ref{fig:population_properties}, the overall progress, but not the detailed changes at each node, of the epidemic can be comprehended.
At time 0, the percentage of susceptible nodes is very nearly one, but 
drops rapidly as susceptible nodes become exposed and then infectious.
At the same time, the percentage of recovered nodes rises rapidly.
During the first 50 to 100 steps, there is considerable change in the state.
After that, the epidemic enters a quasi-steady state period during which the population properties change only slowly. 
However, in the quasi-steady state period, there are still plenty of changes of the compartment labels at the node level, 
but in such a way that the population properties are kept rather stable.

From the point of view of filtering, a reasonable conjecture is that the rapid progress of the epidemic process in the first 50 to 100 steps will make it difficult to track the state distribution at the beginning of the epidemic.
Furthermore, the slower pace of change after this point suggests that state tracking will then become easier.
This conjecture is borne out by the experimental results that follow.
The typical pattern in the applications is that the state errors of the filter are higher near the start of the epidemic, rising to a peak at around 30 time steps and then dropping to a lower, nearly constant, state error after about 100 steps.
(The exact behaviour depends on the application, of course.)
In applications where parameters have to be learned, there is a similar behaviour for the parameter errors.
This is related to the fact that the accuracy of parameter estimation depends significantly on the accuracy of state tracking.
It is necessary to be able to track the state accurately when the epidemic process parameters are known {\em before} attempting to simultaneously estimate the parameters and track the state.

%% file: fig_epi.tex
\tikzstyle{S}=[shape=rectangle,draw=black]
\tikzstyle{E}=[shape=rectangle,draw=black]
\tikzstyle{I}=[shape=rectangle,draw=black]
\tikzstyle{R}=[shape=rectangle,draw=black]

\begin{figure*}[htbp]
    \centering
    \begin{tikzpicture}[baseline=(s1.base)]
        \node[S] (s1) at (0,0) {$\sS$};
        \node[I] (i1) at (3,0) {$\sI$};
        
        \node[S] (s2) at (6,0) {$\sS$};
        \node[E] (e2) at (8,0) {$\sE$};
        \node[I] (i2) at (10,0) {$\sI$};
        \node[R] (r2) at (12,0) {$\sR$};
        
        \draw [->,thick] (s1) to (i1) node[draw=none] at (1.5,-.25) {$\beta$};
        \draw [->,thick,bend right] (i1) to [looseness=.8] (s1) node[draw=none] at (1.5, .65) {$\gamma$};

        \draw [->,thick] (s2) to (e2) node[draw=none] at (7,-.25) {$\beta$};
        \draw [->,thick] (e2) to (i2) node[draw=none] at (9,-.25) {$\sigma$};
        \draw [->,thick] (i2) to (r2) node[draw=none] at (11,-.25) {$\gamma$};
        \draw [->,thick,bend right] (r2) to [looseness=.5] (s2) node[draw=none] at (9, .75) {$\rho$};
    \end{tikzpicture}
    \caption{Compartmental models for epidemics. 
             Left: SIS model with contact rate $\beta$ and recovery rate $\gamma$; 
             Right: SEIRS model with contact rate $\beta$, latency rate $\sigma$, recovery rate $\gamma$, and loss of immunity rate $\rho$.}
    \label{fig:epi}
\end{figure*}

%% file: factored_cpf.tex
\section{Factored Conditional Filters for Epidemic Processes}
\label{sec:factored_conditional_filters}

Consider an epidemic process spreading on a contact network, as described in Section~\ref{sec:epi}.
Knowledge of the states and parameters of the epidemic process is crucial to understand 
how an infectious disease spreads in the population, which may enable effective containment measures. 
In this section, we consider the problem of tracking the states and estimating the parameters of epidemic processes 
on contact networks.

For this kind of application, {\em factored} filters are appropriate since epidemic processes satisfy the requirements discussed in Section~\ref{sec:theory:factored} which ensure that the factored filter gives a good approximation of the state distribution even for very large contact networks.
The key characteristic of epidemics that can be exploited by the factored filter is the sparseness of contact networks.
People become infected in an epidemic by the infectious people that they come into sufficiently close contact with.
But whether people live in very large cities or small towns, in most circumstances, they have similar-sized sets of contacts that are usually quite small.
This sparseness of contact networks translates into transition schemas that only depend on the small number of neighbours of a node;
hence the resulting state distributions are likely to be approximately factorizable into independent factors. 

This sparseness of contact networks is reflected in the datasets of Table~\ref{tab:dataset} that we use.
For example, the average node degree for the {\tt Youtube} network is about 6.  
Thus the factored algorithm works well on this network, even though it has over 1M nodes. 
Similarly, the {\tt Facebook} and {\tt GitHub} networks each have average node degree about 15.
This sparseness property is also likely to hold more generally in geophysical applications.

We empirically evaluate factored filtering given by each of the Algorithms~\ref{algo:fcf}, \ref{algo:fcpf},  and \ref{algo:fcvf} for tracking epidemics transmitted through the SEIRS model on (mostly) large contact networks.
We show how factored filters can exploit structure to accurately approximate state distributions for high-dimensional state spaces.
Factored filtering for the state space is fully factored, where each cluster consists of a single node.
We also estimate parameters in each case.
We consider a factored conditional filter, a factored conditional particle filter, and a factored conditional variational filter in turn.

\subsection{Factored Conditional Filter}
\label{sec:fcf}

We now evaluate factored conditional filtering in the setting of epidemics spreading on contact networks, in particular, we estimate the parameters in the SEIRS transition schema (that is, $\beta, \sigma, \gamma$, and $\rho$) and jointly track the states as the epidemics evolve over time. 

For this application, the state space for the simulation is $C^L$, where $C = \{\sS, \sE, \sI, \sR\}$,
the observation space at the node level is $O = \{\Pos, \Neg, \Unk\}$,
the transition schema is given by Eq.~\eqref{eq:seirs}, 
and the observation schema at each node is given by Eq.~\eqref{eq:seirs:obs}.
Patient zero, having compartment $\sE$, that is, exposed, is chosen uniformly at random to start the simulation.
All other nodes initially have compartment $\sS$, that is, they are susceptible.

Epidemics evolve following the transitions given by Eq.~\eqref{eq:seirs} for 600 time steps 
(or, for a dynamic contact network, the number of underlying graphs available)
and generate observations according to Eq.~\eqref{eq:seirs:obs} at each time step.
In particular, we consider a testing method (for the observation schema) which allows large-scale testing (about $20\%$ of the susceptible population, that is, $\alpha_\sS = 0.2$) but may not be particularly accurate.
The setup also assumes that most (about $90\%$, that is, $\alpha_\sI = 0.9$) infectious individuals will be tested (\eg due to symptoms) with good contact tracing that identifies the majority (about $70\%$, that is, $\alpha_\sE = 0.7$) of latent infected cases, and that the testing results can be received within $24$ hours.
We also assume a small proportion (about $5\%$, that is, $\alpha_\sR = 0.05$) of the recovered population will be tested due to the risk of reinfection. 
In summary, the following values are used throughout the paper for Eq.~\eqref{eq:seirs:obs}:
\begin{equation}
\label{eq:experiment:obs_setup}
\alpha_\sS = 0.2, \ \ \alpha_\sE = 0.7, \ \ \ \alpha_\sI = 0.9, \ \ \ \alpha_\sR = 0.05.
\end{equation}

Two set of parameters corresponding to different infectious diseases are considered.
Firstly, we consider an infection with transmission probability $\beta = 0.2$ (that is, transmission rate per day), the average pre-infectious period $\sigma^{-1} = 3$ days, the average infectious period $\gamma^{-1} = 14$ days, and the average period for immunity loss $\rho^{-1} = 180$ days, that is, an infectious disease similar to COVID-19~\cite{Cohen:2020, Della:2020, Gatto:2020, Lumley:2021}.
Also we use a testing method with $\lambda_{FP} = \lambda_{FN} = 0.1$ (that is, false positive rate = false negative rate = $10\%$), 
which is similar to some 
rapid point-of-care testing for COVID-19~\cite{Dinnes:2021}.
In summary, 
\begin{equation}
\label{eq:experiment:setup1}
    \beta = 0.2, \ \ \ \sigma = \frac{1}{3}, \ \ \ \gamma = \frac{1}{14}, \ \ \ \rho = \frac{1}{180}, \ \ \ \lambda_{FP} = \lambda_{FN} = 0.1.
\end{equation}
Note that the average period of immunity loss is significantly longer than the average pre-infectious and infectious periods, that is, $\rho$ is significantly smaller than $\sigma$ and $\gamma$.

Secondly, we consider an infection with transmission probability $\beta = 0.27$, the average pre-infectious period $\sigma^{-1} = 2$ days, the average infectious period $\gamma^{-1} = 7$ days, and the average period for immunity loss $\rho^{-1} = 90$ days, which imitate an infectious disease like one caused by an influenza A virus before effective vaccines are widely administered~\cite{Camacho:2011, cdc:flu, Slifka:1996, Yang:2009}.
Also we use a testing method with $10\%$ false positive rate ($\lambda_{FP} = 0.1$) and $30\%$ false negative rate ($\lambda_{FN} = 0.3)$, which is comparable to many rapid influenza diagnostic tests that can be used in a physician's office~\cite{cdc:flu_testing}, and therefore widely random testing would be practical.
Thus
\begin{equation}
\label{eq:experiment:setup2}
    \beta = 0.27, \ \ \ \sigma = \frac{1}{2}, \ \ \ \gamma = \frac{1}{7}, \ \ \ \rho = \frac{1}{90}, \ \ \ \lambda_{FP} = 0.1, \ \ \ \lambda_{FN} = 0.3.
\end{equation}

Now we consider filtering for this application.
The pairing of Algorithm~\ref{algo:fcf} (fully factored) and Algorithm~\ref{algo:spf} (using observation schema synthesis and a jittering transition schema) is employed.
The factored conditional filter is used for tracking states and the standard particle filter is used for estimating parameters.
The state space $Y$ for the filter is $C^L$, where $C = \{\sS, \sE, \sI, \sR\}$ and the observation space at the node level is $O = \{\Pos, \Neg, \Unk\}$.
The transition schema for $Y$ is given by Eq.~\eqref{eq:seirs} 
(with an extra argument for the parameter space $X = [0,1]^4$,
see the discussion about the signature for the transition schema at the end of  Section~\ref{sec:nodes_compartments}.)
and the observation schema at each node is given by Eq.~\eqref{eq:seirs:obs}.
The transition schema for the parameter space is a jittering transition schema given in Eq.~\eqref{eq:jittering_transition_model}. 
The observation schema for the parameter space can be computed using observation schema synthesis as described in 
Section~\ref{sec:obs_synthesis}.

The initial state distribution for the filter is as follows.
Let $n_0$ be
patient zero in the simulation.
Also let $(p_1, p_2, p_3, p_4)$ denote the tuple of probabilities for the compartments, in the order 
$\sS, \sE, \sI, \sR$, for a node.
Then the initial state distribution for the filter is
\begin{itemize}[itemsep=.01em,topsep=.3em]
\item $(0.29, 0.4, 0.3, 0.01)$ for node $n_0$,
\item $(0.49, 0.3, 0.2, 0.01)$ for the neighbours of node $n_0$ (that is, nodes that are 1 step away from $n_0$),
\item $(0.69, 0.2, 0.1, 0.01)$ for nodes that are 2 steps away from node $n_0$, and
\item $(0.97, 0.01, 0.01, 0.01)$ for nodes that are 3 steps or more away from node $n_0$.
\end{itemize}

The details for the parameter estimation are as follows.
A parameter particle filter with 300 particles is employed, so that $N = 300$.
The parameter particle family is initialized by sampling from uniform distributions as follows:
\begin{equation}
    \label{eq:param:init}
    x^{(i)}_{0,1} \sim \UCal(0, 0.8), \ \ \ x^{(i)}_{0,2} \sim \UCal(0, 0.8), \ \ \ x^{(i)}_{0,3} \sim \UCal(0, 0.8), \ \ \ x^{(i)}_{0,4} \sim \UCal(0, 0.1),
\end{equation}
where $i = 1,\dots, N$.
An adaptive Gaussian jittering kernel is employed in Algorithm~\ref{algo:spf} with covariance matrix
\begin{equation}
    \label{eq:cov:seirs}
    \Sigma_n = \max(a r^n, b) \diag(1, 1, 1, 0.09), \ n \in \N,
\end{equation}
where $a = 10^{-4}, \, b = 9 \times 10^{-6}, \, r = 0.996$, and $\diag(\cdot)$ is a diagonal matrix with the given diagonal entries.
Parameter estimates are given by Eq.~\eqref{eq:estimate_parameter} in Appendix D.

Recall that, for this application, the transition schema 
$\breve{\tau}_n^{(k)} : [0,1]^4 \times C^L \rightarrow \mathcal{D}(C)$,
for $k = 1, \ldots, L$ and all $n \in \mathbb{N}$, is defined by Eq.~\eqref{eq:seirs}, as explained in Section~\ref{sec:epi}.
Also the approximation $q_{n-1}^{(k)}$ of the empirical belief $\lambda x. \breve{\mu}_{n-1}^{(k)}(h_{n-1}^{(k)}, x)$ has signature 
$q_{n-1}^{(k)} : [0,1]^4 \rightarrow \mathcal{D}(C)$, for $k = 1, \ldots, L$ and all $n \in \mathbb{N}$.
Thus, for all $n \in \N$,
the intermediate distribution at node $k$ given by the transition update is
\[ \overline{q}_n^{(k)} =  \lambda x. \lambda c. \int_{C^L} \lambda s. \breve{\tau}_n^{(k)}(x, s)(c) \; \bigotimes_{k'=1}^L q_{n-1}^{(k')}(x) \; d \upsilon_{C^L}. \]
Since the underlying measure $\upsilon_{C^L}$ on $C^L$ is counting measure, this reduces to
\[  \overline{q}_n^{(k)} = 
         \lambda x. \lambda c. \sum_{s \in C^L}  \breve{\tau}_n^{(k)}(x,s)(c) \; \prod_{k'=1}^L q_{n-1}^{(k')}(x)(s_{k'}). 
\]

It is shown in Appendix B that the intermediate distribution resulting from the transition update is as follows.
\begin{equation}
\label{eq:factored:transition_update1}
\begin{aligned}
& \overline{q}_n^{(k)}(x)(\sS) = \rho \, q_{n-1}^{(k)}(x)(\sR) + \left( \textstyle\prod_{l \in \mathcal{N}_k} (1 - \beta q_{n-1}^{(l)}(x)(\sI)) \right)  q_{n-1}^{(k)}(x)(\sS),  \\
& \overline{q}_n^{(k)}(x)(\sE) = \left( 1 - \textstyle\prod_{l \in \mathcal{N}_k} (1 - \beta q_{n-1}^{(l)}(x)(\sI)) \right)  q_{n-1}^{(k)}(x)(\sS) +  (1 - \sigma) \, q_{n-1}^{(k)}(x)(\sE), \\
& \overline{q}_n^{(k)}(x)(\sI) =  \sigma \, q_{n-1}^{(k)}(x)(\sE) + (1 - \gamma) \, q_{n-1}^{(k)}(x)(\sI), \\
& \overline{q}_n^{(k)}(x)(\sR) =  \gamma \, q_{n-1}^{(k)}(x)(\sI) + (1 - \rho) \, q_{n-1}^{(k)}(x)(\sR),
\end{aligned}
\end{equation}
for all $x \in [0,1]^4$ and $n \in \N$, and $k = 1,\dots,L$.
In effect, Eq.~\eqref{eq:factored:transition_update1} could be used to define directly the transition update without specifying the transition schema.

For the observation update, let $o_n^{(k)} \in O_k$ be the testing outcome for node $k$ at time step $n$.
Then
\begin{equation}
\label{eq:factored:obs_update}
q_n^{(k)}(x)(c) = \frac{\breve{\xi}_n^{(k)}(c)(o_n^{(k)}) \; \overline{q}_n^{(k)}(x)(c)} 
                                  {\sum_{c \in C} \breve{\xi}_n^{(k)}(c)(o_n^{(k)}) \; \overline{q}_n^{(k)}(x)(c)},
\end{equation}
for all $x \in [0,1]^4$, $n \in \N$, and $c \in C$.
Note that Eq.~\eqref{eq:factored:obs_update} is a tractable expression that can be easily simplified to an explicit categorical distribution at each node.

Algorithm~\ref{algo:fcf} returns an approximation $q_n^{(k)} : [0, 1]^4 \rightarrow \mathcal{D}(C)$ 
of the conditional empirical belief $\lambda x. \breve{\mu}_n^{(k)}(h_n^{(k)}, x)$ at time $n$, for $k = 1, \ldots, L$.
From this, one can construct an approximation $\bigotimes_{k=1}^L q_n^{(k)} : [0, 1]^4 \rightarrow \mathcal{D}(C^L)$ of
$\lambda x. \breve{\mu}_n(h_n, x)$ at time $n$.
Note that each $q_n^{(k)} : [0, 1]^4 \rightarrow \mathcal{D}(C)$ is, in principle, defined for all $x \in [0, 1]^4$.
In this case, $\overline{q}_n^{(k)}(x)(c)$ in Eq.~\eqref{eq:factored:transition_update1} and 
$q_n^{(k)}(x)(c)$ in Eq.~\eqref{eq:factored:obs_update}, for each $c \in C$, can be easily evaluated. 
However, obtaining an approximation $q_n^{(k)}$ that is defined everywhere on its domain can be a difficult task -- 
this is the problem of Bayesian estimation of conditional densities generalized to the filtering setting.

For the extensions of the conditional filters to the case where the stochastic process on $X$ is arbitrary, the effort needed to obtain an approximation to the conditional empirical belief given by the filter that is defined everywhere can be justified.
Often there are reasoning tasks where, given some value in $X$, the agent needs to know the distribution on $Y$ given by
the conditional empirical belief for that value in order to assist in the choice of an action.
However, for the case where $X$ is a parameter space and the stochastic process on $X$ produces the same value in $X$ at each time step, the corresponding \emph{total} conditional empirical belief is not particularly meaningful.

More important in such cases is to be able to compute the distributions $(\nu_n \otimes \mu_n)(h_n)$ and $(\nu_n \odot \mu_n)(h_n)$.
Since Algorithm~\ref{algo:spf} is a particle filter, for this it is sufficient to be able to compute $q_n^{(k)}(x)$ just for the particles $x$ in the particle family at time $n$.
Thus suppose that we want to compute $q_n^{(k)}(x_n^{(i)})(c)$, for some $c \in C$, as given by Eq.~\eqref{eq:factored:transition_update1}, where $x_n^{(i)}$ is a particle in the particle family $(x_n^{(i)})_{i=1}^N$ at time $n$.
Substituting this value for $x$ in the equations, the term $q_{n-1}^{(k)}(x_n^{(i)})$ appears on the right hand sides.
Since $q_{n-1}^{(k)}(x_n^{(i)})$ is not (generally) defined, we appeal to the considerations of Section~\ref{sec:partial_def} and choose the $i^* \in \{ 1, \ldots, N \}$ such that $x_{n-1}^{(i^*)}$ approximates $x_n^{(i)}$.
Substituting $x_{n-1}^{(i^*)}$ for $x_n^{(i)}$ in the right hand sides of Eq.~\eqref{eq:factored:transition_update1}, 
the following approximations are obtained.
\begin{equation}
\label{eq:factored:transition_update2}
\begin{aligned}
& \overline{q}_n^{(k)}(x_n^{(i)})(\sS) \approx \rho \, q_{n-1}^{(k)}(x_{n-1}^{(i^*)})(\sR) + \left( \textstyle\prod_{l \in \mathcal{N}_k} (1 - \beta q_{n-1}^{(l)}(x_{n-1}^{(i^*)})(\sI)) \right)  q_{n-1}^{(k)}(x_{n-1}^{(i^*)})(\sS),  \\
& \overline{q}_n^{(k)}(x_n^{(i)})(\sE) \approx \left( 1 - \textstyle\prod_{l \in \mathcal{N}_k} (1 - \beta q_{n-1}^{(l)}(x_{n-1}^{(i^*)})(\sI)) \right)  q_{n-1}^{(k)}(x_{n-1}^{(i^*)})(\sS) +  (1 - \sigma) \, q_{n-1}^{(k)}(x_{n-1}^{(i^*)})(\sE), \\
& \overline{q}_n^{(k)}(x_n^{(i)})(\sI) \approx  \sigma \, q_{n-1}^{(k)}(x_{n-1}^{(i^*)})(\sE) + (1 - \gamma) \, q_{n-1}^{(k)}(x_{n-1}^{(i^*)})(\sI), \\
& \overline{q}_n^{(k)}(x_n^{(i)})(\sR) \approx  \gamma \, q_{n-1}^{(k)}(x_{n-1}^{(i^*)})(\sI) + (1 - \rho) \, q_{n-1}^{(k)}(x_{n-1}^{(i^*)})(\sR),
\end{aligned}
\end{equation}
for all 
$n \in \N$ and $i=1,\dots,N$,
and $k = 1,\dots,L$.
Now, for all $c \in C$, each $\overline{q}_n^{(k)}(x_n^{(i)})(c)$ in Eq.~\eqref{eq:factored:transition_update2} 
and each $q_n^{(k)}(x_n^{(i)})(c)$ in Eq.~\eqref{eq:factored:obs_update}
can be calculated.

\begin{figure*}[htbp]
    \centering
    \setlength{\tabcolsep}{.1in}
    \begin{tabular}{ccc}
    \includegraphics[height=1.2in]{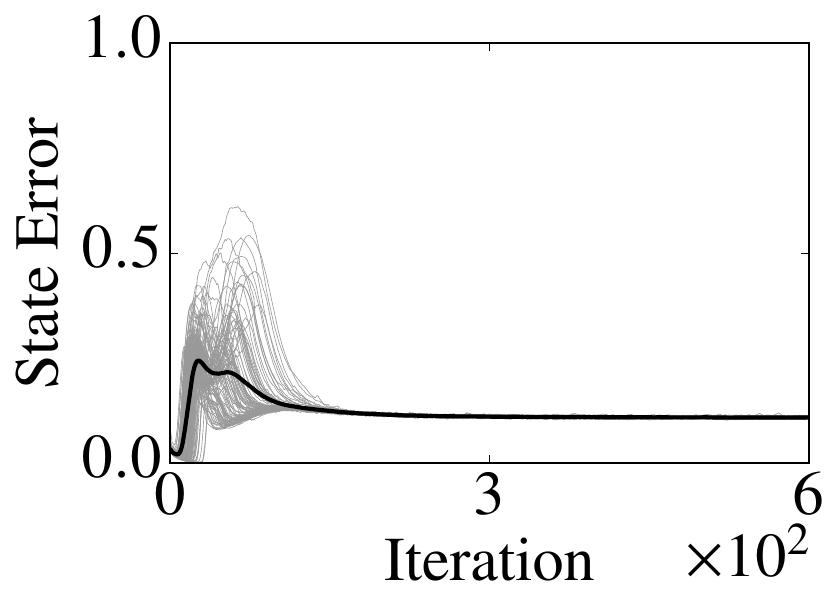} &
    \includegraphics[height=1.2in]{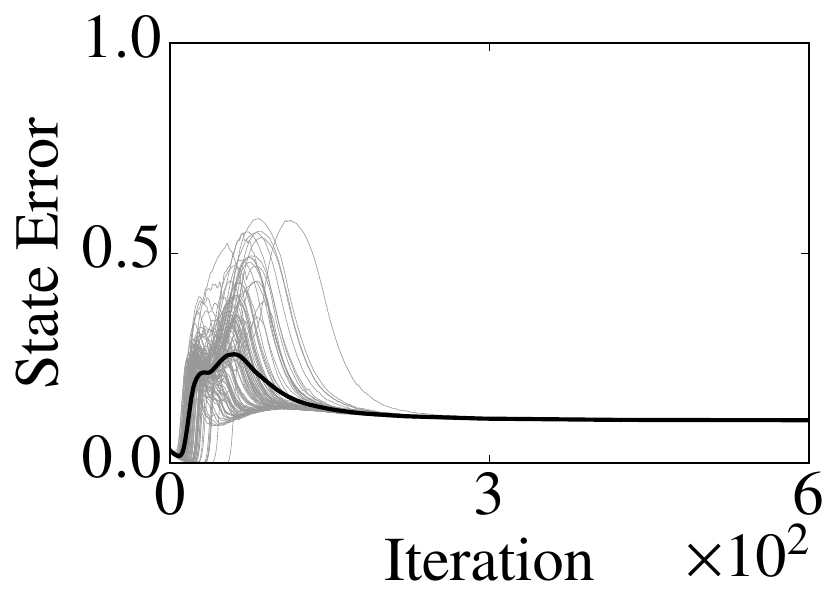} &
    \includegraphics[height=1.2in]{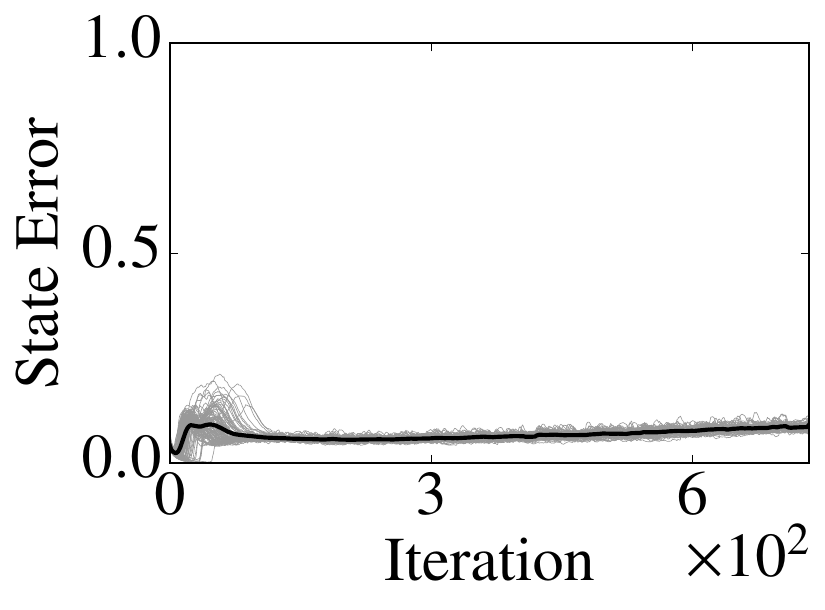} \\
    \includegraphics[height=1.2in]{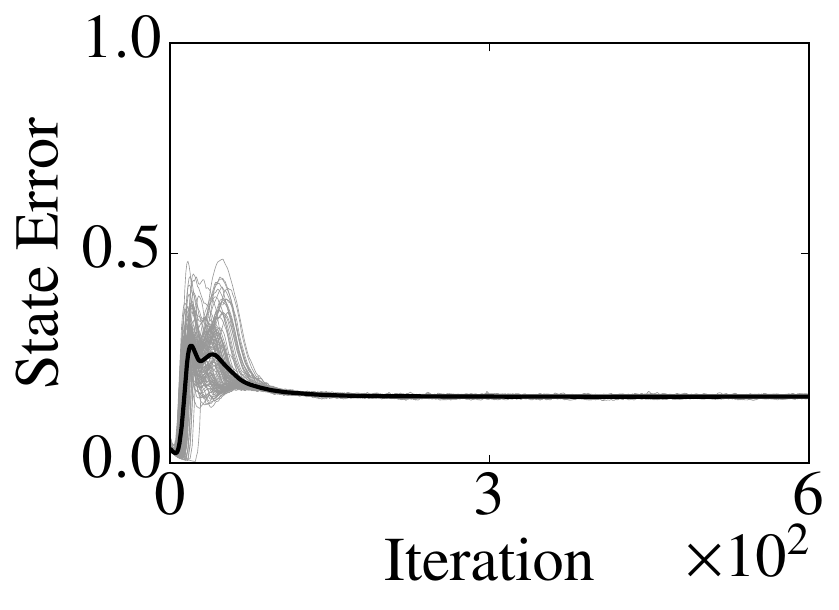} &
    \includegraphics[height=1.2in]{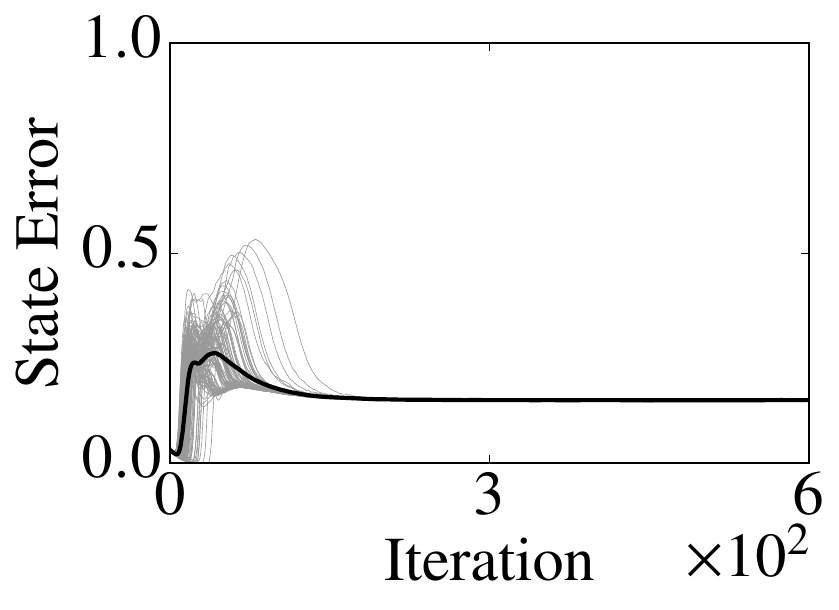} &
    \includegraphics[height=1.2in]{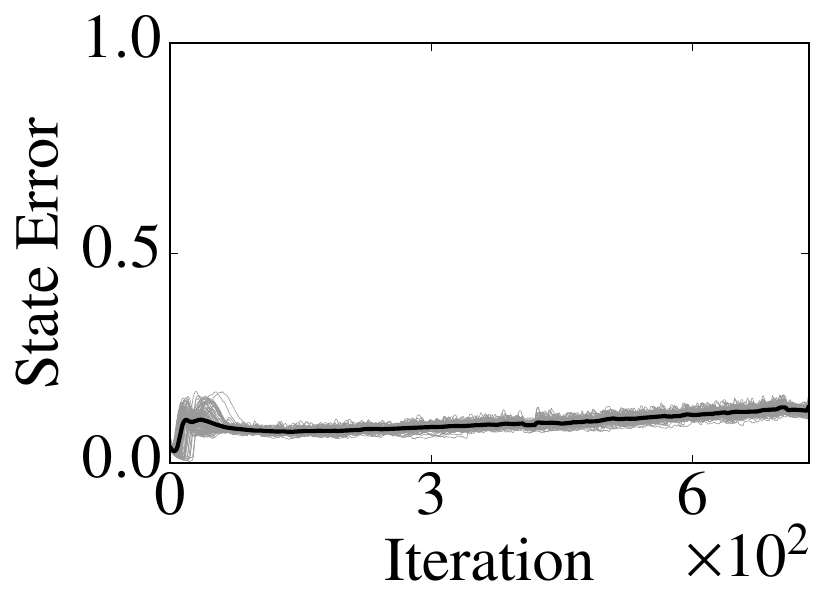} \\
    \end{tabular}
    \caption{
    State errors of an SEIRS epidemic model for three contact networks, 
    {\tt Gowalla}, {\tt Youtube}, and {\tt AS-733},
    using Algorithms~\ref{algo:fcf} and \ref{algo:spf} for 100 independent runs in which the disease does not die out in 600 time steps for the {\tt Gowalla} and the {\tt Youtube} contact networks, and 733 time steps for the {\tt AS-733} dynamic contact network. 
    The left column is for {\tt Gowalla},  the middle column is for {\tt Youtube}, and the right column is for {\tt AS-733}.
    The top row uses the parameters 
    $\beta=0.2, \, \sigma=1/3, \, \gamma=1/14, \,\rho=1/180, \, \lambda_{FP} = 0.1, \,\lambda_{FN} = 0.1$;
    the bottom row uses the parameters 
    $\beta=0.27, \, \sigma=1/2, \, \gamma=1/7, \, \rho=1/90, \, \lambda_{FP} = 0.1, \,\lambda_{FN} = 0.3$.
    Light gray: state errors from individual runs.
    Dark solid: mean state error averaged over 100 independent runs.
    }
    \label{fig:alg10:1}
\end{figure*}

\begin{figure*}[htbp]
    \centering
    \includegraphics[width=\linewidth]{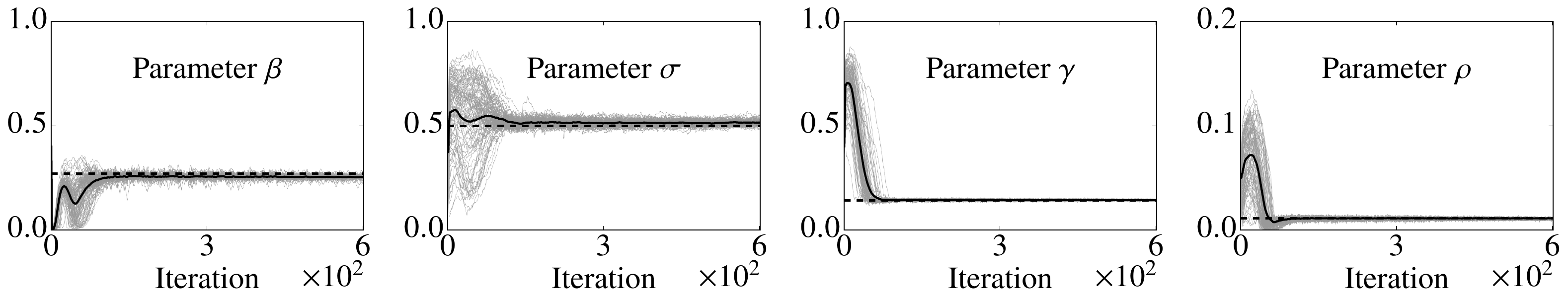}
    \caption{%
    Parameter estimation of an SEIRS epidemic model for the {\tt Gowalla} contact network using Algorithms~\ref{algo:fcf} and \ref{algo:spf} for 100 independent runs in which the disease does not die out in 600 time steps.
    The parameters 
    are $\beta=0.27, \, \sigma=1/2, \, \gamma=1/7, \,\rho=1/90, \, \lambda_{FP} = 0.1, \, \lambda_{FN} = 0.3$.
    Light gray: 
    estimates from individual runs.
    Dark solid: mean 
    estimate averaged over 100 independent runs.
    Dark dashed: true parameter values.
    }
    \label{fig:alg10:2}
\end{figure*}

\begin{figure*}[htbp]
    \centering
    \includegraphics[width=\linewidth]{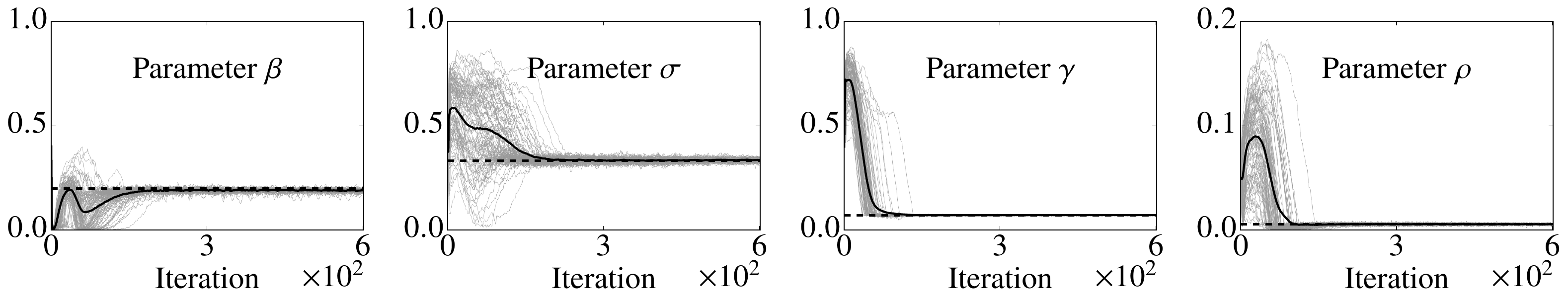}
    \caption{
    Parameter estimation of an SEIRS epidemic model for the {\tt Youtube} contact network using Algorithms~\ref{algo:fcf} and \ref{algo:spf} for 100 independent runs in which the disease does not die out in 600 time steps.
    The parameters 
    are $\beta=0.2, \, \sigma=1/3, \, \gamma=1/14, \, \rho=1/180, \, \lambda_{FP} = 0.1, \, \lambda_{FN} = 0.1$.
    Light gray: 
    estimates from individual runs.
    Dark solid: mean 
    estimate averaged over 100 independent runs.
    Dark dashed: true parameter values.
    }
    \label{fig:alg10:3}
\end{figure*}

\begin{figure*}[htbp]
    \centering
    \includegraphics[width=\linewidth]{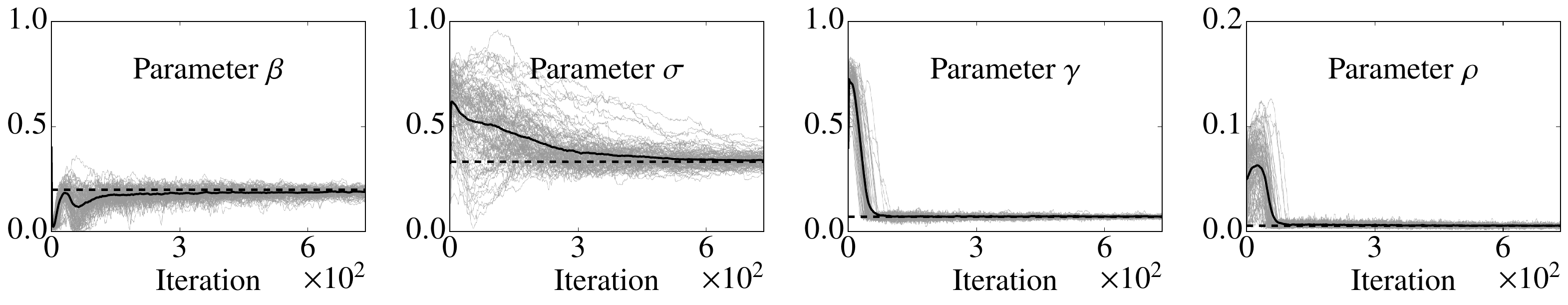}
    \caption{
    Parameter estimation of an SEIRS epidemic model for the {\tt AS-733} dynamic contact network using Algorithms~\ref{algo:fcf} and \ref{algo:spf} for 100 independent runs in which the disease does not die out in 733 time steps.
    The parameters
    are $\beta=0.2, \, \sigma=1/3, \, \gamma=1/14, \, \rho=1/180, \, \lambda_{FP} = 0.1, \, \lambda_{FN} = 0.1$.
    Light gray: 
    estimates from individual runs.
    Dark solid: mean estimate 
    averaged over 100 independent runs.
    Dark dashed: true parameter values.
    }
    \label{fig:alg10:4}
\end{figure*}

Figure~\ref{fig:alg10:1} shows the results of tracking states, and Figures~\ref{fig:alg10:2},  \ref{fig:alg10:3}, and \ref{fig:alg10:4}  the results of estimating parameters using factored conditional filtering (Algorithms~\ref{algo:fcf} and \ref{algo:spf}) for the contact networks {\tt Gowalla}, {\tt Youtube}, and {\tt AS-733} in Table~\ref{tab:dataset}.
Note that the state errors for the \texttt{AS-733} dynamic contact network gradually increase (as shown in
Figure~\ref{fig:alg10:1}), although the estimated parameters stay reasonably accurate (see Figure~\ref{fig:alg10:4}). 
This is likely to be a result of the constant changing of the network structure at each time step in the \texttt{AS-733} dataset.

\subsection{Factored Conditional Particle Filter}
\label{sec:fcpf}
Next we consider the factored conditional particle filter, Algorithm~\ref{algo:fcpf} (fully factored), 
paired with Algorithm~\ref{algo:spf} (using observation schema synthesis and a jittering transition schema).

For the simulation, the settings for this application are exactly the same as presented in Section~\ref{sec:fcf}.
For the filter, the settings are the same as in Section~\ref{sec:fcf}, except that Algorithm~\ref{algo:fcpf} is used instead of Algorithm~\ref{algo:fcf} and there are a larger number of parameter particles.
{For Algorithm~\ref{algo:spf}, we initialize $N=600$ parameter particles by
sampling from the uniform distributions given by Eq.~\eqref{eq:param:init}; for Algorithm~\ref{algo:fcpf}, we sample $M = 1024$ particles on each factor from the initial state distribution for the filter presented in Section~\ref{sec:fcf}.

In this application, both the nonconditional filter Algorithm~\ref{algo:spf} and the conditional filter Algorithm~\ref{algo:fcpf} are particle filters.
Thus the conditional empirical belief given by conditional filter Algorithm~\ref{algo:fcpf} is only partially defined; specifically, at time $n$, the conditional empirical belief is only defined on the particles 
in the particle family $(x_n^{(i)})_{i=1}^N$ given by the nonconditional particle filter at time $n$.
Thus the considerations of Section~\ref{sec:partial_def} apply.
The problem is to know the value of the tuple $\phi_{n-1}^{(l)}(x_n^{(i)})$ in Line 4 of Algorithm~\ref{algo:fcpf}, 
where $x_n^{(i)}$ is a parameter particle at time $n$.
Analogous to the discussion in Section~\ref{sec:partial_def}, there exists $i^* \in \{ 1, \ldots, N \}$ such that
$\phi_{n-1}^{(l)}(x_{n-1}^{(i^*)})$ replaces $\phi_{n-1}^{(l)}(x_n^{(i)})$ in later calculations.

Figures~\ref{fig:alg11:1} and~\ref{fig:alg11:2} show the results of tracking states and estimating parameters for the {\tt OpenFlights} contact network using factored conditional particle filtering (Algorithms~\ref{algo:fcpf} and \ref{algo:spf}) for the
parameters in Eq.~\eqref{eq:experiment:setup2}.
It can be seen that the performance for tracking states and estimating parameters using factored conditional particle filtering is less effective compared to that of factored conditional filtering (Algorithms~\ref{algo:fcf} and~\ref{algo:spf}).
We remark that Algorithm~\ref{algo:fcpf} is also more computationally expensive than Algorithm~\ref{algo:fcf}
and thus unlikely to scale to graphs as large as the {\tt Youtube} contact network.
Nevertheless, Algorithm~\ref{algo:fcpf} could be employed in settings where exact filtering as in Algorithm~\ref{algo:fcf} is not applicable.

\begin{figure}[htbp]
    \centering
    \includegraphics[height=1.21in]{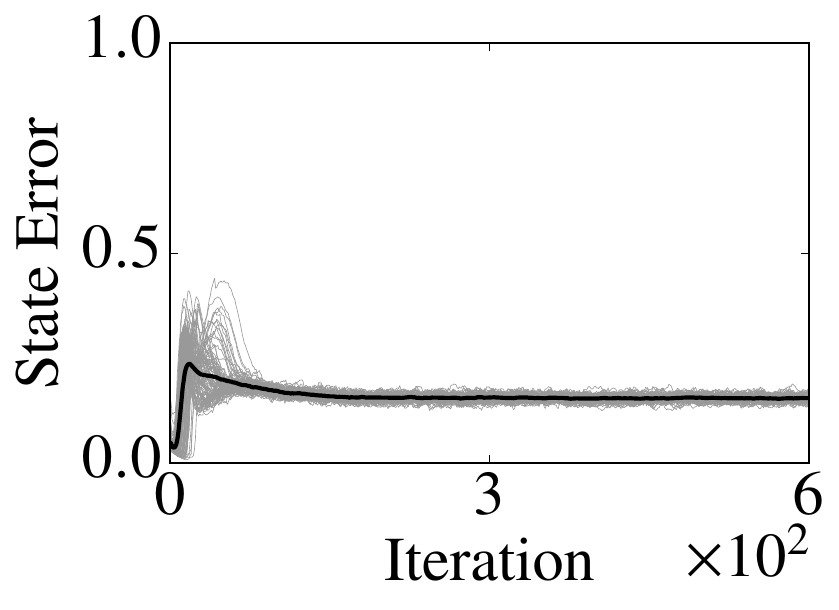}
    \caption{
    State errors of an SEIRS epidemic model for the {\tt OpenFlights} contact network using Algorithms~\ref{algo:fcpf} and \ref{algo:spf} for 100 independent runs in which the disease does not die out in 600 time steps. 
    The parameters 
    are $\beta=0.27, \, \sigma=1/2, \, \gamma=1/7, \,\rho=1/90, \, \lambda_{FP} = 0.1, \, \lambda_{FN} = 0.3$.
    Light gray: state errors from individual runs.
    Dark solid: mean state error averaged over 100 independent runs.
    }
    \label{fig:alg11:1}
\end{figure}

\begin{figure*}[htbp]
    \centering
    \includegraphics[width=\linewidth]{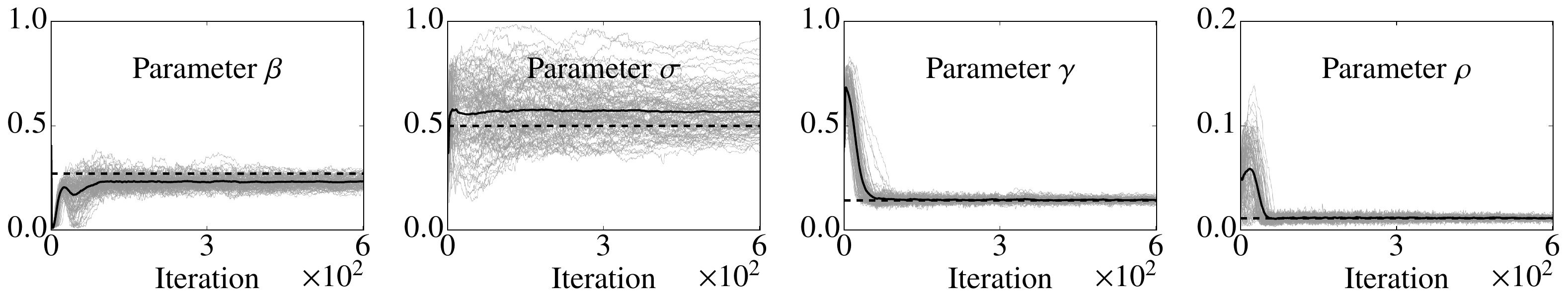}
    \caption{
    Parameter estimation of an SEIRS epidemic model for the {\tt OpenFlights} contact network using Algorithms~\ref{algo:fcpf} and \ref{algo:spf} for 100 independent runs in which the disease does not die out in 600 time steps.
    The parameters 
    are $\beta=0.27, \, \sigma=1/2, \, \gamma=1/7, \, \rho=1/90$.
    Light gray: estimates 
    from individual runs.
    Dark solid: 
    mean estimate averaged over 100 independent runs.
    Dark dashed: true parameter values.
    }
    \label{fig:alg11:2}
\end{figure*}

\subsection{Factored Conditional Variational Filter}
\label{sec:fcvf}

Finally we consider the factored conditional variational filter, Algorithm~\ref{algo:fcvf} (fully factored), paired with Algorithm~\ref{algo:spf} (using observation schema synthesis and a jittering transition schema).

For the subpopulation contact network application of this subsection, the state space for the simulation is $S^L$ and the transition schema is given by Eq.~\eqref{eq:subpopulation:T_seirs_trans2} (together with Eq.~\ref{eq:subpopulation:T_seirs_trans}), for $K = 3$. 
The observation schema is given by Eq.~\eqref{obs_update_multinomial} with $m=5$.
The probability of a node being observed at a time step is $\alpha = 0.7$.
Note that $\mathit{Mult}(m, (s_1, s_2, s_3, s_4))$ in Eq.~\eqref{obs_update_multinomial}
can be written in the form
\vspace{-.3em}
\[ \lambda o. \frac{\Gamma(m+1)}{\prod_{j=1}^4 \Gamma(o_j +1)} \prod_{j=1}^4 s_j^{o_j}. 
\]

The state space for the filter is $S^L$.
The filtering algorithm is a fully factored version of the factored conditional variational filter. 
Recall that
the parameter space is $[0, 1]^4$.
The observation schema for the filter is given by Eq.~\eqref{obs_update_multinomial} (applied at the subpopulation node level).

The details of the initializations are as follows. 
(For subpopulation contact networks, the terminology `subpopulation zero' is used instead of `patient zero'.)
For the simulation, subpopulation zero, labelled by (0.01, 0.97, 0.01, 0.01), is chosen uniformly at random.
All other nodes are initially labelled by (0.97, 0.01, 0.01, 0.01).
For the filter, the initial state distribution is as follows.
Let $n_0$ be subpopulation zero in the simulation.
Then the initial state distribution for the filter is given by the product of the Dirichlet distributions
\begin{itemize}
\item $\mathit{Dir}(0.29, 0.4, 0.3, 0.01)$, for node $n_0$,
\item $\mathit{Dir}(0.49, 0.3, 0.2, 0.01)$, for the neighbours of node $n_0$,
\item $\mathit{Dir}(0.69, 0.2, 0.1, 0.01)$, for nodes that are 2 steps away from $n_0$, and
\item $\mathit{Dir}(0.97, 0.01, 0.01, 0.01)$, for nodes that are 3 or more steps away from $n_0$.
\end{itemize}

Instead of defining a transition schema, a transition update is defined directly.
For the factored conditional variational filter, it is important that the intermediate distribution simplify in an analogous way as the factored conditional filter does in Eq.~\eqref{eq:factored:transition_update1}.
This is because the distribution obtained at the end of the observation update needs to be tractable so that the variational update can be performed.
Since there does not seem to be such a transition schema that could plausibly model the application correctly and also give an analytically simple intermediate distribution, we follow a different path and define the intermediate distribution {\em directly}, without the explicit use of a transition schema.
If it is true that state distributions for this application can be (approximately) modelled by products of Dirichlet distributions, 
then this approach is reasonable.

Let $\mathcal{D}_{P D}(S^L)$ be the set of all distributions on $S^L$ that are products of Dirichlet distributions on each factor $S$.
For all $n \in \mathbb{N}$ and $k = 1, \ldots, L$, define 
\[ \mathcal{T}_n^{(k)}: [0, 1]^4 \times \mathcal{D}_{P D}(S^L) \rightarrow \mathcal{D}(S) 
\vspace{-.5em}
\]
by 
\vspace{-1em}
\[
\mathcal{T}_n^{(k)}((\beta, \sigma, \gamma, \rho), \bigotimes_{k'=1}^L \mathit{Dir}(\alpha_1^{(k')}, \alpha_2^{(k')}, \alpha_3^{(k')}, \alpha_4^{(k')})) = \mathit{Dir}(K\beta_1^{(k)}, K\beta_2^{(k)}, K\beta_3^{(k)}, K\beta_4^{(k)}),
\]
where
\begin{equation}
\label{eq:subpopulation:T_seirs_trans:update}
\begin{aligned}
& \beta_1^{(k)}  \triangleq \rho \overline{\alpha}_{4}^{(k)} + 
    \overline{i_k}\, \overline{i_{\mathcal{N}_k}} \overline{\alpha}_{1}^{(k)}
& \hspace{3em}
& \beta_2^{(k)} \triangleq 
(1 - \overline{i_k}\, \overline{i_{\mathcal{N}_k}}) \overline{\alpha}_{1}^{(k)}
 + (1 - \sigma) \overline{\alpha}_2^{(k)} 
\\
& \beta_3^{(k)} \triangleq \sigma \overline{\alpha}_{2}^{(k)} + (1 - \gamma) \overline{\alpha}_{3}^{(k)} &
& \beta_4^{(k)} \triangleq \gamma \overline{\alpha}_{3}^{(k)} + (1 - \rho) \overline{\alpha}_{4}^{(k)} \\
& \overline{i_k}  \triangleq (1 - \beta \kappa_1 \overline{\alpha}_{3}^{(k)})^{M_k-1} &
& \overline{i_{\mathcal{N}_k}} \triangleq \prod_{l \in \mathcal{N}_k} (1 - \beta \kappa_2 \overline{\alpha}_{3}^{(l)})^{M_l} 
\end{aligned}
\end{equation}
and, for $k = 1, \ldots, L$ and $j = 1, \ldots, 4$, 
\begin{gather*}
\overline{\alpha}_j^{(k)} \triangleq \frac{\alpha_j^{(k)}}{\sum_{i=1}^4 \alpha_i^{(k)}}.
\vspace{-.5em}
\end{gather*}

To understand Eq.~\eqref{eq:subpopulation:T_seirs_trans:update}, note that the transition schema given by Eq.~\eqref{eq:subpopulation:T_seirs_trans2} contains a number of terms of the form $(1-\beta)^{M_l \kappa_2 s_{l,3}}$ that reflect, on average, the probability that a susceptible individual $j_k$ remains uninfected by its neighbouring individuals. 
In the transition update, 
we need to consider the same event but probability-weighted by the number $n(j_k)$ of infectious
neighbouring individuals connected to $j_k$, which is given by
\begin{align*}
    \sum_{d=0}^{M_l} \Pr(n(j_k) = d) (1 - \beta)^{d}
    \ &= \
    \sum_{d=0}^{M_l} \binom{M_l}{d} (\kappa_2 \overline{\alpha}_3^{(l)})^d (1 - \kappa_2 \overline{\alpha}_3^{(l)})^{M_l - d}(1 - \beta)^d  \\
    \ &= \
    ((1 - \kappa_2 \overline{\alpha}_3^{(l)}) + (1 - \beta) \kappa_2 \overline{\alpha}_3^{(l)})^{M_l} \\
    \ &= \
    (1 - \beta \kappa_2 \overline{\alpha}_3^{(l)})^{M_l},
\end{align*} 
where the second-to-last line comes from the binomial theorem.
The $(1-\beta)^{(M_k-1) \kappa_1 s_{k,3}}$ term in Eq.~\eqref{eq:subpopulation:T_seirs_trans2} can be analyzed in a similar way to yield $(1 - \beta \kappa_1 \overline{\alpha}_3^{(k)})^{M_k-1}$.

Then, using the function  $\mathcal{T}_n^{(k)}$, Line 2 
of Algorithm~\ref{algo:fcvf} (specialized to this application) 
\[ \overline{q}_n^{(k)} := \lambda x. \lambda y. \int_{S^L} \lambda y'. \breve{\tau}_n^{(k)}(h_{n-1}^{(k)}, a_n, x, y')(y) \; \bigotimes_{k'=1}^L q_{n-1}^{(k')}(x) \; d \upsilon_{S^L} \]
is replaced by
\[ \overline{q}_n^{(k)} := \lambda x. \mathcal{T}_n^{(k)}(x, \bigotimes_{k'=1}^L q_{n-1}^{(k')}(x)). \]
(Here, $\upsilon_{S^L}$ is the product of the Lebesgue measures on each factor $S$.)
Consequently, for all $x \in [0,1]^4$, $\overline{q}_n^{(k)}(x) \in \mathcal{D}(S)$.
The function $\mathcal{T}_n^{(k)}$ defines the transition update directly without the need for a transition schema $\breve{\tau}_n^{(k)}$.
Defining the transition update directly like this is an alternative to defining it indirectly via a transition schema 
and has the advantage of providing greater control over the set of distributions that can be intermediate distributions.

In this application, the nonconditional filter Algorithm~\ref{algo:spf} is a particle filter.
Furthermore, the conditional empirical belief given by conditional filter Algorithm~\ref{algo:fcvf} is only partially defined; specifically, at time $n$, the conditional empirical belief is only defined on the particles 
in the particle family $(x_n^{(i)})_{i=1}^N$.
Thus the considerations of Section~\ref{sec:partial_def} apply.
So consider 
\[ \overline{q}_n^{(k)}(x_n^{(i)}) = \mathcal{T}_n^{(k)}(x_n^{(i)}, \bigotimes_{k'=1}^L q_{n-1}^{(k')}(x_n^{(i)})), \]
for some $i \in \{1, \ldots, N\}$.
Note that $q_{n-1}^{(k')}(x_n^{(i)})$ may not be defined.
However, according to Section~\ref{sec:partial_def}, $x_n^{(i)}$ is a perturbation of $x_{n-1}^{(i^*)}$, 
for some $i^* \in \{1, \ldots, N\}$.
Thus
\[ \overline{q}_n^{(k)}(x_n^{(i)}) \approx \mathcal{T}_n^{(k)}(x_n^{(i)}, \bigotimes_{k'=1}^L q_{n-1}^{(k')}(x_{n-1}^{(i^*)})) \]
and this approximation is used in the following derivation.

Now we consider the details of Algorithm~\ref{algo:fcvf} for this application.
The transition update corresponding to Line 2 of Algorithm~\ref{algo:fcvf} is described above.
For the observation update in Line 3, it is necessary to calculate, for $k = 1, \ldots, L$ and for $i = 1, \ldots, N$,
\[ \displaystyle{p_n^{(k)}(x_n^{(i)}) := \frac{\lambda y. \breve{\xi}_n^{(k)}(y)(o_n^{(k)}) \; \overline{q}_n^{(k)}(x_n^{(i)})}
              {\int_S \lambda y. \breve{\xi}_n^{(k)}(y)(o_n^{(k)}) \; \overline{q}_n^{(k)}(x_n^{(i)})  \; d \upsilon_S}}. \]
Suppose that 
\[ \overline{q}_n^{(k)}(x_n^{(i)}) = \mathit{Dir}(K\beta_1^{(i,k)}, K\beta_2^{(i,k)}, K\beta_3^{(i,k)}, K\beta_4^{(i,k)}), \]
for $i = 1, \ldots, N$, and $o_n^{(k)} = (o_1, o_2, o_3, o_4)$.
Hence, for $i = 1, \ldots, N$,
\begin{equation*}
\begin{aligned}
p_n^{(k)}(x_n^{(i)}) \
& \propto \ \lambda y. \breve{\xi}_n^{(k)}(y)(o_n^{(k)}) \; \overline{q}_n^{(k)}(x_n^{(i)}) \\  
&= \ \lambda y. \frac{\Gamma(m + 1)}{\prod_{j=1}^4 \Gamma(o_j + 1)} \prod_{j=1}^4 y_j^{o_j} \;
              \mathit{Dir}(K\beta_1^{(i,k)}, K\beta_2^{(i,k)}, K\beta_3^{(i,k)}, K\beta_4^{(i,k)}) \\
&= \ \lambda y. \frac{\Gamma(m + 1)}{\prod_{j=1}^4 \Gamma(o_j + 1)} \prod_{j=1}^4 y_j^{o_j} \;
            \lambda y. \frac{\Gamma(\sum_{j=1}^4 K\beta_j^{(i,k)}))}{\prod_{j=1}^4 \Gamma(K\beta_j^{(i,k)})} \prod_{j=1}^4 y_j^{K\beta_j^{(i,k)} - 1}  \\
&= \ \lambda y. \frac{\Gamma(m + 1)}{\prod_{j=1}^4 \Gamma(o_j + 1)} \; \frac{\Gamma(\sum_{j=1}^4 K\beta_j^{(i,k)})}{\prod_{j=1}^4 \Gamma(K\beta_j^{(i,k)})}  
              \; \prod_{j=1}^4 y_j^{o_j + K\beta_j^{(i,k)} - 1} \\
&= \ \lambda y. \frac{\Gamma(m + 1)}{\prod_{j=1}^4 \Gamma(o_j + 1)} \; \frac{\Gamma(K)}{\prod_{j=1}^4 \Gamma(K\beta_j^{(i,k)})}  
              \; \prod_{j=1}^4 y_j^{o_j + K\beta_j^{(i,k)} - 1}.
              \hspace*{3.5em}
              \left[ \text{by } \sum_{j=1}^4 \beta_j^{(i,k)} = 1 \right]
\end{aligned}
\end{equation*}

The next step is the variational update in Line 4 of Algorithm~\ref{algo:fcvf}.
Let $Q^{(k)}$ be the set of Dirichlet distributions on $S$, 
for $k = 1, \ldots, L$.
The issue is to calculate $q_n^{(k)}(x_n^{(i)})$, where
\[ q_n^{(k)}(x_n^{(i)}) = \argmin_{r \in Q^{(k)}}  KL(p_n^{(k)}(x_n^{(i)}) \| r), 
\] 
for $i = 1, \ldots, N$.

Suppose that $r$ has the form $\mathit{Dir}(\gamma_1^{(i,k)}, \gamma_2^{(i,k)}, \gamma_3^{(i,k)}, \gamma_4^{(i,k)})$.
Then, as shown in Appendix C,
for $i = 1, \ldots, N$,
\begin{equation*}
\begin{aligned}
  KL(p_n^{(k)}(x_n^{(i)}) \| r) 
  = & \; \mathit{Const}  - \left( \left(\log \Gamma (\sum_{j=1}^4 \gamma_j^{(i,k)}) - \log \prod_{j=1}^4 \Gamma (\gamma_j^{(i,k)}) \right) \; + 
  \sum_{j=1}^4 (\gamma_j^{(i,k)} - 1) (\psi(o_j + K \beta_j^{(i,k)}) - \psi(m + K)) \right),
\end{aligned}
\end{equation*}
where $\psi$ is the digamma function.

Let
\[ f \; \triangleq \; \lambda (\gamma_1^{(i,k)}, \gamma_2^{(i,k)}, \gamma_3^{(i,k)}, \gamma_4^{(i,k)}). KL(p_n^{(k)}(x_n^{(i)}) \| r)
\; : \;
\mathbb{R}_+^4 \rightarrow \mathbb{R},
\] 
where $\mathbb{R}_+$ is the set of positive real numbers.
Note that the partial derivatives $D_j f$ of $f$ exist at each point $(\gamma_1^{(i,k)}, \gamma_2^{(i,k)}, \gamma_3^{(i,k)}, \gamma_4^{(i,k)}) \in \mathbb{R}_+^4$ and
\begin{equation*}
\begin{aligned}
D_j f (\gamma_1^{(i,k)}, \gamma_2^{(i,k)}, \gamma_3^{(i,k)}, \gamma_4^{(i,k)}) 
& \ = \
\psi(\gamma_j^{(i,k)}) - \psi(\sum_{j=1}^4 \gamma_j^{(i,k)}) - (\psi(o_j + K \beta_j^{(i,k)}) - \psi(m + K)), 
\end{aligned}
\end{equation*}
for $j = 1, \ldots, 4$.
Thus, 
\[ D_j f(o_1 + K \beta_1^{(i,k)}, \ldots, o_4 + K \beta_4^{(i,k)}) = 0, \]
for $j = 1, \ldots, 4$.
Furthermore, the Hessian matrix of $f$ 
is positive definite.
Consequently, $f$ is a convex function and has a (global) minimum at $(o_1 + K \beta_1^{(i,k)}, \ldots, o_4 + K \beta_4^{(i,k)})$.
In summary,
\[ q_n^{(k)}(x_n^{(i)}) = \mathit{Dir}(o_1 + K \beta_1^{(i,k)}, \ldots, o_4 + K \beta_4^{(i,k)}), \]
for $i = 1, \ldots, N$.
Since it is only necessary to consider $x \in X$ that belong to the parameter particle family $(x_n^{(i)})_{i=1}^N$,
these equations provide a simple and efficient computation of $q_n^{(k)}$ in Line 4 of Algorithm~\ref{algo:fcvf}.

In Appendix E, the following formula for the state error $\mathit{Err}(\reallywidehat{(\nu_n \odot\mu_n)(h_n)})$ is derived:
\begin{equation*}
  \mathit{Err}(\reallywidehat{(\nu_n \odot \mu_n)(h_n)}) 
= \frac{1}{LN} \sum_{k=1}^L \sum_{i=1}^N \int_S \lambda y. \sum_{j=1}^4 | \widetilde{y}_{n,j}^{(k)} - y_j | \; q^{(k)}_n(x^{(i)}_n) \; d \upsilon_S.
\end{equation*}
Thus the state error is the average over all factors $k$ and all parameter particles $x^{(i)}_n$  of the state error for the 
$k$th factor and the $i$th parameter particle.
The integral can be evaluated directly using the fact that, for $j = 1, \ldots, 4$,
\begin{equation*}
\begin{aligned}
\int_S \lambda y. | c - y_j | \; \mathit{Dir}(\alpha_1, \alpha_2, \alpha_3, \alpha_3) \; d \upsilon_S 
=
  2 \left( c I(c; \alpha_j, \alpha_0 - \alpha_j) - \frac{\alpha_j}{\alpha_0} I(c; \alpha_j + 1, \alpha_0 - \alpha_j) \right) + \frac{\alpha_j}{\alpha_0} - c, 
\end{aligned}
\end{equation*}
where $I(z; a, b)$ is the regularized incomplete beta function, $\alpha_0 = \sum_{j=1}^4 \alpha_j$, and $c \in [0, 1]$. 

\begin{figure*}[htbp]
    \centering
    \setlength{\tabcolsep}{.1in}
    \begin{tabular}{ccc}
    \includegraphics[height=1.2in]{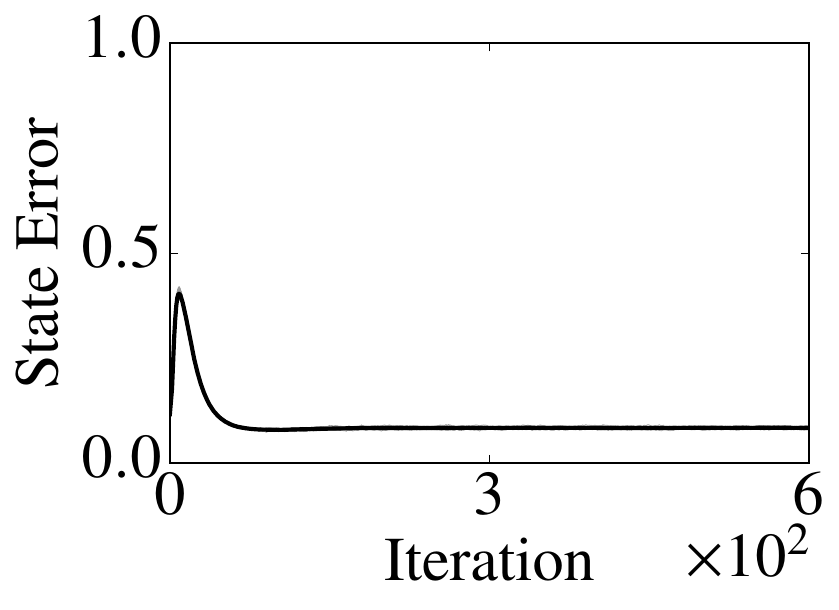} &
    \includegraphics[height=1.2in]{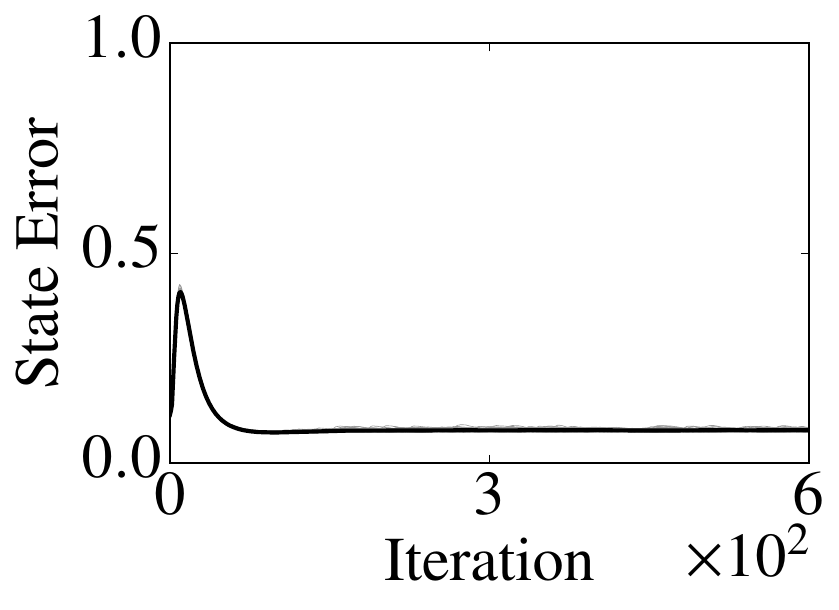} &
    \includegraphics[height=1.2in]{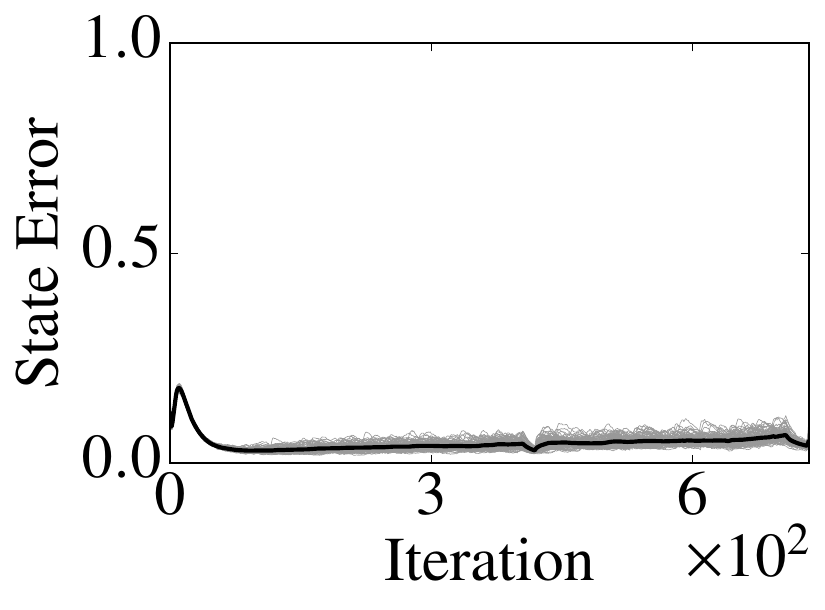} \\
    \includegraphics[height=1.2in]{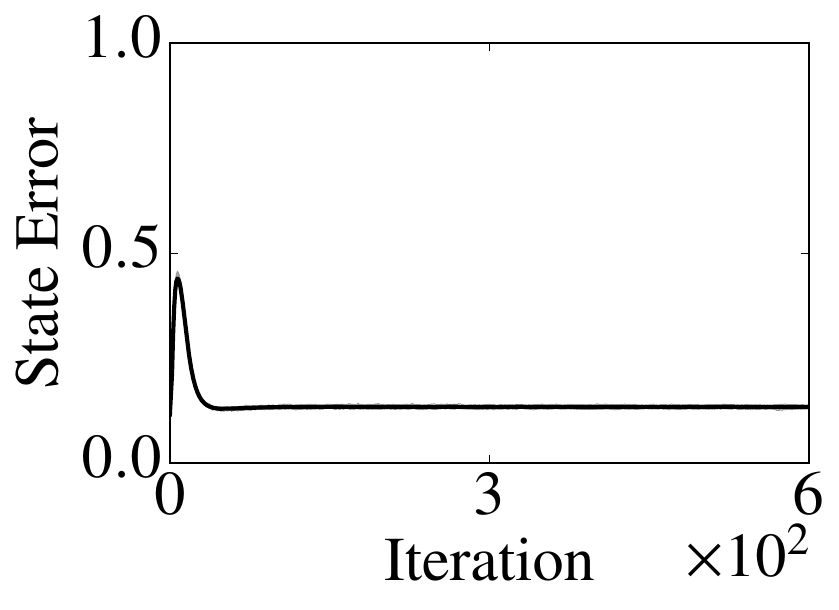} &
    \includegraphics[height=1.2in]{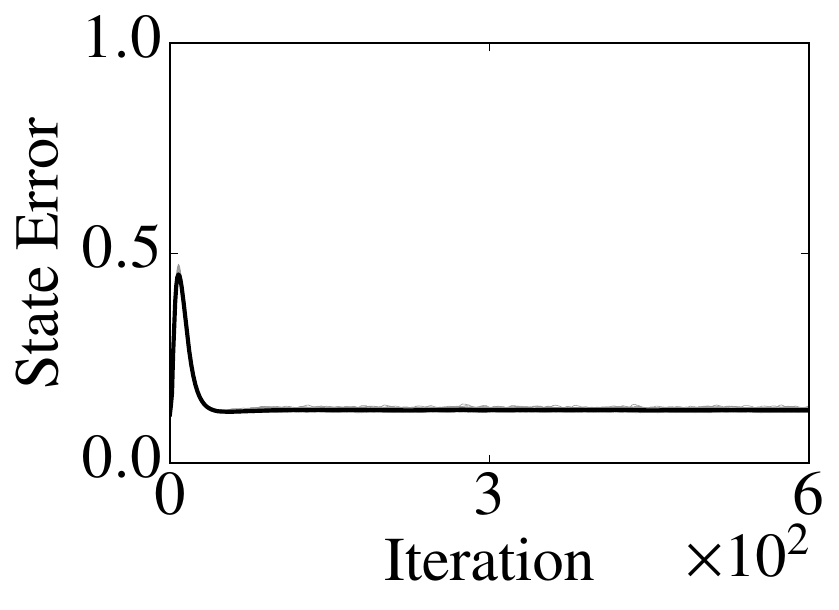} &
    \includegraphics[height=1.2in]{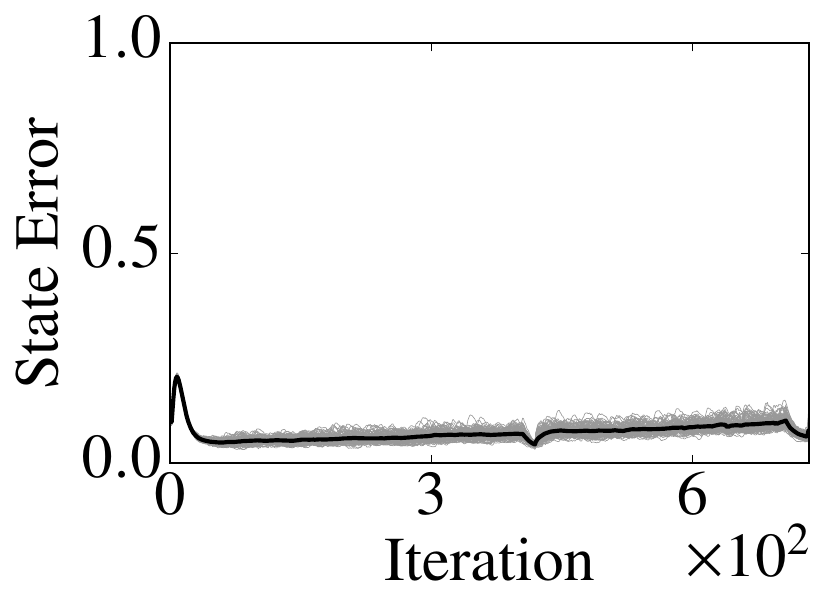}
    \end{tabular}
    \caption{
    State errors of an SEIRS epidemic model for three contact networks, 
    {\tt Facebook}, {\tt Gowalla,} and {\tt AS-733},
    using Algorithms~\ref{algo:fcvf} and \ref{algo:spf} for 100 independent runs in which the disease does not die out in 600 time steps for the {\tt Facebook} and the {\tt Gowalla} contact networks, and 733 time steps for the {\tt AS-733} dynamic contact network. 
    The left column is for {\tt Facebook},  the middle column is for {\tt Gowalla}, and the right column is for {\tt AS-733}.
    The top row uses the parameters 
    $\beta=0.2, \, \sigma=1/3, \, \gamma=1/14, \, \rho=1/180, \, \lambda_{FP} = 0.1, \, \lambda_{FN} = 0.1$; 
    the bottom row uses the parameters 
    $\beta=0.27, \, \sigma=1/2, \, \gamma=1/7, \, \rho=1/90, \, \lambda_{FP} = 0.1, \, \lambda_{FN} = 0.3$.
    Light gray: state errors from individual runs.
    Dark solid: mean state error averaged over 100 independent runs.
    }
    \label{fig:alg12:1}
\end{figure*}

\begin{figure*}[htbp]
    \centering
    \includegraphics[width=\linewidth]{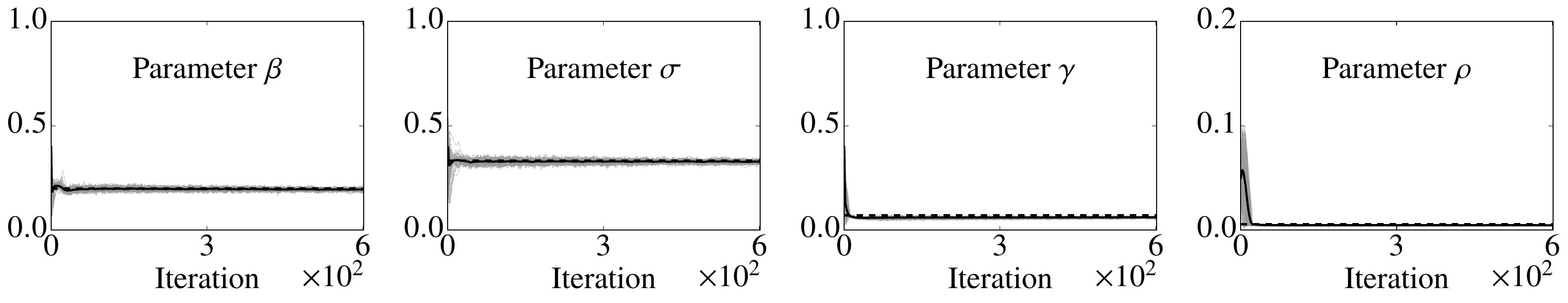}
    \caption{
    Parameter estimation of an SEIRS epidemic model for the {\tt Facebook} contact network using Algorithms~\ref{algo:fcvf} and \ref{algo:spf} for 100 independent runs in which the disease does not die out in 600 time steps.
    The parameters 
    are $\beta=0.2, \, \sigma=1/3, \, \gamma=1/14, \, \rho=1/180, \, \lambda_{FP} = 0.1, \, \lambda_{FN} = 0.1$.
    Light gray: 
    estimates from individual runs.
    Dark solid: mean 
    estimate averaged over 100 independent runs.
    Dark dashed: true parameter values.
    }
    \label{fig:alg12:2}
\end{figure*}

\begin{figure*}[htbp]
    \centering
    \includegraphics[width=\linewidth]{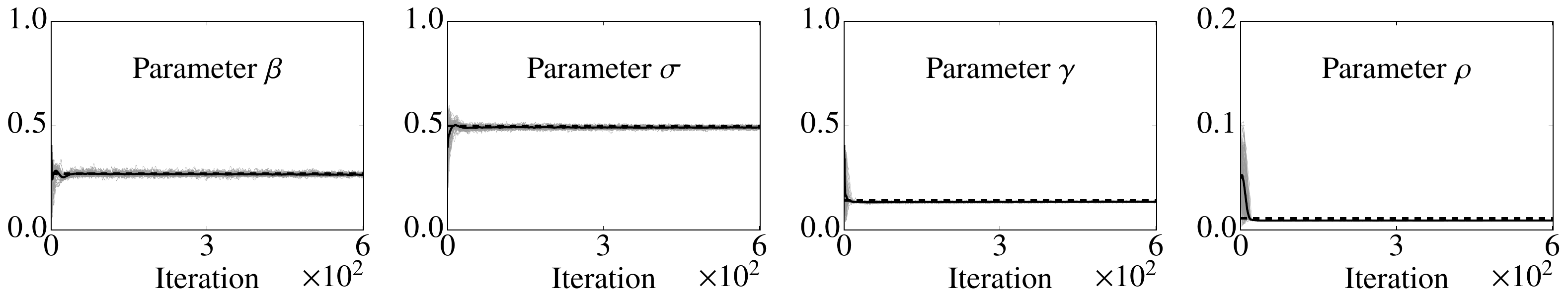}
    \caption{
    Parameter estimation of an SEIRS epidemic model for the {\tt Gowalla} contact network using Algorithms~\ref{algo:fcvf} and \ref{algo:spf} for 100 independent runs in which the disease does not die out in 600 time steps.
    The parameters 
    are $\beta=0.27, \, \sigma=1/2, \, \gamma=1/7, \, \rho=1/90, \, \lambda_{FP} = 0.1, \, \lambda_{FN} = 0.3$.
    Light gray: 
    estimates from individual runs.
    Dark solid: mean 
    estimate averaged over 100 independent runs.
    Dark dashed: true parameter values.
    }
    \label{fig:alg12:3}
\end{figure*}

\begin{figure*}[htbp]
    \centering
    \includegraphics[width=\linewidth]{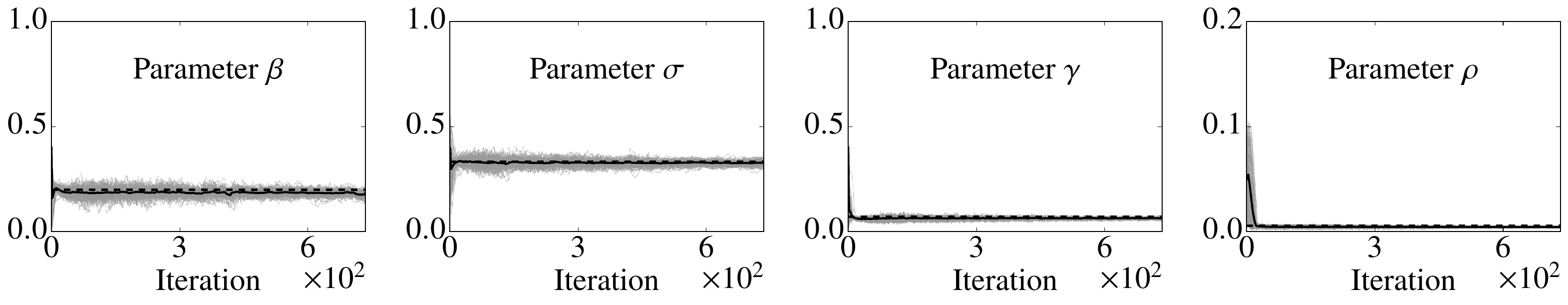}
    \caption{
    Parameter estimation of an SEIRS epidemic model for the {\tt AS-733} dynamic contact network using Algorithms~\ref{algo:fcvf} and \ref{algo:spf} for 100 independent runs in which the disease does not die out in 733 time steps.
    The parameters 
    are $\beta=0.2, \, \sigma=1/3, \, \gamma=1/14, \, \rho=1/180, \, \lambda_{FP} = 0.1, \, \lambda_{FN} = 0.1$.
    Light gray: 
    estimates from individual runs.
    Dark solid: mean 
    estimate averaged over 100 independent runs.
    Dark dashed: true parameter values.
    }
    \label{fig:alg12:4}
\end{figure*}

Figure~\ref{fig:alg12:1} shows the results of tracking states, and Figures~\ref{fig:alg12:2}, \ref{fig:alg12:3}, and  \ref{fig:alg12:4}  the results of estimating parameters using factored conditional variational filtering (Algorithms~\ref{algo:fcvf} and \ref{algo:spf}) for the {\tt Facebook}, {\tt Gowalla}, and {\tt AS-733}  contact networks.
The (hyper-)parameter values for the simulation transition schema and filter transition update are $K = 3$,  $\kappa_1 = 0.2$, $\kappa_2 = 0.1$, and $M_k = 10$, for $k = 1, \ldots, L$.
The parameter value for the simulation and filter observation schemas is $m = 5$. 
The probability of a node being observed at a time step is $\alpha = 0.7$.
The number of parameter particles is $N = 300$.

%% file: related_work.tex
\section{Related Work}
\label{sec:related}

The idea of factoring, which exploited the structure of a large dynamic Bayesian network (DBN), was introduced in \cite{BoyenKoller:1998,BoyenKoller:1999} 
for efficient approximate inference.
The basic idea was to partition the dimension space to obtain tractable approximations of distributions on high-dimensional state spaces.
Later, \cite{Ng:2002} combined factoring with particle filtering to obtain a version of factored particle filters
that is similar to our approach.
The main difference between factored particle filters as presented here and the version in \cite{Ng:2002}
is that here the particle family for the entire state space is maintained only in an implicit form;
if the state distribution is needed, for example, to compute an integral, particles for the (entire) state are sampled from the implicit form.
Later work along these lines includes that in \cite{Das:2005}.
The approach here is also similar to that in \cite{djuric-bugallo, djuric-lu-bugallo}, which use the terminology {\em multiple} particle filter.
In parallel with the papers above in the artificial intelligence literature, factored particle filters were also studied in the data assimilation literature but using the term {\em local} particle filter.
See the discussion on the origins of localization in \cite{vanleeuwen}.
Particularly relevant is the paper \cite{Rebeschini:2015} in which a local particle filter, called a block particle filter, is presented that is similar, but not identical, to the factored particle filter in Section~\ref{sec:theory:factored}.
(A block is the same as a cluster.)
This paper also contains a theorem that gives an approximation error bound for the block particle filter that could be adapted to the factored particle filter here.
A recent account of the use of local particle filters for data assimilation in large-scale geophysical applications, especially numerical weather prediction, is provided in \cite{poterjoy-wicker-buehner}.

Coverage of relevant literature on parameter estimation using particle methods in state-space models can be found in \cite{andrieu2004particle, carvalho2010particle, kantas15}. 
The origins of the nested particle filtering algorithm were discussed in \cite{Crisan:2018}.
An early discussion of conditional particle filters in the artificial intelligence literature can be found in \cite{montemerlo-thrun-whittaker}.
The problem considered is that of simultaneously estimating the pose of a mobile robot and the positions of nearby people in a previously mapped environment.
The algorithm proposed to solve the problem is a conditional particle filter that in effect treats the pose of the robot as the parameter space and the position of nearby people as the state space.
(The meaning of `conditional' in \cite{montemerlo-thrun-whittaker} is not explicitly conditional, in the sense of this paper, but has a similar motivation.)
A related approach is to augment the state with the parameter and compute joint posterior distributions over the state space and parameter. 
The EKF-SLAM filter utilizes such an approach \cite{durrant2006simultaneous}. 
Conversely, if one considers the state as originally comprising of the parameter, conditional particle filtering can be viewed as hierarchically decomposing the state into two components. 
This is the approach taken in \cite{doucet-johansen}.
There are also other meanings of the term conditional particle filter in the literature.
For example, in \cite{svensson-schon-kok}, the conditioning is with respect to a state space trajectory. A variational approach to the problem of state and parameter estimation has also been considered in \cite{ye15}. 

Whilst a wide array of exact and approximate approaches for conditional filtering and factored filtering have been explored in the literature, the combination of the two has been relatively unexplored. 
A recursive maximum-likelihood algorithm to tackle the problem of sensor registration and localization is proposed in \cite{kantas06} . 
Each sensor is represented as a node in a graph and maintains its own coordinate system with a matrix of parameters denoting the offset of a given node in another node's coordinate system. 
The parameters are estimated via gradient ascent on the log-likelihood function and each node maintains its own state filter. 
A crucial difference to our approach is that arbitrary clustering is only allowed via message passing in the observation update whilst a node's transition update depends only upon its previous state, independent of neighbours. 
In this manner, our approach is strictly more general.

%% file: conclusion.tex
\section{Conclusion}
\label{sec:conclusion}

In this paper, we outlined a theory of empirical beliefs which provides a suitable framework for belief acquisition.
Taking the viewpoint of belief acquisition as stochastic filtering, we presented a framework for investigating the space of filtering algorithms.
In particular, we presented the factored conditional filter, a filtering algorithm having three versions, for simultaneously tracking states and estimating parameters in high-dimensional state spaces.
The conditional nature of the algorithm is used to estimate parameters and the factored nature is used to decompose the state space into low-dimensional subspaces in such a way that filtering on these subspaces gives distributions whose product is a good approximation to the distribution on the entire state space.
We provided experimental results on tracking epidemics and estimating parameters in large contact networks that show the effectiveness of our approach.

There are several directions for future work.
First, the three variations of the factored conditional filter we have presented are essentially the simplest possible algorithms that are both factored and conditional.
It is likely that there are numerous improvements and optimizations that could be made to them.
Typical such elaborations for a local particle filter intended for applications in the field of geophysics and similar in nature to the factored particle filter here are given in \cite{poterjoy-wicker-buehner}.
While that algorithm is specialized to numerical weather prediction, it seems likely that many of the improvements and optimizations in 
\cite{poterjoy-wicker-buehner} will also apply to the algorithm here.

Second, the filters introduced here need further theoretical foundations.
Convergence theorems for their nested particle filter are given in \cite{Crisan:2017,Crisan:2018}.
Also an error approximation bound for their local particle filter is given in \cite{rebeschini, Rebeschini:2015} . 
These results could be the basis of similar results for the filters of this paper.
However, a different approach is required, since a key idea of this paper is that of separating nonconditional filters and conditional ones, which leads to 36 paired filters.
Thus convergence/approximation results first need to be suitably formulated and proved for conditional filters.
Then a methodology needs to be developed for combining the convergence/approximation results for each filter $F_\nu$ and $F_\mu$ in a pair of filters to obtain convergence/approximation results for the combination of $F_\nu$ and $F_\mu$ for the computation of, say, $(\nu_n \otimes \mu_n)(h_n)$ and $(\nu_n \odot \mu_n)(h_n)$.

Third, for a particular application, the optimal partition for factored filters
lies somewhere between the finest and the coarsest partitions.
It would be useful to develop criteria for choosing the appropriate partition.
The partition cannot be finer than the observations allow, but there is scope for combining the smallest clusters allowed by the observations into larger clusters thus producing a coarser partition that may reduce the approximation error.
On the other hand, a partition that is too coarse may lead to degeneracy problems.

Fourth, the experiments presented here are concerned only with the special case of state distributions (under the Markov assumption) and do not exploit the full generality of the empirical belief framework. 
It would be interesting to explore, for example, applications of a cognitive nature such as acquiring beliefs about the beliefs of other agents or applications where (a suitable summarization of) the history argument in the transition and/or observation schema was needed.

Fifth, the variational versions of the algorithms could be developed more deeply.
For example, it would be interesting to study a version of Algorithm~\ref{algo:svf} that employed variational inference, expectation propagation, or similar in the variational step and compare the results with those obtained by Algorithm~\ref{algo:fvf} for which factorization is handled outside the variational step.
For this, the ideas in \cite{MinkaDivergence} are likely to be useful.
More generally, there is promise in developing variational filters instead of the more usual particle filters for appropriate applications.

Sixth, the six conditional filters in Table~\ref{tab:filter_algorithms} all depend on the assumption that, almost surely, the stochastic process on $X$ produces the same value in $X$ at each time step. 
More generally, the stochastic process on $X$ can be arbitrary. 
For this situation, the transition schema for $F_\mu$ is more complicated having an additional $X$ argument;
intuitively, there is an $X$ argument for the current time and another $X$ argument for the previous time.
In this more general case, there are six more filters, one corresponding to each of the six conditional filters in Table~\ref{tab:filter_algorithms}.
The theory of this general case is presented in \cite{lloyd-empirical-beliefs}.
It would be interesting to investigate the practical application of these general filters.

Finally, the practical application of standalone conditional filters needs investigation.
In this case, there is no paired nonconditional filter 
and the problem becomes one of defining a conditional empirical belief on the whole of its domain $X$.
If $X$ is not a Euclidean space, a natural solution is for conditional empirical beliefs to be piecewise-constant functions.
Thus it is necessary to find a satisfactory way to partition $X$ so that a conditional empirical belief is constant on each equivalence class in the partition.
For this purpose, the concept of a predicate rewrite system in \cite{LogicforLearning, lloyd-empirical-beliefs} that can be used to generate candidate partitions is relevant.

%% file: filters_cmp.tex
\section{Empirical Comparison with Related Filtering Algorithms}
\label{appendixA}

We empirically compare the pairing of Algorithm~\ref{algo:spf} (using observation schema synthesis and a jittering transition schema) and Algorithm~\ref{algo:fcpf} for learning parameters in high-dimensional state spaces with two closely related filtering approaches: the blocked particle filter~\cite{Finke:2017} and the nested particle filter~\cite{Crisan:2018}.

The first application is a linear Gaussian model evaluated by~\cite{Finke:2017} where the state of the stochastic system at time $n$ is represented by a $V$-dimensional random vector $y_n \in \R^V$.
Let $\normal(\mu, \Sigma)$ be the multivariate Gaussian distribution with mean vector $\mu$ and covariance matrix $\Sigma$, and $a_0, a_1, \sigma_0, \sigma_1 \in \R, \sigma_0, \sigma_1 > 0$ the parameters of the system, the transition schema is
\begin{equation}
\label{appendix:eq:lg:transition}
\begin{aligned}
y_{n+1} \sim \normal(A y_n, \sigma_0^2 I_V),
\end{aligned}
\end{equation}
where $I_V$ is the identity matrix of size $V$,
and $A \in \R^{V \times V}$ is a tridiagonal matrix with entries
\begin{equation}
\begin{aligned}
A_{i,j} = \begin{cases}
a_0, & \text{if } i = j, \\
a_1, & \text{if } j \in \{i-1, \, i+1\}, \\
0,   & \text{otherwise}.
\end{cases}
\end{aligned}
\end{equation}

A noisy measurement of the state at time $n$ is produced by the observation schema
\begin{equation}
\label{appendix:eq:lg:observe}
o_n \sim \normal(y_n, \sigma_1^2 I_V).
\end{equation}

In the experiment, we follow~\cite{Finke:2017} and use the true
parameter values
$a_0 = 0.5, \, a_1 = 0.2, \, \sigma_0 = \sigma_1 = 1$.
The state dimension $V = 100$, and the initial state of the system $y_\ast = \mathbf{1}_V$ is a $V$-dimensional vector with all entries equal to $1$.

\begin{figure*}[!t]
    \centering
    \setlength{\tabcolsep}{.08in}
    \vspace{-1em}
    \begin{tabular}{l}
    {\hspace{-2pt}}\includegraphics[height=1.125in]{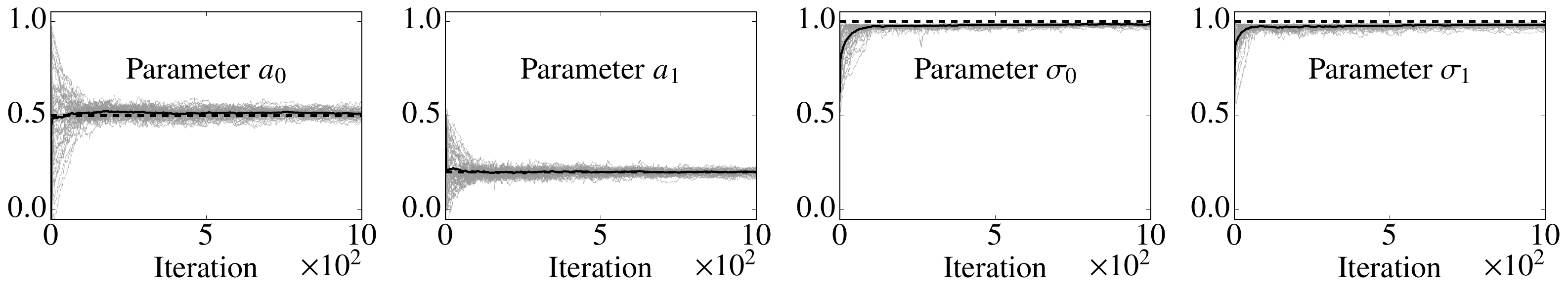} \\
    {\hspace{115pt}}(a) Estimating parameters.{\vspace{10pt}} \\
    {\hspace{-7pt}}\includegraphics[height=1.16in]{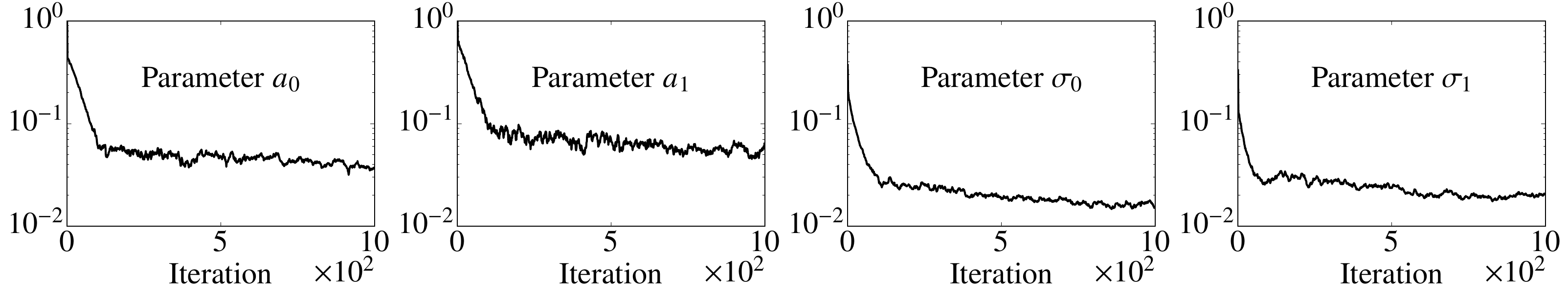} \\
    {\hspace{115pt}}(b) Normalized parameter estimation errors.
    \end{tabular}
    \caption{
    Estimating 
    parameters of a linear Gaussian system~\cite{Finke:2017} with Algorithms~\ref{algo:spf} and \ref{algo:fcpf} in 50 independent runs.
    The number of particles employed are $N = 300, M = 30$. 
    (a): Each of the four parameters estimated from individual runs (light gray), mean estimate averaged over 50 independent runs (dark solid), and the true parameter values (dark dashed).
    (b): The normalized error of each parameter estimation (Eq.~\ref{appendix:eq:error:param}) averaged over the same 50 independent runs.
    }
    \label{appendix:fig:lg:param}
\end{figure*}

To estimate the parameters $a_0, a_1, \sigma_0, \sigma_1$ using Algorithms~\ref{algo:spf} and~\ref{algo:fcpf}, the initial conditional particle family $((x_0^{(i)}, (y_0^{(i,j)})_{j=1}^M))_{i=1}^N$ are produced by
\begin{equation}
\begin{aligned}
& x_{0, 1}^{(i)} \sim \UCal(-1, 1),  \ \ \
x_{0, 2}^{(i)} \sim \UCal(-1, 1), \ \ \ 
x_{0, 3}^{(i)} \sim \UCal(0, 1), \ \ \
x_{0, 4}^{(i)} \sim \UCal(0, 1), \\
& y_0^{(i,j)} \sim \normal(\mathbf{1}_V, I_V),
\text{ for } i = 1, \dots, N \text{ and } j = 1,\dots,M,
\end{aligned}
\end{equation}
where $\UCal(a, b)$ is the uniform distribution over interval $(a, b)$, 
and $N = 300$ and $M = 30$ specify the number of parameter particles and conditional particles respectively.

Further, an adaptive Gaussian jittering kernel is employed in Algorithm~\ref{algo:spf} with covariance matrix
\begin{equation}
\Sigma_n = \max(a r^n, b) \diag(1, 1, 1, 1), \ n \in \N,
\end{equation}
where $a = 10^{-4}, \, b = 9 \times 10^{-6}, \, r = 0.996$, and $\diag(\cdot)$ is a diagonal matrix with the given diagonal entries. 

Figure~\ref{appendix:fig:lg:param} shows the results of estimating parameters for the linear Gaussian system using Algorithms~\ref{algo:spf} and~\ref{algo:fcpf}.
We performed 50 independent runs of the experiment to evaluate the expected behavior.
Following~\cite{Crisan:2018}, we also evaluated each of the estimated parameters at time step $n$ using the normalized error 
\begin{equation}
\label{appendix:eq:error:param}
| (\widehat{x}_{n,k} - x_k) / x_k |, \text{ for all }  n \in \N \text{ and }  k = 1, 2, 3, 4,
\end{equation}
where the estimate $\widehat{x}_{n,k} = \frac{1}{N} \sum_{i=1}^N x^{(i)}_{n,k}$, for $k = 1,2,3,4$, and the true parameter values $x = (0.5, 0.2, 1.0, 1.0)$.
The experimental results in Figure~\ref{appendix:fig:lg:param}
are comparable to those of the blocked particle filters presented in~\cite{Finke:2017}.

The second application here is that of a stochastic discrete-time Lorenz system studied in detail in~\cite{Chorin:2004, Crisan:2018}.
Usually, factorization is not necessary for filters to work in a low-dimensional state space as for this application, but to validate the factored conditional particle filter, here we employ Algorithms~\ref{algo:spf} and \ref{algo:fcpf} to track states and estimate parameters.
The state of the Lorenz system at time $n$ is represented by a three-dimensional random vector $y_n = (y_{n,1}, \, y_{n,2}, \, y_{n,3})$,
which are governed by the initial states $y_\ast = (-5.91652, -5.52332, \, 24.5723)$ and the following stochastic transition schema
\begin{equation}
\label{appendix:eq:lorenz:transition}
\begin{aligned}
& y_{n+1,1} \sim \normal(y_{n,1} - \Delta t \, \theta_1 (y_{n,1} - y_{n,2}), \, \Delta t), \\
& y_{n+1,2} \sim \normal(y_{n,2} + \Delta t \, (\theta_2 y_{n,1} - y_{n,2} - y_{n,1} y_{n,3}), \, \Delta t), \\
& y_{n+1,3} \sim \normal(y_{n,3} + \Delta t \, (y_{n,1} y_{n,2} - \theta_3 y_{n,3}), \, \Delta t), \text{ for all $n \in \N_0$},
\end{aligned}
\end{equation}
where $\normal(\mu, \sigma^2)$ denotes the Gaussian distribution with mean $\mu$ and variance $\sigma^2$. 
The step size $\Delta t = 0.001$ and the true values of parameters $\theta_1 = 10, \ \theta_2 = 28, \ \theta_3 = 8/3$.

Further, the state is partially observed with Gaussian noise every $40$ steps,
\begin{equation}
\label{appendix:eq:lorenz:obs}
\begin{aligned}
& o_{n,1} \sim \normal(\theta_4 y_{n,1}, \, 0.1), \
& o_{n,3} \sim \normal(\theta_4 y_{n,3}, \, 0.1),
\end{aligned}
\end{equation}
for all $n \in \{40t: t \in \N \}$,
and the true value of the scale parameter 
$\theta_4 = 4/5$.
This observation schema is factorizable and thus Algorithm~\ref{algo:fcpf} is applicable. 
(The missing observation schema for the second coordinate can simply be 
ignored by the filters.)

A similar setup to that presented in~\cite{Crisan:2018} was adopted.
First, the initial conditional particle family $((x_0^{(i)}, (y_0^{(i,j)})_{j=1}^M))_{i=1}^N$ are produced by
\begin{equation}
\begin{aligned}
\label{appendix:eq:lorenz:init}
& x_{0,1}^{(i)} \sim \UCal(5, 20), \ \ \ x_{0,2}^{(i)} \sim \UCal(18, 50), \ \ \ 
  x_{0,3}^{(i)} \sim \UCal(1, 8),  \ \ \ x_{0,4}^{(i)} \sim \UCal(0.5, 3), \\
& y_0^{(i, j)} \sim \NCal(y_\ast, 10 I_3), \text{ for } i = 1, \dots, N \text{ and } j = 1,\dots,M,
\end{aligned}
\end{equation}
where 
$N = 300$ and $M = 50$ are the number of parameter and conditional particles respectively.

A Gaussian jittering kernel was adopted with the following covariance matrix, 
\begin{equation}
\Sigma_n = \max(a r^n, b) \, \diag(15^2, 32^2, 7^2, 2.5^2), \text{ for all }  n \in \N,
\end{equation}
where $a = 4 \times 10^{-4}, \, b = 4 \times 10^{-6}, \, r = 0.996$. Note that the diagonal entries of $\Sigma_n$ are derived from the ranges of the parameters in Eq.~\eqref{appendix:eq:lorenz:init}.

\begin{figure*}[!t]
    \centering
    \setlength{\tabcolsep}{.08in}
    \vspace{-1em}
    \begin{tabular}{l}
    {\hspace{0pt}}\includegraphics[height=1.16in]{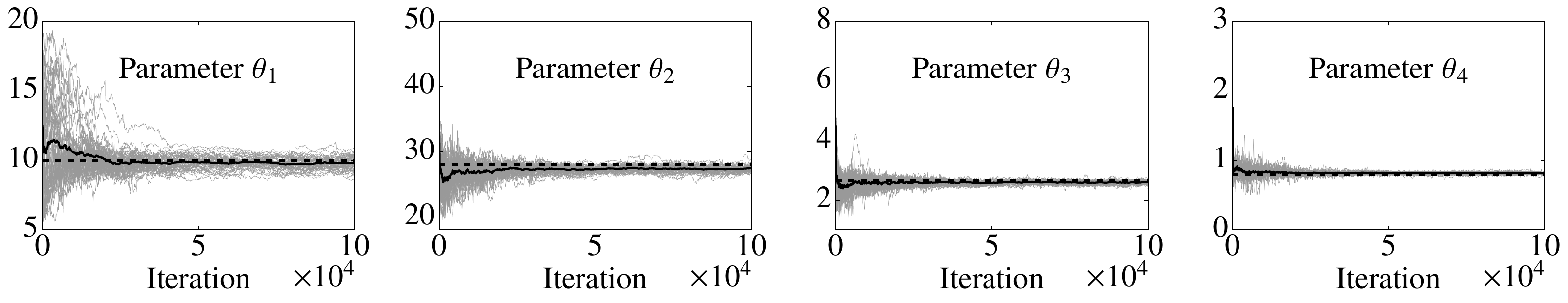} \\
    {\hspace{115pt}}(a) Estimating parameters.{\vspace{10pt}} \\
    {\hspace{-7pt}}\includegraphics[height=1.16in]{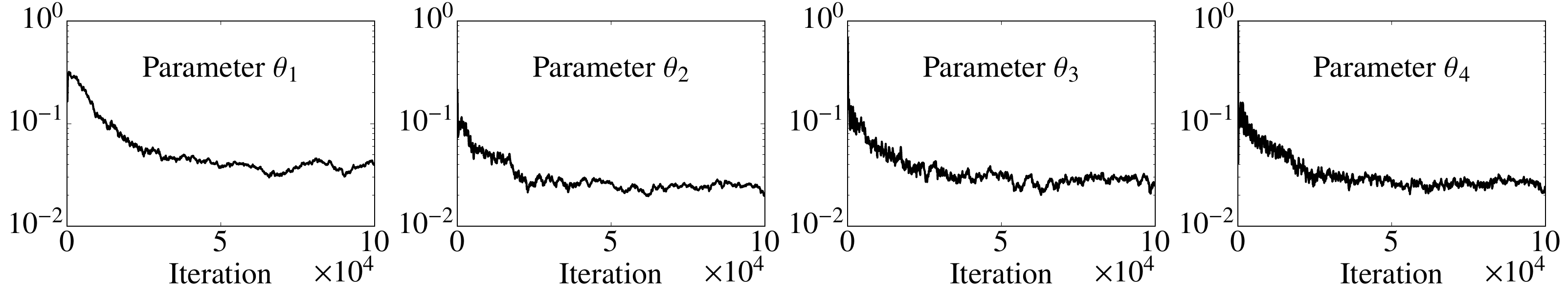} \\
    {\hspace{115pt}}(b) Normalized parameter estimation errors.
    \end{tabular}
    \caption{
    Estimating each of the four parameters of a discrete time Lorenz system with Algorithms~\ref{algo:spf} and \ref{algo:fcpf} in 50 independent runs.
    The number of particles employed are $N = 300, M = 50$. 
    (a): Each of the four parameters estimated from individual runs (light gray), mean estimate averaged over 50 independent runs (dark solid), and the true parameter values (dark dashed).
    (b): The normalized error of each parameter estimation (Eq.~\ref{appendix:eq:error:param}) averaged over the same 50 independent runs.
    }
    \label{appendix:fig:lorenz:param}
\end{figure*}

Fifty independent runs of the experiment were performed to evaluate the expected behavior, and the results of estimating parameters for the Lorenz system using Algorithms~\ref{algo:spf} and~\ref{algo:fcpf} are presented in Figure~\ref{appendix:fig:lorenz:param}.

These results can be compared with those of the nested particle filter in~\cite{Crisan:2018}.
The principles underlying this filter can be found in \cite{Johansen}.
The nested particle filter has a hierarchical structure with an outer filter for parameters and a set of inner filters for states, one for each sample generated by the outer filter.
The nested particle filter first updates the state particle family for each parameter particle and then updates the parameter particles, while Algorithms~\ref{algo:spf} and~\ref{algo:fcpf} update the parameter particles before updating the state particle family corresponding to each parameter particle.
Furthermore, the nested particle filter handles resampling of parameters differently in that conditional particles that consist of both a parameter particle and a state particle family are resampled based on the weights of the parameter particles alone.
Finally, Algorithm~\ref{algo:fcpf} is factored while the nested particle filter is not.

In spite of the differences between the algorithms, the respective results are quite similar.
The accuracy of the parameter estimation from Algorithms~\ref{algo:spf} and \ref{algo:fcpf} is only slightly worse than that of the nested particle filter which employs a larger number of particles ($N=M=300$).
This is likely to be a result of the approximation introduced by the factored nature of Algorithm~\ref{algo:fcpf}.

For another comparison, we also experimented with the pairing of Algorithms~\ref{algo:spf}~and~\ref{algo:cpf}.
The differences between the nested particle filter and Algorithms~\ref{algo:spf} and~\ref{algo:cpf} are the same as for Algorithms~\ref{algo:spf} and \ref{algo:fcpf}, except that Algorithm~\ref{algo:cpf} is not factored.
Algorithms~\ref{algo:spf} and~\ref{algo:cpf} gave an accuracy of parameter estimation that corresponded closely to that of the nested particle filter. 
We refer the reader to the GitHub repository for these results.

%% file: transition_observation_update.tex
\section{Derivation of Transition Update}
\label{appendixB}

Here is the derivation of the transition update 
\[
\overline{q}_n^{(k)}(x)(\sS) = \rho \, q_{n-1}^{(k)}(x)(\sR) + \left( \textstyle\prod_{l \in \mathcal{N}_k} (1 - \beta q_{n-1}^{(l)}(x)(\sI)) \right)  q_{n-1}^{(k)}(x)(\sS),
\]
for all $x \in X$ and $n \in \N$, and $k = 1,\dots,L$,
for compartment $\sS$ employed by the factored conditional filter in Section~\ref{sec:fcf}.
The derivations for the other compartments, $\mathsf{E}$, $\mathsf{I}$, and $\mathsf{R}$, are similar.

Let $r_1, \ldots, r_q$ be the indices of the nodes that are the neighbours of node $k$, so that $\mathcal{N}_k = \{ r_1, \ldots, r_q \}$.
Then, for all $n \in \N$,
\begin{align*}
  & \; \overline{q}_n^{(k)}(x)(\mathsf{S}) \\
= & \; \sum_{s \in C^L}  \breve{\tau}_n^{(k)}(x, s)(\mathsf{S}) \; \prod_{p=1}^L q_{n-1}^{(p)}(x)(s_p) \\
= & \; \sum_{\substack{s \in C^L \\ s_k = \mathsf{R}}}  \breve{\tau}_n^{(k)}(x, s)(\mathsf{S}) \; \prod_{p=1}^L q_{n-1}^{(p)}(x)(s_p) +
        \sum_{\substack{s \in C^L \\ s_k = \mathsf{S}}}  \breve{\tau}_n^{(k)}(x, s)(\mathsf{S}) \; \prod_{p=1}^L q_{n-1}^{(p)}(x)(s_p) \\
= & \; \rho \sum_{\substack{s \in C^L \\ s_k = \mathsf{R}}}  \prod_{p=1}^L q_{n-1}^{(p)}(x)(s_p) +
        \sum_{\substack{s \in C^L \\ s_k = \mathsf{S}}}  \breve{\tau}_n^{(k)}(x, s)(\mathsf{S}) \; \prod_{p=1}^L q_{n-1}^{(p)}(x)(s_p) \\
= & \; \rho \, q_{n-1}^{(k)}(x)(\mathsf{R}) \, \sum_{\substack{s \in C^L \\ s_k = \mathsf{R}}}  \prod_{\substack{p=1, \ldots, L \\ p \neq k}} q_{n-1}^{(p)}(x)(s_p)+
        \sum_{\substack{s \in C^L \\ s_k = \mathsf{S}}}  \breve{\tau}_n^{(k)}(x, s)(\mathsf{S}) \; \prod_{p=1}^L q_{n-1}^{(p)}(x)(s_p) \\
= & \; \rho \, q_{n-1}^{(k)}(x)(\mathsf{R}) +
        \sum_{\substack{s \in C^L \\ s_k = \mathsf{S}}}  \breve{\tau}_n^{(k)}(x, s)(\mathsf{S}) \; \prod_{p=1}^L q_{n-1}^{(p)}(x)(s_p) \\
= & \; \rho \, q_{n-1}^{(k)}(x)(\mathsf{R}) +
        \sum_{\substack{s \in C^L \\ s_k = \mathsf{S}}}  (1 - \beta)^{d_k(s)} \; \prod_{p=1}^L q_{n-1}^{(p)}(x)(s_p) \\
= & \; \rho \, q_{n-1}^{(k)}(x)(\mathsf{R}) +
        \left( \sum_{\substack{s \in C^L \\ s_k = \mathsf{S}}}  (1 - \beta)^{d_k(s)} \; \prod_{\substack{p=1, \ldots, L \\ p \neq k}} q_{n-1}^{(p)}(x)(s_p) \right) \, q_{n-1}^{(k)}(x)(\mathsf{S}) \\
= & \; \rho \, q_{n-1}^{(k)}(x)(\mathsf{R}) + 
        \left( \sum_{(c_1, \ldots, c_q) \in C^{|\mathcal{N}_k|}} (1 - \beta)^{\sum_{l=1}^q \mathbb{I}(c_l = \mathsf{I})} \prod_{l=1}^q q_{n-1}^{(r_l)}(x)(c_l) \right)  \, q_{n-1}^{(k)}(x)(\mathsf{S}) \\
= & \; \rho \, q_{n-1}^{(k)}(x)(\mathsf{R}) + 
        \left( \sum_{(c_1, \ldots, c_q) \in C^{|\mathcal{N}_k|}} \prod_{l=1}^q (1 - \beta)^{\mathbb{I}(c_l = \mathsf{I})} \, q_{n-1}^{(r_l)}(x)(c_l) \right)  \, q_{n-1}^{(k)}(x)(\mathsf{S}) \\
= & \; \rho \, q_{n-1}^{(k)}(x)(\mathsf{R}) + 
        \left( \prod_{l=1}^q \sum_{c_l \in C} (1 - \beta)^{\mathbb{I}(c_l = \mathsf{I})} \, q_{n-1}^{(r_l)}(x)(c_l) \right)  \, q_{n-1}^{(k)}(x)(\mathsf{S}) \\
= & \; \rho \, q_{n-1}^{(k)}(x)(\mathsf{R}) + 
        \left( \prod_{l=1}^q (1 - \beta q_{n-1}^{(r_l)}(x)(\mathsf{I})) \right)  \, q_{n-1}^{(k)}(x)(\mathsf{S}) \\  
= & \; \rho \, q_{n-1}^{(k)}(x)(\mathsf{R}) + 
        \left( \prod_{l \in \mathcal{N}_k} (1 - \beta q_{n-1}^{(l)}(x)(\mathsf{I})) \right)  \, q_{n-1}^{(k)}(x)(\mathsf{S}).
\end{align*}

%% file: KL_derivation.tex
\section{Derivation of KL Divergence}
\label{appendixC}

Here is the derivation of the expression for the KL divergence from Section~\ref{sec:fcvf}.

Let 
$M_i \triangleq \int_S  \lambda y. \frac{\Gamma(m + 1)}{\prod_{j=1}^4 \Gamma(o_j + 1)} \; \frac{\Gamma(K)}{\prod_{j=1}^4 \Gamma(K\beta_j^{(i,k)})} \;            \; \prod_{j=1}^4 y_j^{o_j + K\beta_j^{(i,k)} - 1} \; d \upsilon_S$, for $i = 1, \ldots, N$.
Then, for $i = 1, \ldots, N$,
\[ M_i =  \frac{\Gamma(m + 1)}{\prod_{j=1}^4 \Gamma(o_j + 1)} \; \frac{\Gamma(K)}{\prod_{j=1}^4 \Gamma(K\beta_j^{(i,k)})} 
                     \; \frac{\prod_{j=1}^k \Gamma(o_j + K \beta_j^{(i,k)})}{\Gamma(m + K)}. \]
                     
Suppose that $r$ has the form $\mathit{Dir}(\gamma_1^{(i,k)}, \gamma_2^{(i,k)}, \gamma_3^{(i,k)}, \gamma_4^{(i,k)})$.
Note that $\log f$, for some function $f$, below means the composition $\log \comp f$.
Then, for $i = 1, \ldots, N$,
\begin{align*}
  & \; KL(p_n^{(k)}(x_n^{(i)}) \| r) \\
= & \; \int_S p_n^{(k)}(x_n^{(i)}) \log \frac{p_n^{(k)}(x_n^{(i)})}{\mathit{Dir}(\gamma_1^{(i,k)}, \gamma_2^{(i,k)}, \gamma_3^{(i,k)}, \gamma_4^{(i,k)})} \; d \upsilon_S \\
= & \; \mathit{Const} - \int_S p_n^{(k)}(x_n^{(i)}) \log \mathit{Dir}(\gamma_1^{(i,k)}, \gamma_2^{(i,k)}, \gamma_3^{(i,k)}, \gamma_4^{(i,k)}) \; d \upsilon_S \\
= & \; \mathit{Const} - M_i^{-1} \int_S \lambda y. \frac{\Gamma(m + 1)}{\prod_{j=1}^4 \Gamma(o_j + 1)} \; \frac{\Gamma(K)}{\prod_{j=1}^4 \Gamma(K\beta_j^{(i,k)})} 
                      \prod_{j=1}^4 y_j^{o_j + K\beta_j^{(i,k)} - 1}
 \log \mathit{Dir}(\gamma_1^{(i,k)}, \gamma_2^{(i,k)}, \gamma_3^{(i,k)}, \gamma_4^{(i,k)}) \; d \upsilon_S \\
= & \; \mathit{Const} - M_i^{-1} \int_S \lambda y. \frac{\Gamma(m + 1)}{\prod_{j=1}^4 \Gamma(o_j + 1)} \; \frac{\Gamma(K)}{\prod_{j=1}^4 \Gamma(K\beta_j^{(i,k)})}  
              \; \prod_{j=1}^4 y_j^{o_j + K\beta_j^{(i,k)} - 1} \\*
  & \hspace*{9em}  \left (\log \Gamma (\sum_{j=1}^4 \gamma_j^{(i,k)}) - \log \prod_{j=1}^4 \Gamma (\gamma_j^{(i,k)}) + \lambda y. \sum_{j=1}^4 (\gamma_j^{(i,k)} - 1) \log y_j \right) \; d \upsilon_S \\
= & \; \mathit{Const}  - \left( \left(\log \Gamma (\sum_{j=1}^4 \gamma_j^{(i,k)}) - \log \prod_{j=1}^4 \Gamma (\gamma_j^{(i,k)}) \right) \; + \right. \\*
  & \hspace*{1em} M_i^{-1} \int_S \lambda y. \frac{\Gamma(m + 1)}{\prod_{j=1}^4 \Gamma(o_j + 1)} \; \frac{\Gamma(K)}{\prod_{j=1}^4 \Gamma(K\beta_j^{(i,k)})}  
              \;  \prod_{j=1}^4 y_j^{o_j + K\beta_j^{(i,k)} - 1}
    \left. \lambda y. \sum_{j=1}^4 (\gamma_j^{(i,k)} - 1) \log y_j \; d \upsilon_S \right) \\
= & \; \mathit{Const}  - \left( \left(\log \Gamma (\sum_{j=1}^4 \gamma_j^{(i,k)}) - \log \prod_{j=1}^4 \Gamma (\gamma_j^{(i,k)}) \right) \; +
  M_i^{-1}  \frac{\Gamma(m + 1)}{\prod_{j=1}^4 \Gamma(o_j + 1)} \; \frac{\Gamma(K)}{\prod_{j=1}^4 \Gamma(K\beta_j^{(i,k)})}  
              \;  \frac{\prod_{j=1}^4 \Gamma(o_j +K \beta_j^{(i,k)})}{\Gamma(m + K)} \right. \\*
  & \hspace*{1em}   \left. \int_S \lambda y. \sum_{j=1}^4 (\gamma_j^{(i,k)} - 1) \log y_j \; \mathit{Dir}(o_1 + K \beta_1^{(i,k)}, o_2 + K \beta_2^{(i,k)}, o_3 + K\beta_3^{(i,k)}, o_4 + K\beta_4^{(i,k)}) \; d \upsilon_S \right) \\
= & \; \mathit{Const}  - \left( \left(\log \Gamma (\sum_{j=1}^4 \gamma_j^{(i,k)}) - \log \prod_{j=1}^4 \Gamma (\gamma_j^{(i,k)}) \right) \; +
  \sum_{j=1}^4 (\gamma_j^{(i,k)} - 1) (\psi(o_j + K \beta_j^{(i,k)}) - \psi(\sum_{p=1}^4 (o_p + K \beta_p^{(i,k)})) \right) \\
= & \; \mathit{Const}  - \left( \left(\log \Gamma(\sum_{j=1}^4 \gamma_j^{(i,k)}) - \log \prod_{j=1}^4 \Gamma (\gamma_j^{(i,k)}) \right) \; +
  \sum_{j=1}^4 (\gamma_j^{(i,k)} - 1) (\psi(o_j + K \beta_j^{(i,k)}) - \psi(m + K)) \right),
\end{align*}
where $\psi$ is the digamma function.

%% file: performance_evaluation.tex
\section{Evaluating the Performance of Filters}
\label{appendixD}

This appendix presents the metrics that are used to evaluate the performance of the filters studied in this paper.

The first issue is how well a filter tracks the state.
Consider first the nonconditional case. 
Suppose that $Y$ is the state space for both the simulation and the filter.
Also suppose that $\widetilde{y}_n \in Y$ is the ground truth state at time $n$ given by the simulation.
(Strictly, $\widetilde{y} : \Omega \rightarrow Y^{\mathbb{N}_0}$ is a stochastic process, so that each 
$\widetilde{y}_n : \Omega \rightarrow Y$ is a random variable.)
Let $(\mu_n : H_n \rightarrow \mathcal{P}(Y))_{n \in \mathbb{N}_0}$ be the relevant empirical schema so that $\mu_n(h_n) : \mathcal{P}(Y)$ is the empirical belief at time $n$.
Let $\reallywidehat{\mu_n(h_n)} : \mathcal{P}(Y)$ be the approximation of $\mu_n(h_n)$ given by the filter at time $n$.
Intuitively, we need a suitable measure of the `distance' between the state $\widetilde{y}_n \in Y$ and  the distribution $\reallywidehat{\mu_n(h_n)} \in \mathcal{P}(Y)$.
Let $\rho : Y \times Y \rightarrow \mathbb{R}$ be a metric on $Y$.
Then let
\begin{equation}
\label{eq:error_noncond}
\mathit{Err}(\reallywidehat{\mu_n(h_n)}) \triangleq \int_Y  \lambda y. \rho(\widetilde{y}_n, y)  \; d \reallywidehat{\mu_n(h_n)}.
\end{equation}
Thus $\mathit{Err}(\reallywidehat{\mu_n(h_n)})$ is a random variable whose value is the average error using the metric $\rho$ with respect to  the filter's estimate $\reallywidehat{\mu_n(h_n)}$ of the ground truth state $\widetilde{y}_n$ at time $n$.

For the conditional case, there is an empirical schema $(\nu_n : H_n \rightarrow \mathcal{P}(X))_{n \in \mathbb{N}_0}$ for parameters and an empirical schema 
$(\mu_n : H_n \times X \rightarrow \mathcal{P}(Y))_{n \in \mathbb{N}_0}$ for states conditional on the parameters.
It can be shown that $(\nu_n \odot \mu_n : H_n \rightarrow \mathcal{P}(Y))_{n \in \mathbb{N}_0}$ is an empirical schema for states.
Let $\reallywidehat{(\nu_n \odot \mu_n)(h_n)} \in \mathcal{P}(Y)$ be the approximation of $(\nu_n \odot \mu_n)(h_n)$ given by the filter at time $n$. 
In this case, let
\begin{equation}
\label{eq:error_cond}
\mathit{Err}((\reallywidehat{\nu_n \odot \mu_n)(h_n)}) \triangleq \int_Y  \lambda y. \rho(\widetilde{y}_n, y)  \; d \reallywidehat{(\nu_n \odot \mu_n)(h_n)}. 
\end{equation}
In both the nonconditional and conditional cases, the average is with respect to the filter's estimate of the empirical belief on $Y$.
Both $\mathit{Err}(\reallywidehat{\mu_n(h_n)})$ and $\mathit{Err}(\reallywidehat{(\nu_n \odot \mu_n)(h_n)})$ are referred to as the {\em state error} (at time $n$).

An alternative approach to the above is to identify the ground truth state with the Dirac measure at that state. 
In this case, we require instead the definition of a distance between probability measures.
The most natural definition of such a distance is the total variation metric that produces a value in the range $[0, 1]$.
However only in exceptional circumstances is the total variation metric computable, although it may be possible in some applications to compute reasonably accurate bounds on the value of this metric.
This approach is not pursued here.
Also note that there is a fundamental difference between the simulation case considered here and a similar problem of defining a metric that arises in convergence theorems for filters.
For convergence theorems, one also requires a metric that defines the distance between two probability measures.
One is the conditional probability defined by the underlying stochastic process of the state distribution given a particular history.
The other is the approximation of this distribution for the same history given by the filtering algorithm employed.
In this case, the total variation metric is a natural metric for defining the distance between these two probability measures, but the fact that this metric is not computable is of little consequence for convergence theorems.

Often $Y$ is a product space $\prod_{l=1}^p Y_l$.
In this case, let $\rho_l : Y_l \times Y_l \rightarrow \mathbb{R}$ be a metric on $Y_l$, for $l = 1, \ldots, p$.
Then the metric $\rho : \prod_{l=1}^p Y_l \times \prod_{l=1}^p Y_l \rightarrow \mathbb{R}$ can be defined by
\[ \rho(y, z) = \frac{1}{p} \sum_{l=1}^p \rho_l(y_l, z_l), \] 
for all $y,z \in \prod_{l=1}^p Y_l$.
The constant $1/p$ is introduced to make the error less dependent on the size of the dimension $p$.
Of course, there are many other ways of defining $\rho$ that could be used instead -- the definition here is convenient for our purpose.
Suppose now that $\reallywidehat{\mu_n(h_n)}$ can be factorized so that 
$\reallywidehat{\mu_n(h_n)} = \bigotimes_{l=1}^p \reallywidehat{\mu_n^{(l)}(h_n^{(l)})}$,
where $\reallywidehat{\mu_n^{(l)}(h_n^{(l)})}$ is an approximation of the empirical belief $\mu_n^{(l)}(h_n^{(l)}) \in \mathcal{P}(Y_l)$,
for $l = 1, \ldots, p$.
Then
\begin{align*}
  & \; \int_Y  \lambda y. \rho(\widetilde{y}_n, y)  \; d \reallywidehat{\mu_n(h_n)} \\
= & \; \frac{1}{p} \int_{\prod_{l=1}^p Y_l} \lambda (y^{(1)}, \ldots, y^{(p)}). \sum_{l=1}^p \rho_l(\widetilde{y}_n^{(l)}, y^{(l)})  \; d \bigotimes_{l=1}^p \reallywidehat{\mu_n^{(l)}(h_n^{(l)})} \\
= & \; \frac{1}{p} \int_{Y_p} \left( \lambda y^{(p)}. \int_{Y_{p-1}} \left( \lambda y^{(p-1)}. \; \ldots \; \int_{Y_1} \lambda y^{(1)}. \sum_{l=1}^p \rho_l(\widetilde{y}_n^{(l)}, y^{(l)}) \; d\reallywidehat{\mu_n^{(1)}(h_n^{(1)})} \; \ldots \; \right) \; d\reallywidehat{\mu_n^{(p-1)}(h_n^{(p-1)})} \right) \; d\reallywidehat{\mu_n^{(p)}(h_n^{(p)})} \\
= & \; \frac{1}{p} \sum_{l=1}^p \int_{Y_l} \lambda y^{(l)}. \rho_l(\widetilde{y}_n^{(l)}, y^{(l)}) \; d \reallywidehat{\mu_n^{(l)}(h_n^{(l)})}.
\end{align*}
Thus 
\begin{equation} 
\label{eq:error_noncond_factored}
\mathit{Err}(\reallywidehat{\mu_n(h_n)}) = \frac{1}{p} \sum_{l=1}^p \int_{Y_l} \lambda y^{(l)}. \rho_l(\widetilde{y}_n^{(l)}, y^{(l)}) \; d \reallywidehat{\mu_n^{(l)}(h_n^{(l)})}.
\end{equation}
Intuitively, $\mathit{Err}(\reallywidehat{\mu_n(h_n)})$ is a random variable whose value is the average over all $p$ clusters of the average error using the metric $\rho_l$ with respect to the filter's estimate $\reallywidehat{\mu_n^{(l)}(h_n^{(l)})}$ of the $l$th component $\widetilde{y}_n^{(l)}$ of the ground truth state  at time $n$.
There is an analogous expression for the conditional case.

In the applications considered here, the state space is either $C^L$ or $S^L$.
For the first case, a metric on $C$ is needed and for this purpose the discrete metric is the obvious choice.
In this case, the state error is bounded above by $1$.
For $S$, the metric $d : S \times S \rightarrow \mathbb{R}$ defined by 
$d(x, y) = \sum_{j=1}^4 | x_j - y_j |$, for all $x, y \in S$, 
where $x_4 \triangleq 1 -\sum_{i=1}^3 x_i$ and $y_4 \triangleq 1 -\sum_{i=1}^3 y_i$, is employed.
The definition of the metric is independent of the choice of fill-up argument.
In this case, the state error is bounded above by $2$.

The second issue is how well the filter learns the parameters of the simulation.
Let $X = \prod_{j=1}^k X_j= \mathbb{R}^k$, for some $k \geq 1$, and $\widetilde{x} = (\widetilde{x}_1, \ldots, \widetilde{x}_k)$ be the ground truth parameter value.
Let 
\begin{equation}
\label{eq:error_parameter}
\mathit{Err}_j(\reallywidehat{\nu_n(h_n)}) \triangleq \frac{1}{\widetilde{x}_j} \int_X  \lambda x. | \widetilde{x}_j - x_j |  \; d \reallywidehat{\nu_n(h_n)}, 
\end{equation}
for $j = 1, \ldots, k$.
Thus $\mathit{Err}_j(\reallywidehat{\nu_n(h_n)})$ is a random variable whose value is the average normalized absolute error with respect to the filter's estimate $\reallywidehat{\nu_n(h_n)}$ of the ground truth parameter (component) $\widetilde{x}_j$ at time $n$, for $j = 1, \ldots, k$.
$\mathit{Err}_j(\reallywidehat{\nu_n(h_n)})$ is referred to as the {\em $j$th parameter error} (at time $n$).

In all the applications in this paper, the filter for the parameters is a particle filter.
Suppose that $(x^{(i)}_n)_{i=1}^N$ is the parameter particle family at time $n$.
Thus $\reallywidehat{\nu_n(h_n)} = \frac{1}{N} \sum_{i=1}^N \delta_{x^{(i)}_n}$ and hence
\begin{equation}
\label{eq:error_param_particle}
\mathit{Err}_j(\reallywidehat{\nu_n(h_n)}) = \frac{1}{\widetilde{x}_j} \frac{1}{N} \sum_{i=1}^N | \widetilde{x}_j - x^{(i)}_{n, j} |,
\end{equation}
for $j = 1, \ldots, k$.
The $j$th  {\em parameter estimate} (at time $n$), for $j = 1, \ldots, k$, is given by
\begin{equation}
\label{eq:estimate_parameter}
\frac{1}{N} \sum_{i=1}^N x_{n,j}^{(i)}.
\end{equation}

%% file: state_error_derivation.tex
\section{Derivation of the State Error}
\label{appendixE}

Here is the derivation of the formula for the state error $\mathit{Err}(\reallywidehat{(\nu_n \odot\mu_n)(h_n)})$ from Section~\ref{sec:fcvf}.
\begin{align*}
  & \; \mathit{Err}(\reallywidehat{(\nu_n \odot \mu_n)(h_n)}) \\
= & \; \frac{1}{L} \sum_{k=1}^L \int_S \lambda y. \rho_k(\widetilde{y}_n^{(k)}, y) \; d \reallywidehat{(\nu_n \odot \mu_n^{(k)})(h_n^{(k)})} \hspace*{3em} \text{[Conditional version of Eq.~\eqref{eq:error_noncond_factored}]} \\
= & \; \frac{1}{L} \sum_{k=1}^L \int_S \lambda y. \rho_k(\widetilde{y}_n^{(k)}, y) \; d ((\frac{1}{N} \sum_{i=1}^N \delta_{x^{(i)}_n}) \odot (q^{(k)}_n \cdot \upsilon_S)) \\
= & \; \frac{1}{L} \sum_{k=1}^L \int_S \lambda y. \rho_k(\widetilde{y}_n^{(k)}, y) \; d \frac{1}{N} \sum_{i=1}^N (\delta_{x^{(i)}_n} \odot (q^{(k)}_n \cdot \upsilon_S)) \\
= & \; \frac{1}{LN} \sum_{k=1}^L \sum_{i=1}^N \int_S \lambda y. \rho_k(\widetilde{y}_n^{(k)}, y) \; d (\delta_{x^{(i)}_n} \odot (q^{(k)}_n \cdot \upsilon_S)) \\
= & \; \frac{1}{LN} \sum_{k=1}^L \sum_{i=1}^N \int_S \lambda y. \rho_k(\widetilde{y}_n^{(k)}, y) \; d (q^{(k)}_n(x^{(i)}_n) \cdot \upsilon_S) \\
= & \; \frac{1}{LN} \sum_{k=1}^L \sum_{i=1}^N \int_S \lambda y. \rho_k(\widetilde{y}_n^{(k)}, y) \; q^{(k)}_n(x^{(i)}_n) \; d \upsilon_S \\
= & \; \frac{1}{LN} \sum_{k=1}^L \sum_{i=1}^N \int_S \lambda y. \sum_{j=1}^4 | \widetilde{y}_{n,j}^{(k)} - y_j | \; q^{(k)}_n(x^{(i)}_n) \; d \upsilon_S.
\end{align*}
In the derivation above, $\widetilde{y}_n^{(k)} = (\widetilde{y}_{n,1}^{(k)},  \widetilde{y}_{n,2}^{(k)}, \widetilde{y}_{n,3}^{(k)})$ is the ground truth state for the factor $k$ at time $n$ from the simulation,
$\widetilde{y}_{n,4}^{(k)} \triangleq 1 - \sum_{i=1}^3 \widetilde{y}_{n,i}^{(k)}$, 
$y = (y_1, y_2, y_3)$, and $y_4 \triangleq 1 - \sum_{i=1}^3 y_i$.
Also, if  $h : X \rightarrow \mathcal{D}(Y)$ is a conditional density and $\nu$ a measure on $Y$, 
then $h \cdot \nu : X \rightarrow \mathcal{P}(Y)$ is the probability kernel defined by 
$h \cdot \nu = \lambda x. \lambda B. \int_Y \mathbf{1}_B \, h(x) \, d\nu$.
There is an analogous definition if $h : \mathcal{D}(Y)$ is a density.
So $q^{(k)}_n \cdot \upsilon_S$ is a probability kernel that has $q^{(k)}_n$ as its corresponding conditional density.

%% file: appendix_practical_consideration.tex
\section{Practical Considerations}
\label{appendix:prac}

Some practical considerations of applying the particle versions of the filtering algorithms, namely, Algorithms~\ref{algo:spf},~\ref{algo:cpf},~\ref{algo:fpf},~and~\ref{algo:fcpf}, are presented here.

In the standard particle filter (Algorithm~\ref{algo:spf}), the transition update involves sampling from a mixture of Dirac measures
\[
\widebar{y}_n^{(i)} \sim \frac{1}{N} \sum_{i'=1}^N \tau_n(h_{n-1}, a_n, y_{n-1}^{(i')}), \text{ for } i=1,\dots,N.
\]
Typically, this approach can be implemented by first sampling an index $i'$ from the set $\{1, \dots, N\}$ uniformly at random, then applying the $n$th component of the transition schema $\tau$ to the particle $y_{n-1}^{(i')}$ (along with the history and action arguments).
The average-case running time of sampling an index
is $O(1)$, irrespective of the value of $N$, using a typical pseudorandom number generator and a sampling method like the algorithm in~\cite{Lemire2019fast};
the worst-case running time is $O(2^W)$ ($W$ is the bit width of the computer, for example, $W=64$, for a 64-bit computer) which is constant time complexity for a fixed $W$.
But, in practice, an implementation such as 
\[
\widebar{y}_n^{(i)} \sim \tau_n(h_{n-1}, a_n, y_{n-1}^{(i)}), \text{ for } i=1,\dots,N,
\]
is usually more efficient and produces a particle family with lower variance~\cite{Pitt2001} and thus is preferred.
Similar remarks apply to 
Algorithms~\ref{algo:cpf},~\ref{algo:fpf},~and~\ref{algo:fcpf}.

Another consideration with regard to the particle filtering algorithms is the resampling of particles.
Many resampling methods have been developed in the literature, as summarized in~\cite{Li2015resampling}.
To preserve the diversity of the parameter particle family, we adopt a modified resampling approach~\cite{Liu2001theoretical} with respect to the effective sample size (ESS)~\cite{Doucet2000sequential, Kong1994sequential} of particles when Algorithm~\ref{algo:spf} is paired with Algorithms~\ref{algo:fcf},~\ref{algo:fcpf},~and~\ref{algo:fcvf} to estimate parameters.
In particular, the (normalized) weights of particles $(x_n^{(i)})_{i=1}^N$ at time step $n$ are
\[
\overline{w}_n^{(i)} \propto (w_n^{(i)})^{\frac{1}{T}} \text{ for } i = 1,\dots,N,
\]
and we determine the value of $T$ such that the estimated ESS of the particles
\[
\widehat{\text{ESS}}((x_n^{(i)})_{i=1}^N) = \frac{1}{\sum_{i=1}^N (\overline{w}_n^{(i)})^2},
\]
is no less than a given threshold.